\pdfoutput=1

\documentclass[11pt]{article}

\usepackage[preprint]{acl}

\usepackage{times}
\usepackage{latexsym}

\usepackage[T1]{fontenc}

\usepackage[utf8]{inputenc}

\usepackage{microtype}

\usepackage{inconsolata}

\usepackage{graphicx}
\usepackage{amsmath}
\usepackage{amssymb}
\usepackage{booktabs}
\usepackage{multirow}
\usepackage{enumitem}
\usepackage{color}
\usepackage{tikz}
\usepackage[edges]{forest}
\definecolor{hidden-draw}{RGB}{0,0,0}
\definecolor{hidden-pink}{rgb}{0.98, 0.94, 0.75}
\definecolor{level0}{rgb}{0.67, 0.88, 0.69}
\definecolor{level1}{rgb}{0.98, 0.92, 0.84}
\definecolor{level2}{rgb}{0.8, 0.8, 1.0}
\definecolor{level3}{rgb}{1.0, 0.71, 0.76}

\newcommand{\grayshade}{\colorbox{gray!25}}

\newtheorem{definition}{Definition}[section]

\title{Deep Learning and Foundation Models for Weather Prediction: A Survey}

\author{
 \textbf{Jimeng Shi\textsuperscript{1}}$^{\dag}$, 
 \textbf{Azam Shirali\textsuperscript{1}},
 \textbf{Bowen Jin}\textsuperscript{2},
 \textbf{Sizhe Zhou}\textsuperscript{2}, 
 \textbf{Wei Hu}\textsuperscript{2}, 
 \textbf{Rahuul Rangaraj\textsuperscript{1}},\\
 \textbf{Shaowen Wang}\textsuperscript{2},
 \textbf{Jiawei Han}\textsuperscript{2},
 \textbf{Zhaonan Wang\textsuperscript{3}}, 
 \textbf{Upmanu Lall\textsuperscript{4,5}}, \\
 \textbf{Yanzhao Wu\textsuperscript{1}},
 \textbf{Leonardo Bobadilla\textsuperscript{1}},
 \textbf{Giri Narasimhan\textsuperscript{1}$^{\dag}$}
\\
\\
 \textsuperscript{1}Florida International University \quad 
 \textsuperscript{2}University of Illinois Urbana-Champaign \quad \\ 
 \textsuperscript{3}New York University Shanghai \quad
 \textsuperscript{4}Arizona State University \quad
 \textsuperscript{5}Columbia University
\\
\\
 $^{\dag}\textrm{Corresponding authors:}$ \texttt{\{jshi008,giri\}@fiu.edu}
}

\begin{document}
\maketitle
\begin{abstract}
Physics-based numerical models have been the bedrock of atmospheric sciences for decades, offering robust solutions but often at the cost of significant computational resources.
Deep learning (DL) models have emerged as powerful tools in meteorology, capable of analyzing complex weather and climate data by learning intricate dependencies and providing rapid predictions once trained. 
While these models demonstrate promising performance in weather prediction, often surpassing traditional physics-based methods, they still face critical challenges.
This paper presents a comprehensive survey of recent deep learning and foundation models for weather prediction.
We propose a taxonomy to classify existing models based on their training paradigms: \textit{deterministic predictive} learning, \textit{probabilistic generative} learning, and \textit{pre-training and fine-tuning}. 
For each paradigm, we delve into the underlying model architectures, address major challenges, offer key insights, and propose targeted directions for future research. 
Furthermore, we explore real-world applications of these methods and provide a curated summary of open-source code repositories and widely used datasets, aiming to bridge research advancements with practical implementations while fostering open and trustworthy scientific practices in adopting cutting-edge artificial intelligence for weather prediction.
The related sources are available at \url{https://github.com/JimengShi/DL-Foundation-Models-Weather}.
\end{abstract}

\section{Introduction}
\label{sec:intro}
\begin{figure}[ht]
\centering
    \includegraphics[width=0.999\columnwidth]{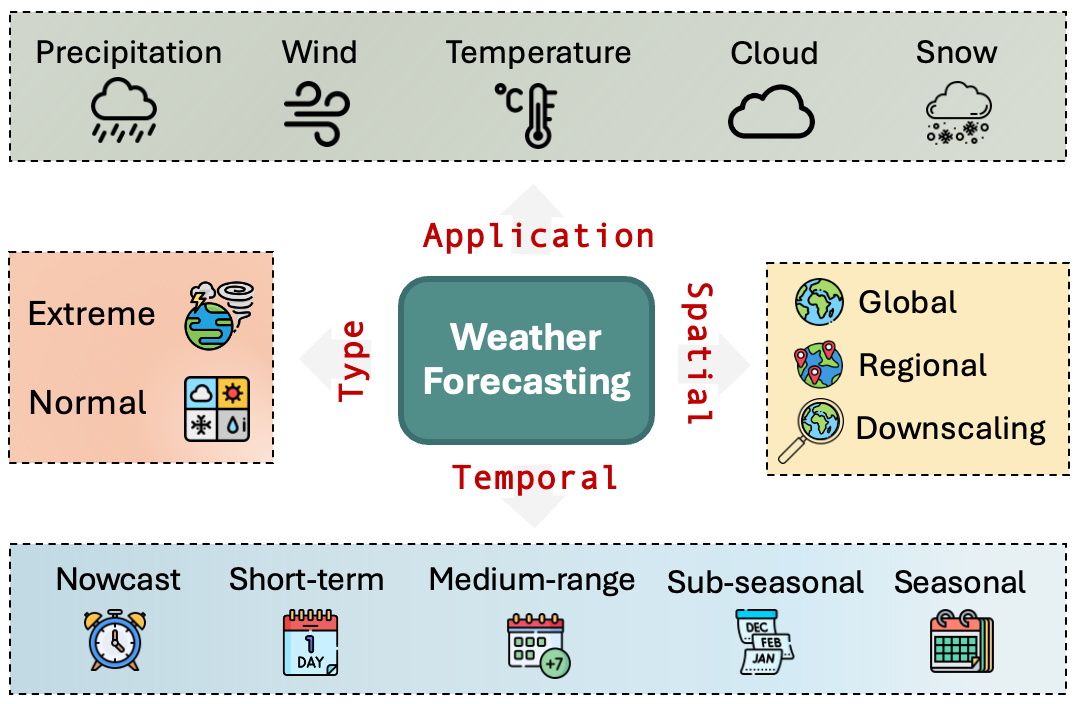}
    \caption{Perspectives of weather forecasting.}
\label{fig:weather_forecast}
\vspace{-3mm}
\end{figure}
%
%
Global climate change has increased the frequency of extreme weather events, such as heatwaves, extreme cold spells, intense rainfall, storms, and hurricanes, leading to disasters such as droughts, floods, and air pollution. These changes have profound implications across multiple domains, affecting human health and activities~\cite{flandroy2018impact}, compromising environmental sustainability~\cite{abbass2022review}, disrupting economic stability~\cite{carleton2016social}, and altering ecosystem dynamics~\cite{descombes2020novel}.
In this context, developing accurate and timely weather prediction is critical to mitigating these impacts and supporting adaptive strategies.

Physics-based models, including General Circulation Models (GCMs)~\cite{ravindra2019generalized} and Numerical Weather Prediction (NWP) models~\cite{coiffier2011fundamentals}, have been the cornerstone of weather prediction. These models simulate future weather scenarios by numerically approximating solutions to the differential equations that govern the complex physical dynamics of interconnected atmospheric, terrestrial, and oceanic systems~\cite{nguyen2023climax}.
Despite significant advancements, these models face notable limitations. 
Firstly, they are computationally intensive due to the high-dimensional and nonlinear nature of the governing equations~\cite{ren2021deep}. 
Secondly, the underlying equations often rely on simplified assumptions about atmospheric dynamics, which can limit their ability to capture intricate, uncommon processes~\cite{palmer2005representing}. 
Lastly, these physics-based models typically produce deterministic forecasts based on initial conditions, falling short of explicitly capturing model uncertainties in weather evolution even though perturbation of initial conditions has been used to represent the input uncertainty~\cite{bulte2024uncertainty}.

ARIMA (AutoRegressive Integrated Moving Average) is a statistical model widely used for weather prediction \cite{box2015time}. Non-seasonal ARIMA models analyze patterns in historical data but cannot handle seasonality, while seasonal ARIMA extends this framework to account for regular cycles, making it effective for variables like rainfall or temperature \cite{lai2020use,khan2023short}. 
However, ARIMA models have limitations, including difficulty capturing nonlinear relationships, sensitivity to outliers, and the need for careful parameter selection. Bayesian nonparametric nonhomogeneous hidden Markov model is another statistical method that has been studied for predicting daily rainfall~\cite{cao2024predictability} and ENSO impacts~\cite{zhang2024potential}. However, these methods are usually applied to univariate or low-dimensional responses.

In recent years, data-driven machine learning (ML) and deep learning (DL) models have been increasingly applied to weather and climate modeling, demonstrating remarkable advances in precision, computational efficiency, and uncertainty quantification~\cite{chen2023artificial, nguyen2023climatelearn}.
For example, deterministic models such as \texttt{Pangu}~\cite{bi2023accurate} and \texttt{GraphCast}~\cite{lam2022graphcast} have achieved state-of-the-art performance in medium-range (10-day) global weather prediction, surpassing or matching traditional methods in accuracy while dramatically reducing computational costs (up to three orders of magnitude). However, their predictions are often blurry since they are trained by minimizing point-wise loss functions.
To overcome this limitation, probabilistic generative models have emerged as powerful tools for weather prediction while achieving uncertainty quantification in those predictions.
They consider weather prediction as probabilistic sampling (i.e., generation) conditioning on necessary constraints.
Models like \texttt{CasCast}~\cite{gong2024cascast} and \texttt{Gencast}~\cite{price2023gencast} leverage probabilistic diffusion techniques for tasks such as precipitation nowcasting and weather prediction, delivering both high-quality predictions and calibrated uncertainty estimates.
More recently, foundation models have gained traction in climate and weather modeling as an emerging paradigm~\cite{bodnar2024aurora, schmude2024prithvi}. These models are pre-trained on massive historical weather datasets to learn generalizable and comprehensive knowledge, which can then be fine-tuned for diverse downstream tasks~\cite{chen2023foundation}. Foundation models offer two key advantages: (1) the ability to learn robust and transferable weather representations from large-scale data, and (2) the flexibility to adapt to downstream applications without the need for task-specific models trained from scratch~\cite{miller2024survey, zhu2024foundations}.

With the rapid advancement of deep learning in weather and climate science, a systematic and up-to-date survey is critical to consolidating knowledge and guiding future research. While several surveys were published in recent years, each has a distinct focus.
\citet{ren2021deep} reviewed DL models for weather prediction, emphasizing their architectural designs.
\citet{molina2023review} summarized DL applications in climate modeling, covering feature detection, extreme weather prediction, downscaling, and bias correction.
Moreover, surveys~\cite{fang2021survey,materia2023artificial} focused on DL techniques for weather forecasting in specific scenarios, such as extreme weather events.
\citet{mukkavilli2023ai} discussed state-of-the-art DL models across diverse meteorological applications, highlighting their effectiveness over varying spatial and temporal scales.
\citet{chen2023foundation} categorized DL models for weather and climate science by data modality (e.g., time series, text) and their applications.
Distinct from existing surveys, our work provides a novel perspective by reviewing the literature through the lens of training paradigms and offering a broader discussion on future research directions. Our contributions are:
\begin{itemize}[itemsep=0pt, topsep=1pt]
    \item \textbf{Novel Taxonomy.} We introduce a systematic categorization of existing DL models for weather prediction based on their training paradigms: predictive learning, generative learning, and pre-training and fine-tuning.
    \item \textbf{Comprehensive Overview.} We present a detailed survey of the state-of-the-art models, analyzing their strengths, limitations, and applications in weather prediction.
    \item \textbf{Extensive Resources.} We compile an extensive repository of resources, including benchmark datasets, open-source codes, and real-world applications to support further research.
    \item \textbf{Future Directions.} We outline a forward-looking roadmap, highlighting \emph{ten} critical research directions across \emph{five} key avenues to advance DL methods for weather prediction.
\end{itemize}

\section{Background and Preliminaries}
\label{sec:background}

\subsection{Weather Data Representation}
There are two primary types of weather data commonly used: \textit{station-based observation} data and \textit{gridded reanalysis} data. 
Each offers unique advantages and limitations and both play critical roles in advancing weather and climate research.

\paragraph{Station-Based Observation Data.}
It originates from weather stations distributed across the globe, collecting high-resolution meteorological measurements at specific locations. 
These stations provide precise monitoring data, for example, temperature, humidity, wind speed and direction, precipitation, atmospheric pressure, and more.
Station-based observations are typically of high temporal resolution, with data recorded hourly or daily, enabling detailed insights into local weather patterns and trends. 
However, station coverage is often uneven, with a high concentration in populated or economically significant areas and sparse coverage in remote regions such as the oceans, mountains, and deserts. 
This uneven distribution can limit global-scale analyses, though it remains invaluable for localized forecasting, trend analysis, and model validation.

\paragraph{Gridded Reanalysis Data.}
It offers a global view by dividing the Earth's surface into a grid, with each cell assigned values representing averaged weather conditions over its area. 
It is often called reanalysis data, derived from a combination of sources, including station observations, satellite measurements, and numerical weather prediction (NWP) models.
Gridded data provide consistent spatial coverage, including remote areas and oceans, where station-based observations are sparse or nonexistent.
Gridded data are typically available at varying resolutions, with common grid sizes ranging from $1^\circ\times1^\circ$ to $0.25^\circ\times0.25^\circ$ (each degree corresponds to about 100 km). Temporal resolution can also vary, offering hourly or daily intervals, allowing for detailed temporal analysis.
\subsection{Weather Prediction Formulation}
As shown in Figure \ref{fig:weather_forecast}, we discuss four types of weather forecasting.
(1) \emph{Temporal}: forecasts predict atmospheric variables of interest for future time point(s), $t+\Delta t$, given observation(s) from the recent past. It includes weather and climate forecasts based on the lead time $\Delta t \approx$ \{hours, days, weeks, months, years\} and encompasses nowcast, medium-range forecast, sub-seasonal, and seasonal forecast. 
(2) \emph{Spatial}: methods predict global and regional weather forecasts for any given time point.
(3) \emph{Applications}: focus on predicting weather variables of interest.
(4) \emph{Event Type}: Weather forecasts may be for extreme events, such as heatwaves, snowstorms, hurricanes, and tropical cyclones. Forecasts could also be for regular, non-extreme periods.

Deterministic weather and climate forecasting can be formulated as follows:
\begin{equation}
    \left[X_{t - (\alpha-1)}, \dots, X_t \right] \xrightarrow{\mathcal{F}(\theta)} \left[ Y_{t+1}, \dots, Y_{t+\beta} \right],
\end{equation}
where $X$ and $Y$ are sets of input and output variables; $\alpha$ and $\beta$ are the temporal lengths of the input and output windows; $\mathcal{F}(\theta)$ represents the model with the learnable parameters $\theta$.
$\mathcal{F}(\cdot)$ can also denote a probabilistic function, i.e., $Y \sim \mathcal{P}(Y|X)$.

\subsection{Preliminaries}
We identify three types of weather models.
\begin{definition}[General-Purpose Large Models] They are typically trained on large, diverse global datasets that include information on \underline{multiple} meteorological variables of interest, enabling \underline{global} weather prediction across a broad spectrum of applications. 
\end{definition}
\begin{definition}[Domain-Specific Models] They focus on predicting a \underline{single} variable, applied to \underline{regional} weather prediction. 
\end{definition}
\begin{definition}[Foundation Models] They are large models \underline{pre-trained} on diverse, massive datasets, allowing for subsequent \underline{fine-tuning} or adaptation for various downstream tasks. 
\end{definition}

\noindent Based on the modeling algorithm, we have deterministic and probabilistic training paradigms. Both general-purpose large models and domain-specific models can be trained with deterministic predictive learning (Section \ref{sec:predictive}) or probabilistic generative learning (Section \ref{sec:generative}). Foundation Models are pre-trained and then fine-tuned (Section \ref{sec:foundations}).

\begin{figure*}[ht!]
\centering
  \includegraphics[width=2\columnwidth]{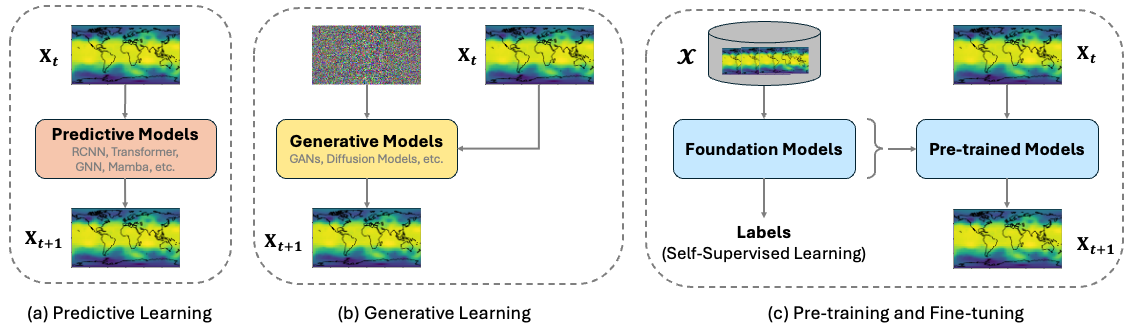}
  \caption{The illustration of various frameworks of training deep learning models on weather prediction. For clarity, this visualization focuses exclusively on single-step predictions for a single variable.}
  \label{fig:frameworks}
\end{figure*}
\section{Overview and Taxonomy}
\label{sec:overview}
\begin{figure*}[t]
\centering
    \resizebox{0.98\textwidth}{!}{
        \begin{forest}
            forked edges,
            for tree={
                fill=level0!80,
                grow=east,
                reversed=true,
                anchor=base west,
                parent anchor=east,
                child anchor=west,
                base=left,
                font=\large,
                rectangle,
                draw=hidden-draw,
                rounded corners,
                align=left,
                minimum width=6em,
                edge+={darkgray, line width=1pt},
                s sep=3pt,
                inner xsep=2pt,
                inner ysep=3pt,
                line width=0.8pt,
                ver/.style={rotate=90, child anchor=north, parent anchor=south, anchor=center},
            },
            where level=1{text width=10.5em,font=\normalsize,fill=level1!90,}{},
            where level=2{text width=9em,font=\normalsize,fill=level2!80,}{},
            where level=3{text width=5.5em,font=\normalsize,fill=level3!60,}{},
            where level=4{text width=28.5em,font=\normalsize,fill=level1!40,}{},
            [AI Models \\ for Weather \\ Prediction
                [
                    Deterministic Predictive \\ Learning (Section \ref{sec:predictive})
                    [
                        General-Purpose \\
                        \textbf{Large Models}
                        [
                            Transformer
                            [
                                FourCastNet~\cite{pathak2022fourcastnet}{,} 
                                FuXi~\cite{chen2023fuxi}{,} \\
                                FengWu~\cite{chen2023fengwu}{,} 
                                FengWu-4DVar~\cite{xiao2023fengwu}{,}\\
                                SwinVRNN~\cite{hu2023swinvrnn}{,}
                                SwinRDM~\cite{chen2023swinrdm}{,} \\
                                Pangu-Weather~\cite{bi2023accurate}{,}
                                Stormer~\cite{nguyen2023scaling}{,}\\
                                HEAL-ViT~\cite{ramavajjala2024heal}{,}
                                TianXing~\cite{yuan2025tianxing}
                            ]
                        ]
                        [
                            GNN
                            [   
                                GraphCast~\cite{lam2022graphcast}{,} GnnWeather~\cite{keisler2022forecasting}{,}\\
                                AIFS~\cite{lang2024aifs}{,}
                                GraphDOP~\cite{alexe2024graphdop}
                            ]
                        ]
                        [
                            PhysicsAI
                            [
                                ClimODE~\cite{verma2024climode}{,}
                                WeatherODE~\cite{liu2024mitigating}{,}\\
                                NeuralGCM~\cite{kochkov2024neural}{,}
                                Conformer~\cite{saleem2024conformer}
                            ]
                        ]
                    ]
                    [
                        Domain-Specific \\ Models
                        [
                            Transformer
                            [
                                SwinUnet~\cite{bojesomo2021spatiotemporal}{,}
                                Earthformer~\cite{gao2022earthformer}{,}\\
                                Rainformer~\cite{bai2022rainformer}{,}
                                U-STN~\cite{chattopadhyay2022towards}{,}\\
                                OMG-HD~\cite{zhao2024omg}{,}
                                PFformer~\cite{xu2024pfformer}
                            ]
                        ]
                        [
                            GNN
                            [
                                HiSTGNN~\cite{ma2023histgnn}{,} 
                                $w$-GNN~\cite{chen2024coupling}{,}\\
                                WeatherGNN~\cite{wu2024weathergnn}{,} MPNNs~\cite{yang2024multi}
                            ]
                        ]
                        [
                            RNN\&CNN
                            [   
                                MetNet~\cite{sonderby2020metnet,espeholt2022deep}{,}\\
                                MetNet-3~\cite{andrychowicz2023deep}{,}
                                PredRNN~\cite{wang2022predrnn}{,}\\
                                MM-RNN~\cite{ma2023mm}{,}
                                ConvLSTM~\cite{shi2015convolutional}
                            ]
                        ]
                        [
                            Mamba
                            [
                                MetMamba~\cite{qin2024metmamba}{,}
                                MambaDS~\cite{liu2024mambads}
                            ]
                        ]
                        [
                            PhysicsAI
                            [
                                NowcastNet~\cite{zhang2023skilful}{,}
                                PhysDL~\cite{de2019deep}{,}\\
                                PhyDNet~\cite{guen2020disentangling}{,}
                                DeepPhysiNet~\cite{li2024deepphysinet}
                            ]
                        ]
                    ]
                ]
                [
                    Probabilistic Generative \\ Learning (Section \ref{sec:generative})
                    [
                        General-Purpose \\ \textbf{Large Models}
                        [
                            Diffusion
                            [ 
                                GenCast~\cite{price2023gencast}{,} CoDiCast~\cite{shi2024codicast}{,}\\
                                SEEDs~\cite{li2023seeds}{,}
                                ContinuousEnsCast~\cite{andrae2024continuous}
                            ]
                        ]
                    ]
                    [
                        Domain-Specific \\ Models
                        [
                            Diffusion
                            [
                                LDMRain~\cite{leinonen2023latent}{,} PreDiff~\cite{gao2023prediff}{,}\\
                                CasCast~\cite{gong2024cascast}{,} SRNDiff~\cite{ling2024srndiff}{,}\\
                                DiffCast~\cite{yu2024diffcast}{,} 
                                GEDRain~\cite{asperti2023precipitation}
                            ]
                        ]
                        [
                            GANs
                            [
                                GANrain~\cite{ravuri2021skilful}{,}
                                MultiScaleGAN~\cite{luo2022experimental}{,}\\
                                STGM~\cite{wang2023physical}{,}
                                PCT-CycleGAN~\cite{choi2023pct}
                            ]
                        ]
                    ]
                ]
                [   
                    Pretraining \& Finetuning  \\ (Section \ref{sec:foundations})
                    [
                        \textbf{Foundation Models}
                        [
                            Transformer
                            [
                                ClimaX~\cite{nguyen2023climax}{,} W-MAE~\cite{man2023w}{,}\\
                                Aurora~\cite{bodnar2024aurora}{,} Prithvi WxC~\cite{schmude2024prithvi}
                            ]
                        ]
                    ]
                ]
            ]
        \end{forest}
    }
\caption{A comprehensive taxonomy of deep learning and foundation models for weather prediction from the perspectives of training paradigms (dark yellow), model scopes (purple), and model architectures (pink).}
\vspace{-3mm}
\label{fig:taxonomy}
\end{figure*}
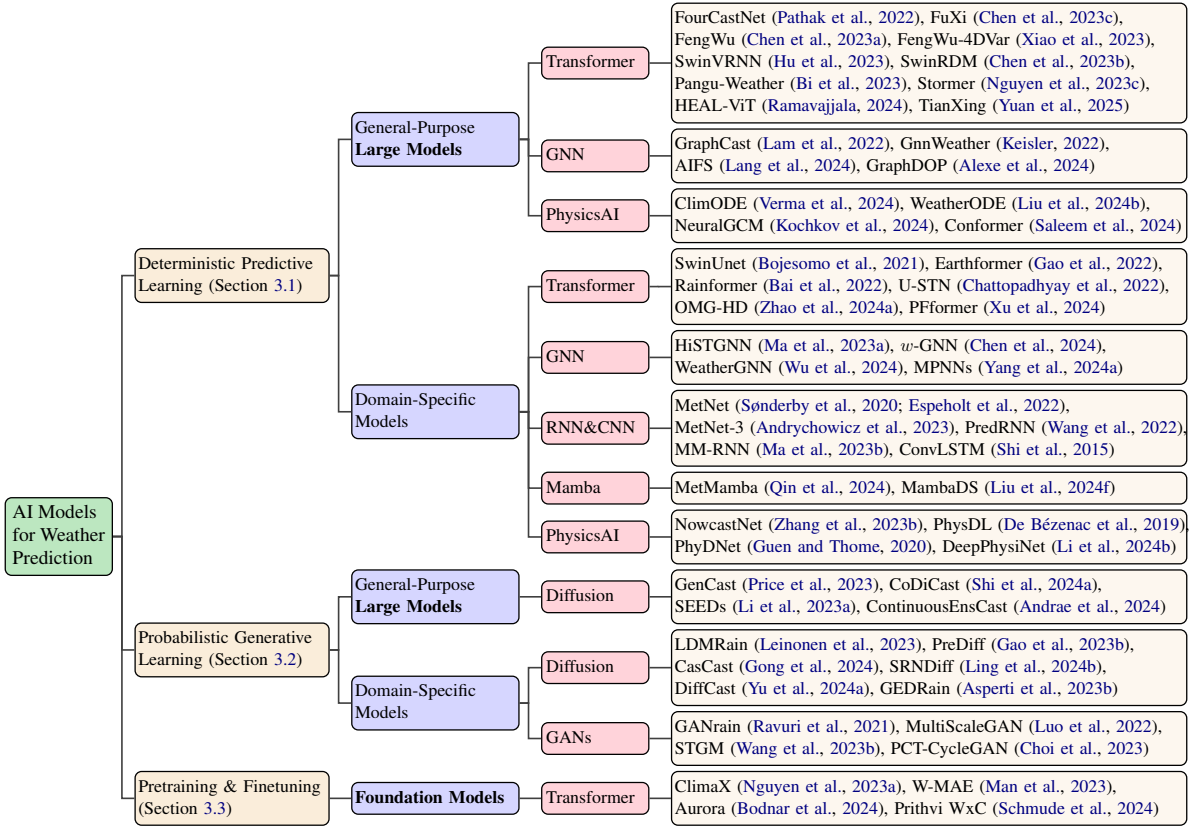

This section provides an overview and categorization of deep learning (DL) models for weather forecasts. Our survey mainly focuses on three aspects: modeling paradigm, model backbone, and application domain. 
The modeling paradigm includes deterministic \emph{predictive} learning, probabilistic \emph{generative} learning, and \emph{pre-training and fine-tuning} (see Figure \ref{fig:frameworks}).
Weather and climate models can be categorized based on model backbones, such as Recurrent Neural Networks, Transformers, Graph Neural Networks, Mamba, Generative Adversarial Networks, and Diffusion Models. 
The theoretical details of these models are provided in Appendix \ref{sec:model_arch}.
At the application level, the existing models can be divided into general-purpose and domain-specific models. We present a detailed comparison and summary in Table \ref{tab:genral_domain_compare} and Figure \ref{fig:taxonomy}.
\begin{table}[ht]
\centering
\caption{General-Purpose Large Models vs Domain-Specific Models.}
\vspace{-2mm}
\resizebox{.999\columnwidth}{!}{
    \begin{tabular}{l|l|l}
    \toprule
    \textbf{ }      & \textbf{General-Purpose Large Models} & \textbf{Domain-Specific Models}   \\ 
    \midrule
    Scope           & Global, multi-variable                & Regional forecasts, single-variable \\ 
    Spatial         & Coarse ($0.25^\circ\sim5.625^\circ$)  & High ($\leq0.1^\circ$)  \\ 
    Temporal        & Coarse ($6~12$ hours)                 & High ($5 \text{ mins} \sim 1 \text{ hour}$)  \\
    Training Data   & $\geq 10$ Years                       & Days, Months, Years  \\ 
    Architectures   & Transformer, GNN                      & Transformer, GNN, RNN, CNN, Mamba \\ 
    \bottomrule
    \end{tabular}
}
\label{tab:genral_domain_compare}
\end{table}

\tikzstyle{my-box}=[
    rectangle,
    draw=hidden-draw,
    rounded corners,
    text opacity=1,
    minimum height=1.5em,
    minimum width=5em,
    inner sep=2pt,
    align=center,
    fill opacity=.4,
    line width=0.8pt,
]
\tikzstyle{leaf}=[my-box, minimum height=1.5em,
    fill=hidden-pink!80, text=black, align=left, font=\normalsize,
    inner xsep=2pt,
    inner ysep=4pt,
    line width=0.8pt,
]

\subsection{Predictive Learning}
\label{sec:predictive}
\emph{Predictive} learning methods are usually \emph{deterministic}, where models aim to predict future states of weather variables (like temperature, humidity, wind speed, and precipitation) based on past and present observations. These models are typically built to recognize weather patterns or dependencies in historical data by minimizing a point-wised loss function (e.g., mean absolute errors). We systematically categorize these predictive models into general-purpose large models and domain-specific models. Each categorization is discussed with various model architectures.
\subsubsection{General-Purpose Large Models}
Large Language Models (LLMs)~\cite{zhao2023survey} have garnered significant attention in recent years. Similarly, large-scale weather models have been developed to address global weather prediction tasks across multiple meteorological variables, leveraging deterministic predictive frameworks.

\paragraph{Transformer-based models.} Transformer models~\cite{vaswani2017attention} are widely used as a backbone.  \texttt{FourCastNet}~\cite{pathak2022fourcastnet} is developed for global data-driven weather forecasting by employing a Fourier transform-based token-mixing scheme~\cite{guibas2021adaptive} with a vision transformer (ViT)~\cite{dosovitskiy2020image}. 
The multiple-time step prediction is achieved by using trained models in autoregressive inference mode. 
\texttt{FengWu}~\cite{chen2023fengwu} processes each weather variable separately, using multiple encoders to extract individual feature embeddings. Then, an elaborately designed transformer network fuses these embeddings to capture the interaction among all variables. As with \texttt{FourCastNet}, \texttt{Fengwu} also autoregressively forecasts multiple steps over a long range.
\texttt{FengWu-4DVar}~\cite{xiao2023fengwu} integrates \texttt{FengWu} with the Four-Dimensional Variational (4DVar) assimilation algorithm~\cite{rabier1998extended}, accomplishing both global weather forecasting and data assimilation.
\texttt{SwinVRNN}~\cite{hu2023swinvrnn} utilizes the Swin Transformer~\cite{liu2022swin} and RNN for weather prediction, but with a perturbation module to generate ensemble forecasts. SwinRDM~\cite{chen2023swinrdm} uses SwinRNN for prediction and a diffusion model for super-resolution output. 
\texttt{HEAL-ViT}~\cite{ramavajjala2024heal} explores Vision Transformers on a spherical mesh, benefiting from both spatial homogeneity inherent in graphical models and efficient attention mechanisms. 
The \texttt{TianXing} model~\cite{yuan2025tianxing} proposes a variant attention mechanism with linear complexity for global weather prediction, significantly diminishing GPU resource demands, with only a marginal compromise in accuracy.\\
\indent While these models have achieved impressive performance, any iterative inference process accumulates errors as the length of the prediction window increases. 
The \texttt{Pangu-Weather}~\cite{bi2023accurate} model uses a hierarchical temporal aggregation algorithm to alleviate cumulative forecast errors.  
They train four individual models for lead times of 1, 3, 6, and 24 hours. In the testing stage, the greedy algorithm is used to guarantee the minimal number of iterations of the trained models for a forecast window. 
Furthermore, they design a 3D Earth Specific Transformer (3DEST) architecture that formulates the height (pressure level) information into cubic data, capturing more intricate spatiotemporal dynamics.
Similarly, the \texttt{FuXi} model~\cite{chen2023fuxi} employed a combination of FuXi-Short, FuXi-Medium, and FuXi-Long models to produce 15-day forecasts, where each model generates 5-day forecasts. Its backbone is a U-transformer, coupling U-Net~\cite{ronneberger2015u}, and a Swin Transformer~\cite{liu2022swin}. 
In addition to the integration of direct and iterative forecasting, the \texttt{Stormer} model~\cite{nguyen2023scaling} needs the explicit time point, $t+\Delta t$ to guide the models for predictions.

\paragraph{GNN-based models.} \citet{keisler2022forecasting} introduced an approach to global weather prediction using graph neural networks (GNNs)~\cite{wu2020comprehensive}. By modeling the Earth as a graph with nodes representing spatial locations and edges encoding their relationships, the model captures spatial dependencies in weather patterns. 
This GNN-based method effectively integrates local and global weather dynamics. 
Another GNN-based model, \texttt{GraphCast}~\cite{lam2022graphcast}, forecasts hundreds of weather variables with a longer forecast range (up to 10 days ahead) at a higher spatial resolution (0.25 degree) after training with reanalysis gridded ERA5 data~\cite{rasp2023weatherbench}. 
It also provides better support for severe weather compared to the European Centre for Medium-Range Weather Forecasts (ECMWF)’s High-RESolution forecast (HRES), a component of the Integrated Forecast System (IFS).
More recently, ECMWF also proposed GNN-based models, \texttt{AIFS}~\cite{lang2024aifs} and \texttt{GraphDOP}~\cite{alexe2024graphdop}. The latter is a model that operates solely on inputs and outputs in observation space, with no gridded climatology and/or NWP (re)analysis inputs or feedback.

\paragraph{Physics-AI-based models.} Although data-driven methods have demonstrated high accuracy and efficiency, they operate as black-box models that frequently overlook underlying physical mechanisms, such as turbulence, convection, and atmospheric airflow.
\texttt{ClimODE}~\cite{verma2024climode} implements a key principle of \emph{advection} to model a spatiotemporal continuous-time process, namely, weather changes due to the spatial movement over time.
It aims to precisely describe the value-conserving dynamics of weather evolution with continuity ODE~\cite{marchuk2012numerical}, learning global weather transport as a neural flow.
It also includes a Gaussian emission network for predicting uncertainties and source variations.
To solve the advection equation more accurately, \texttt{WeatherODE}~\cite{liu2024mitigating} adopts wave equation theory~\cite{evans2022partial} and a time-dependent source model and designs the CNN-ViT-CNN sandwich structure, facilitating efficient learning dynamics tailored for distinct yet interrelated tasks with varying optimization biases.
\texttt{NeuralGCM}~\cite{kochkov2024neural} employs a differentiable dynamical core for solving \emph{more} primitive equations, including momentum equations, the second law of thermodynamics, a thermodynamic equation of state, continuity equation, and hydrostatic approximation.
It also develops a learned physics module that parameterizes physical processes with a neural network, predicting the effect of unresolved processes such as cloud formation, radiative transport, precipitation, and subgrid-scale dynamics.
\texttt{Conformer}~\cite{saleem2024conformer} is a spatiotemporal Continuous Vision Transformer for weather forecasting, learning the continuous weather evolution over time by implementing continuity in the multi-head attention mechanism.

\subsubsection{Domain-Specific Models} 
We present domain-specific predictive models for regional or single-variable weather predictions.
\paragraph{Transformer-based models.} 
\texttt{SwinUnet}~\cite{bojesomo2021spatiotemporal} employs the hybrid model of Swin Transformer and U-Net for regional weather forecasts in Europe.
\texttt{Earthformer}~\cite{gao2022earthformer} proposes a generic, flexible, and efficient space-time attention block (Cuboid Attention) Earth system forecasting, which can decompose the data into cuboids and apply cuboid-level self-attention in parallel. 
\texttt{Rainformer}~\cite{bai2022rainformer} combines CNN and Swin Transformer for precipitation nowcasting.
\texttt{PFformer}~\cite{xu2024pfformer} utilizes i-Transformer~\cite{liu2023itransformer} to learn spatial dependencies among multiple observation stations for short-term precipitation forecasting.
Vision transformer~\cite{dosovitskiy2020image} has been applied to estimate lightning intensity in Ningbo City, China \cite{lu2022vision}.
\texttt{NowcastingGPT}~\cite{meo2024extreme} develops Transformer-based models with Extreme Value Loss (EVL) regularization~\cite{von1921variationsbreite}
for extreme precipitation nowcasting.
The \texttt{U-STN} model~\cite{chattopadhyay2022towards} integrates data assimilation with a deep spatial-transformer-based U-NET to predict the global geopotential while the \texttt{OMG-HD} model~\cite{zhao2024omg} leverages the Swim Transformer for regional high-resolution weather forecast trained with multiple observational data, including stations, radar, and satellite.

\paragraph{GNN-based models.}
\texttt{HiSTGNN}~\cite{ma2023histgnn} incorporates an adaptive graph learning module comprising a global graph representing regions and a local graph capturing meteorological variables for each region. 
The \texttt{$w$-GNN} model~\cite{chen2024coupling} leverages Graph Neural Networks coupled with physical factors for precipitation forecast in China.
\texttt{WeatherGNN}~\cite{wu2024weathergnn} proposes a fast hierarchical Graph Neural Network (FHGNN) to extract the spatial dependencies.
The \texttt{MPNN} model~\cite{yang2024multi} exploits heterogeneous GNNs for both station-observed and gridded weather data, where the node at the prediction location aggregates information from its heterogeneous neighboring nodes by message passing.

\paragraph{RNN- \& CNN-based models.}
The \texttt{ConvLSTM} model~\cite{shi2015convolutional} couples CNNs and LSTMs as the model backbone for precipitation nowcasting, usually with a lead time between 1 to 3 hours. 
Similar works include \texttt{MetNet-1}~\cite{sonderby2020metnet} and \texttt{MetNet-2} models~\cite{espeholt2022deep} for precipitation forecasting for lead times of 8 and 12 hours. 
\texttt{MetNet-3}~\cite{andrychowicz2023deep} significantly extends both the lead times (up to 24 hours) and variables (precipitation, wind, temperature, and dew point) by learning from both dense and sparse data sensors. 
\texttt{MM-RNN}~\cite{ma2023mm} introduces knowledge of elements to guide precipitation prediction and learn the underlying atmospheric motion laws using RNNs.
Based on the original LSTMs, \texttt{PredRNN}~\cite{wang2022predrnn} proposes a zigzag memory flow that propagates in both a bottom-up and top-down fashion across all layers, enabling the dynamic communication at various levels of RNNs.
Other variants of ConvLSTM for precipitation nowcasting include \texttt{TrajGRU}~\cite{shi2017deep} and \texttt{Predrnn++}~\cite{wang2018predrnn}.

\paragraph{Mamba-based models.}
\texttt{MetMamba}~\cite{qin2024metmamba} exploits Mamba’s selective scan to achieve token (spatial, temporal) mixing and channel mixing to capture more complex spatiotemporal dependencies in weather data.
\texttt{MambaDS}~\cite{liu2024mambads} attempts to use the selective state space model (Mamba) for the meteorological field downscaling.
\texttt{VMRNN}~\cite{tang2024vmrnn} develops an innovative architecture tailored for spatiotemporal forecasting by integrating Vision Mamba and LSTM, surpassing established vision models in both efficiency and accuracy.

\paragraph{Physics-AI-based models.}
\texttt{NowcastNet}~\cite{zhang2023skilful} is a nonlinear nowcasting model for extreme precipitation that unifies physical-evolution schemes and conditional-learning methods into a neural network framework.
\texttt{PhysicsAI}~\cite{das2024hybrid} has evaluated \texttt{NowcastNet} model with a case study on the Tennessee Valley Authority (TVA) service area, outperforming the High Resolution Rapid Refresh (HRRR) model.
\texttt{PhysDL}~\cite{de2019deep} presents how physical knowledge (\emph{advection} and \emph{diffusion}) could be used as a guideline for designing efficient deep-learning models, exemplifying sea surface temperature predictions.
\texttt{PhyDNet}~\cite{guen2020disentangling} is a two-branch deep learning architecture that explicitly disentangles known PDE dynamics from unknown complementary information.
\texttt{DeepPhysiNet}~\cite{li2024deepphysinet} incorporates atmospheric physics into the loss function of deep learning methods as hard constraints for accurate weather modeling.

More generally, we provide state-of-the-art predictive models for time series forecasting across various domains. While these models are not specific for weather modeling, they offer insightful modeling advancements since weather data is often represented as time series. Representative models include but not limited to \texttt{iTransformer}~\cite{liu2023itransformer}, \texttt{PatchTST}~\cite{nie2022time}, \texttt{FEDformer}~\cite{zhou2022fedformer}, \texttt{DLinear}~\cite{zeng2023transformers}, \texttt{Autoformer}~\cite{chen2021autoformer}. 
More recently, \citet{han2024weather} collected worldwide meteorological monitoring data, created a benchmark dataset, and completed a comprehensive evaluation with those advanced models above.

\subsection{Generative Models}
\label{sec:generative}
Generative models can be used for weather \emph{prediction} by treating them as \emph{generative} processes conditioned on observations from the past. More significantly, since these generative models are probabilistic, they are well suited to generate ensemble forecasts that can help quantify the uncertainty in the predictions, facilitating informed decision-making.
\subsubsection{General-Purpose Large Models}
\paragraph{Diffusion-based models.} Some researchers have developed generative models for global weather prediction. 
\texttt{GenCast}~\cite{price2023gencast} uses diffusion models for probabilistic weather forecasts conditioning on the past two observations, generating an ensemble of stochastic $15$-day global forecasts, at $12$-hour steps and $0.25^\circ$ latitude-longitude resolution, for over 80 surface and atmospheric variables.
As a variant of \texttt{GenCast}, \texttt{CoDiCast}~\cite{shi2024codicast} leverages a \emph{pre-trained} encoder to learn embeddings from observations from the recent past and a \emph{cross-attention} mechanism to guide the generation process to predict future weather states.
Similar work includes \texttt{SEEDs}~\cite{li2023seeds} for the global weather forecast.
The three methods above are trained on a single forecasting step and rolled out autoregressively.
However, they are computationally expensive and accumulate errors for high temporal resolution due to the many rollout steps.
\texttt{ContinuousEnsCast}~\cite{andrae2024continuous} addresses these limitations by proposing a continuous forecasting diffusion model that takes lead time as input and forecasts the future weather state in a single step while maintaining a temporally consistent trajectory for each ensemble member.

\subsubsection{Domain-Specific Models}
Here we discuss domain-specific models for generative learning with generative adversarial networks (GANs)~\cite{goodfellow2014generative,mirza2014conditional} and diffusion models~\cite{ho2020denoising}.
\paragraph{GAN-based models.}
\texttt{GANrain}~\cite{ravuri2021skilful} employs a conditional generative adversarial network (GAN) for the precipitation prediction problem, where the generator generates future precipitation frames and the discriminator learns to distinguish whether a sample is coming from the original training data or was generated by the generator.
\texttt{MultiScaleGAN}~\cite{luo2022experimental} evaluates GANs~\cite{goodfellow2014generative} and Wasserstein-GAN~\cite{arjovsky2017wasserstein} for precipitation nowcasting in Guangdong province, China, and indicates that GAN-based models outperform the traditional ConvGRU, ConvLSTM, and multiscale CNN models.
\texttt{STGM}~\cite{wang2023physical} introduces a task-segmented, synthetic-data generative model (STGM) for heavy rainfall nowcasting by utilizing real-time radar observations in conjunction with physical parameters derived from the Weather Research and Forecasting (WRF) model.
\texttt{PCT-CycleGAN}~\cite{choi2023pct} extends the idea of the cycle-consistent adversarial networks (CycleGAN)~\cite{zhu2017unpaired} and proposes a paired complementary temporal CycleGAN for radar-based precipitation nowcasting.

\paragraph{Diffusion-based models.}
\texttt{LDMRain}~\cite{leinonen2023latent} uses the architecture of latent diffusion model~\cite{rombach2022high} for precipitation nowcasting -- short-term forecasting based on the latest observational data.
Similar works include \texttt{SRNDif}~\cite{ling2024srndiff} and \texttt{GEDRain}~\cite{asperti2023precipitation}.
\texttt{DiffCast}~\cite{yu2024diffcast} models the precipitation process from two perspectives: the deterministic component accounts for predicting a global motion trend by a coarse forecast, while the stochastic component aims to learn local stochastic variations with the residual mechanism. 
\texttt{CasCast}~\cite{gong2024cascast} develops a cascaded framework consisting of a deterministic predictive model to output blurry predictions, and a probabilistic diffusion model with inputs as both past observations and deterministic predictions beforehand. Because the deterministic predictions are the future frames, such frame-wise guidance in the diffusion model can provide a frame-to-frame correspondence between blurry predictions and latent vectors, resulting in a better generation of small-scale patterns. 
However, directly applying diffusion models might generate physically implausible predictions.
To tackle these limitations, \texttt{Prediff}~\cite{gao2023prediff} proposes a conditional latent diffusion model for probabilistic forecasts and then aligns forecasts with domain-specific physical constraints. This is achieved by estimating the deviation from imposed constraints at each denoising step and adjusting the transition distribution accordingly.

\texttt{TimeDiff}~\cite{shen2023non}, \texttt{TimeDDPM}~\cite{dai2023timeddpm}, \texttt{LTD}~\cite{feng2024latent}, \texttt{TimeGrad}~\cite{rasul2021autoregressive}, and \texttt{Dyffusion}~\cite{ruhling2024dyffusion} are examples that have applied diffusion models to general time series modeling, which could be adapted to weather time series. \citet{yang2024survey} provides a comprehensive survey of such methods.

\subsection{Foundation Models}
\label{sec:foundations}
Foundation Models (FMs) have garnered significant research interest due to their powerful prior knowledge acquired through pre-training on massive data and their remarkable adaptability to downstream tasks with fine-tuning strategies~\cite{he2024foundation}. 
While foundation models may be large language models (LLMs), a few foundation models in the weather domain have been proposed.

\texttt{ClimaX}~\cite{nguyen2023climax} is a versatile and generalizable deep-learning model developed for weather and climate science. It is trained on heterogeneous datasets encompassing diverse variables, spatiotemporal coverage, and physical principles with CMIP6 datasets and it can be fine-tuned for a wide range of weather and climate applications, including those involving atmospheric variables and spatiotemporal scales not encountered during pre-training. 
\texttt{W-MAE}~\cite{man2023w} is pre-trained with similar data, but using reconstruction tasks with the Masked Autoencoder model~\cite{he2022masked}. The pre-trained model can be fine-tuned for various tasks, e.g., multi-variate forecasting.
\texttt{Aurora}~\cite{bodnar2024aurora} is a large-scale foundation model pre-trained on over a million hours of diverse weather and climate data. Unlike the two foundation models above, \texttt{Aurora} can be fine-tuned in one of two ways: short-time fine-tuning (i.e., fine-tuning the entire architecture through one or two roll-out steps) and rollout fine-tuning for long-term multi-step predictions with low-rank adaption (LoRA)~\cite{hu2021lora}.
\texttt{Prithvi WxC}~\cite{schmude2024prithvi} is a foundation model with 2.3 billion parameters developed using 160 variables. It is essentially a scalable and flexible 2D vision transformer with varying sizes of tokens or windows. During the pre-training, the Masked Autoencoder model~\cite{he2022masked} is pre-trained by masking different ratios of tokens and windows to capture both regional and global dependencies in the input data.
It can be fine-tuned for nowcasting, forecasting, and downscaling tasks.
More recently, \texttt{AtmosArena}~\cite{nguyen2024atmosarena} benchmarks foundation models for atmospheric sciences across various atmospheric variables.

The large foundation models designed for general time series data, including \texttt{TimeFM}~\cite{das2023decoder}, \texttt{Moment}~\cite{goswami2024moment}, \texttt{Timer}~\cite{liu2024timer}, \texttt{Moirai}~\cite{woo2024unified}, and \texttt{Chronos}~\cite{ansari2024chronos} may be adapted for weather forecasting.

\section{Applications and Resources}
\label{sec:applications}
This section introduces the diverse applications of deep learning models in weather and climate science. We provide an overview of the available datasets, summarized in detail in Table \ref{tab:datasets} in Appendix \ref{sec:datasets}.


\subsection{Precipitation}
Precipitation prediction has witnessed significant advances driven by deep learning (DL) applications, focusing mainly on precipitation nowcasting \cite{Gao2020, gao2021deep, ashok2022systematic, verma2023deep, salcedo2024analysis, an2024deep}.
CNN-based architectures, particularly U-Net, have been widely utilized for their ability to extract local features through convolutional layers, effectively capturing high-dimensional spatio-temporal dynamics of precipitation \cite{lebedev2019precipitation, ayzel2020rainnet, han2021convective, ehsani2022nowcasting, seo2022domain, kim2022region, zhang2023skilful}.
RNN-based models, Transformers, and their hybrid designs combining convolutions represent another dominant approach, optimized for long-term dependency modeling \cite{shi2015convolutional, wang2017predrnn, park2022nowformer, gao2022earthformer, bai2022rainformer, geng2024ms, bodnar2024aurora, zhao2024weathergfm, schmude2024prithvi}. 
Generative models have also played a critical role, with adversarial models (e.g., GANs) \cite{jing2019aenn, liu2020mpl, ravuri2021skilful, wang2023skillful, she2023self, choi2023pct, yin2024precipitation, franch2024gptcast} contributing to precipitation synthesis. Moreover, probabilistic generative diffusion models have gained attention for their superior stability, controllability, and fine-grained synthesis capabilities \cite{leinonen2023latent, gao2023prediff, yu2024diffcast,gong2024cascast}. 


\subsection{Air Quality}
Air quality prediction is of critical importance to society. \citet{zheng2013u} employ artificial neural network (ANN) with spatially-related features to predict the air quality in Beijing, \citet{waseem2022forecasting} employed a CNN-Bi-LSTM architecture for air quality prediction in Xi'an, China, and \citet{yi2018deep} propose a model combining a spatial transformation component and a deep distributed fusion network to predict air quality in nine major cities in China. 
More recently, \citet{shi2022time} evaluate various deep learning models, including RNNs, LSTMs, GRUs, and Transformers, for air quality prediction in Beijing. 
Nationwide air quality forecasting in China has leveraged advanced architectures such as hierarchical group-aware graph neural networks (GAGNN)~\cite{chen2023group}, spatiotemporal graph neural networks (STGNNs)~\cite{wang2020pm2}, and Transformer-based models~\cite{liang2023airformer, yu2025mgsfformer}. 
Additionally, RNNs have been utilized for air quality prediction in India~\cite{arora2022air} and Pakistan~\cite{waseem2022forecasting}, while hybrid CNN-LSTM architectures have been applied for predictions in Barcelona and Turkey~\cite{gilik2022air}.

\subsection{Sea Surface Temperature}
The change in Sea Surface Temperature can cause El Niño/Southern Oscillation (ENSO) and La Niña phenomena, largely impacting the global extreme climate, such as increasing the chances of floods, droughts, heat waves, and cold seasons~\cite{wang2023enso}. 
Niño 3.4 index, an important indicator for ENSO prediction, has been predicted using different deep learning (DL) models, such as RNN-based~\cite{huang2019analyzing, geng2021spatiotemporal}, CNN-based~\cite{ham2019deep,liu2021forecasting}, residual CNNs~\cite{hu2021deep}, ConvLSTM~\cite{he2019dlenso}, GNN-based~\cite{cachay2020graph}, and Transformer-based models~\cite{ye2021transformer,zhou2023self,song2023spatial}. 
More recently, an adaptive graph spatial-temporal attention network (AGSTAN) has been proposed for longer lead (i.e., 23 months) ENSO prediction~\cite{liang2024adaptive}.
\citeauthor{mu2021enso} evaluates multiple DL models for the Niño 3 index, Niño 3.4 index, and Niño 4 index with a multivariate air–sea coupler. Similar evaluation work involves comparing deep learning models for ENSO forecasting and presenting ENSO dataset~\cite{mir2024enso}.
Moreover, some researchers directly predict the sea surface temperature using spatiotemporal graph attention networks~\cite{gao2023global} and physical knowledge-enhanced generative adversarial networks~\cite{meng2023physical}. 
ENSO impacts have also been studied, including river flows~\cite{liu2023explainable}, rainfall~\cite{he2024data}, and heatwaves~\cite{he2024interpretable}.

\subsection{Flood}
Accurate flood prediction is essential for mitigating the adverse impacts of flooding. 
Recent advances in deep learning (DL) have led to the development of various models tailored for flood forecasting and mapping, such as CNN-based~\cite{adikari2021evaluation}, RNN-based and LSTM~\cite{nevo2022flood, ruma2023particle}, and CNN-RNN hybrid models such as ConvLSTM ~\cite{li2022prediction, moishin2021designing}, and LSTM-DeepLabv3+~\cite{situ2024improving}. 
\citet{situ2024attention} employs the \emph{attention} mechanism for urban flood damage and risk assessment with improved flood prediction and land use segmentation.
Furthermore, graph-based models have also gained attention for flood prediction~\cite{kirschstein2024merit}. FloodGNN-GRU combines GNNs and Gated Recurrent Units (GRUs) for spatiotemporal flood prediction by incorporating vector features like velocities~\cite{kazadi2024floodgnn} while Graph Transformer Network (FloodGTN) integrates GNNs and Transformers to learn spatiotemporal dependencies in water levels~\cite{shi2023graph,shi2024fidlar}. 
Additionally, physics-guided models further enhance flood prediction by embedding physical laws into model training. 
For instance, the DK-Diffusion model incorporates flood physics into its loss function to align predictions with hydrological principles~\cite{shao2024data}. 
\texttt{DRUM} leverages diffusion model for operational flood forecasting and long-term risk assessment~\cite{ou2024drum}.
Moreover, conditional GANs have been explored for flood predictions across untrained catchments~\cite{do2023generalizing}, demonstrating their versatility in diverse hydrological conditions.

\subsection{Drought}
Drought, driven by a complex interplay of meteorological, agricultural, hydrological, and socio-economic factors, manifests across diverse spatial and temporal scales~\cite{wilhite2016drought,gyaneshwar2023contemporary}. 
We focus on DL methods that consider meteorological drivers, such as precipitation deficits, wind patterns, and temperature anomalies, to predict various drought indices. 
LSTMs have been widely used to predict spatial precipitation patterns (dry-wet)~\cite{gibson2021training} and drought indices related to precipitation, such as the standardized precipitation index (SPI)~\cite{poornima2019drought, dikshit2021explainable} and the standardized precipitation evapotranspiration index (SPEI)~\cite{tian2021drought, dikshit2021long, xu2022application}, excelling at capturing long-term dependencies. 
Beyond SPI and SPEI~\cite{adikari2021evaluation, dhyani2021deep, hao2023forecasting}, CNNs have been applied for predicting other indices, such as the soil moisture index (SMI)~\cite{dhyani2021deep} and soil moisture condition index (SMCI)~\cite{zhang2024multivariate}, aiding agricultural drought prediction. 
Hybrid models like ConvLSTM and CNN-LSTM have demonstrated significant improvements in multi-temporal predictions for SPEI~\cite{danandeh2023novel, nyamane2024harnessing} and SPI~\cite{park2020short}, as well as indices like the scaled drought condition index (SDCI)~\cite{park2020short}, composite drought index (CDI)~\cite{zhang2023construction}, and Palmer drought severity index (PDSI)~\cite{elbeltagi2024advanced}. 
Specifically, the CNN-GRU model has effectively forecasted daily reference evapotranspiration (ET)~\cite{ahmed2022hybrid}. 
Swin Transformer was used for drought prediction across multiple scales ~\cite{zhang2024multiscale}.
Meanwhile, GANs have emerged as robust tools for drought prediction, with applications spanning vegetative drought prediction~\cite{shukla2023deep}, and SMI~\cite{ferchichi2024spatio}.

\subsection{Tropical Storms/Cyclones and Hurricanes}
Accurate forecasting of tropical storms, cyclones, and hurricanes is crucial for mitigating their devastating impacts.
CNN-based models have been increasingly employed to predict various aspects of these phenomena, focusing on targets such as storm formation~\cite{zhang2021predicting, nguyen2024predicting}, intensity~\cite{kim2024improvement}, track~\cite{giffard2020tropical, lian2020novel}, and associated rainfall~\cite{kim2022near}.
Hybrid models, such as CNN-LSTM, further improve the accuracy of intensity prediction~\cite{alijoyo2024advanced}, extend lead times up to 60 hours~\cite{kumar2022forecastingformationtropicalcyclone}, and effectively capture landfall in terms of location and time~\cite{kumar2021predicting}.
GANs have also proven valuable in downscaling tropical cyclone rainfall to hazard-relevant spatial scales~\cite{vosper2023deep} and in multitask frameworks for simultaneously forecasting cyclone paths and intensities~\cite{wu2021tropical}. 
Recent approaches like diffusion models have been explored for forecasting cyclone trajectories and precipitation patterns~\cite{nath2023forecasting}. GNNs integrated with GRUs have been utilized to model storm surge dependencies across observation stations, offering improvements in spatial and temporal forecasting~\cite{jiang2024advancing}.

\subsection{Wildfire}
Accurate wildfire prediction is critical for disaster management and mitigation. 
CNN-based models have demonstrated strong capabilities in wildfire spread prediction~\cite{khennou2021forest, shadrin2024wildfire}, including forecasting fire weather with high spatial resolution~\cite{son2022deep}, generating spread maps~\cite{huot2022next}, and modeling large-scale fire dynamics using multi-kernel architectures~\cite{marjani2023large}. 
RNNs, including GRUs and LSTMs, excel in modeling wildfire risk and predicting spread, with GRU-LSTM showing superior performance in longer time series data~\cite{perumal2020comparison, dzulhijjah2023comparative, gopu2023comparative}. 
Hybrid CNN-LSTM models further enhance prediction accuracy, offering near-real-time daily wildfire spread forecasting~\cite{marjani2024cnn} and incorporating multi-temporal dynamics for prediction~\cite{marjani2023firepred}. 
ConvLSTM models capture a wide range of temporal scales in wildfire prediction, from short-term intervals of 15 minutes~\cite{burge2023recurrent} to longer-term forecasts extending up to 10 days~\cite{masrur2023spatiotemporal, masrur2024capturing}.
Other advancements include GANs, which have been utilized for wildfire risk prediction through conditional tabular data augmentation~\cite{chowdhury2021mitigating}, and GNNs, which simulate wildfire spread in variable-scale landscapes, effectively addressing landscape heterogeneity~\cite{jiang2022modeling}. Additionally, researchers have also explored Transformer models for wildfire prediction~\cite{miao2023time, cao2024forest}.

\section{Challenges and Future Directions}
\label{sec:outlook}
In this section, we introduce primary challenges and suggest promising future research opportunities from the perspectives of DL models (Subsections \ref{sec:trustAI}-\ref{sec:mml}) and data (Subsections \ref{sec:mml}-\ref{sec:data}).

\subsection{Trustworthy AI}
\label{sec:trustAI}
We discuss trustworthy AI models paying careful attention to robustness, generalization, explainability, scalability, and uncertainty quantification.

\paragraph{Robustness:} 
Weather data is often subject to observational or collection biases, leading to significant performance degradation in AI models. These biases may stem from inconsistent data collection methods, non-uniformity or limited spatial or temporal coverage, and inaccuracies in sensor measurements. As a result, AI models trained on such biased data sets may struggle to generalize effectively.
\grayshade{\texttt{Opportunities:}} 
(1) Bias correction with statistical adjustments~\cite{durai2014evaluation} and data assimilation~\cite{berry2017correcting} can be applied to reduce biases in the data.
(2) Adversarial training~\cite{wang2024evaluating}, a technique originally developed to defend against adversarial attacks in machine learning, can mitigate vulnerabilities by exposing models to challenging or perturbed examples during training, allowing them to generalize better to real-world biases or anomalies. It involves creating perturbed versions of weather data representing scenarios with systematic biases and incorporating adversarial examples alongside clean data during training to improve its robustness to biased data sets~\cite{schmalfuss2023distracting}.

\paragraph{Generalization:}
AI models often fail to perform effectively on rare extreme weather or anomalous events that fall outside the distribution (OOD) of the training samples.
\grayshade{\texttt{Opportunities:}} 
(1) Physical laws represent precious wisdom from domain pioneers, but they are rarely explicitly incorporated into AI models~\cite{feng2023physics}. Leveraging physics-informed or physics-guided AI approaches can increase reliability and consistency with the physical world~\cite{chen2021physics,meng2021physics,yin2023physic}, particularly while addressing extreme or unseen scenarios. 
Although significant progress has been made in the integration of physics and AI (see ``Physics-AI'' in Section \ref{sec:overview}), further exploration is needed to optimize and refine these approaches.
(2) DL models perform poorly in extreme weather events due to their rarity and limited representation in the training data. Effective data augmentation with generative diffusion models~\cite{trabucco2023effective, mardani2023generative} is a promising method to address or alleviate this challenge. By augmenting the training set with more extreme samples, DL models are better equipped to understand these rare events comprehensively, enhancing their generalizability. Therefore, it is worth exploring how to effectively augment data with extreme samples.

\paragraph{Explainablity:} 
Neural networks are frequently referred to as ``black boxes'' due to the opacity of their internal processes, making it challenging to interpret how they produce outputs~\cite{guidotti2018survey}. In the weather and climate domains, understanding the underlying mechanisms of these models is of paramount importance and a necessity to ensure reliability and trustworthiness.
\grayshade{\texttt{Opportunities:}} 
Explainable AI tools, such as SHAP (Shapley Additive Explanations)~\cite{lundberg2017unified}, LIME (Local Interpretable Model-Agnostic Explanations)~\cite{ribeiro2016should}, Grad-CAM~\cite{selvaraju2017grad}, and causal analysis~\cite{zhang2011causality} have gained prominence in addressing this challenge. 
Furthermore, the principle of information bottleneck (IB) has been used for explainable learning in the time series domain~\cite{feng2024timesieve,liu2024timex}. Given that weather data inherently constitute time series, we advocate exploring how the information bottleneck method can enhance the explainability of weather modeling.
Leveraging these techniques can help determine whether DL models are producing meaningful results based on legitimate patterns or merely fabricating outputs, reinforcing trustworthiness and accountability in model predictions.

\paragraph{Varying Resolution:}
In weather and climate science, is it common to deal with varying data resolutions.
For example, weather data have differing temporal and spatial resolutions across modalities. Meteorological observations might have an hourly temporal resolution from sparse sensors, radar echo data could feature six-minute temporal intervals and a spatial resolution of 1–4 km, and satellite imagery might exhibit a temporal resolution of 30 minutes with a spatial resolution of 5–12 km. 
These discrepancies complicate the task of harmonizing information across modalities for robust model development~\cite{chen2023foundation}.
\grayshade{\texttt{Opportunities:}}
Therefore, an important challenge is to build models that can handle training data of varying resolutions and also reliably predict at a different resolution. Such models could revolutionize how we integrate data from various sources, including observations, satellite imagery, and numerical simulations, which often differ in granularity and format.
\texttt{Aurora} processes input data with varying patch sizes~\cite{bodnar2024aurora}, and \texttt{IPOT} (Inducing-point operator transformer) uses a smaller number of inducing points,  flexibly handling any discretization formats of input~\cite{lee2024inducing}.

\paragraph{Uncertainty Quantification:} Given the chaotic nature of the atmosphere, quantifying uncertainty in weather predictions is essential to allow informed decision-making. Approaches such as initial conditions perturbation and Monte Carlo dropout have been studied~\cite{bulte2024uncertainty}; however, they only simulate the aleatoric uncertainty, i.e., the inherent randomness in from weather data or the epistemic uncertainty from the model itself due to the limited knowledge.
\grayshade{\texttt{Opportunities:}} Generative diffusion models address both aleatoric and epistemic uncertainty simultaneously.
Diffusion models learn the full probability distribution of the data, capturing aleatoric uncertainty through stochastic sampling, where the spread of outcomes reflects inherent data variability. 
When conditioned on the inputs, added stochastic noise incorporates input variability, further representing data-driven uncertainty. Additionally, by initializing from different noise points, diffusion models capture epistemic uncertainty~\cite{du2023diffusion,price2023gencast}, with greater variability in regions of sparse training data. 
This inherent stochasticity makes diffusion models a robust tool for quantifying both aleatoric and epistemic uncertainties.

\subsection{Retrieval-augmented Foundation Models}
\label{sec:retrieval}
Retrieval-augmented generation (RAG)~\cite{gao2023retrieval} has emerged as a promising approach to enhance foundation models by integrating external domain knowledge.
\grayshade{\texttt{Opportunities:}} While RAG has been extensively explored in domains such as medicine~\cite{xiong2024benchmarking}, its application to weather and climate modeling remains underexplored. Depending on whether the foundation model uses diffusion models~\cite{yang2023diffusion} or large language models (LLMs)~\cite{zhao2023survey} as its underlying architecture, different opportunities arise for leveraging retrieval augmentation:
(1) Diffusion Models for Weather Forecasting: In the context of diffusion-based weather models~\cite{shi2024codicast}, retrieval augmentation can be leveraged to fetch historical weather patterns similar to the current state, allowing it to recreate historical conditions that may have appeared in the past and that can serve as references to refine predictions, potentially improving accuracy and robustness~\cite{liu2024retrieval,ravuru2024agentic}. 
It holds significant potential to enhance performance in extreme weather scenarios by addressing the challenges posed by data rarity.
(2) LLMs for Weather Text Analysis: For tasks involving textual analysis of weather-related corpora, such as extreme weather reports or climatological summaries~\cite{colverd2023floodbrain}, retrieval augmentation can provide valuable context by identifying and incorporating relevant documents. This approach can significantly enhance the model's ability to generate informed and contextually relevant outputs~\cite{juhasz2024responsible}. 
By bridging retrieval-based methodologies with foundation models, RAG helps to maximize the power of foundation models, presenting an exciting avenue for advancing both accuracy and interpretability in weather and climate applications.

\subsection{Generative AI with Weather Constraints}
\label{sec:GenAI}
Generative models have achieved enormous success in image generation~\cite{goodfellow2014generative, ho2020denoising}. More interestingly, controllable generative models can synthesize customized images according to conditions provided by users~\cite{gauthier2014conditional, rombach2022high}. 
\grayshade{\texttt{Opportunities:}} In the weather domain, weather \emph{prediction} can be formulated as weather \emph{generation} conditioned on temporal and spatial similarities. These conditions or constraints could come from (1) partial differential continuity equations~\cite{broome2014pde,palmer2019stochastic}, which describe the weather as a flux, a spatial movement of quantities over time; (2) Tobler's First law of Geography~\cite{tobler2004first}, which states that everything is related to everything else, but near things are more related than distant things; and (3) Tobler's Second law of Geography~\cite{tobler1999linear}, which states that the phenomenon external to a geographic area of interest affects what goes on inside; and (4) other modalities, such as station-based, satellite-based~\cite{qu2024deep,xiang2024adaf}, and even text data~\cite{li2024cllmate}. 
By leveraging the weather constraints as prior knowledge, these models could learn more robust and precise representations from the complex weather data.
Besides, accelerating training and inference is important~\cite{song2020denoising} since diffusion models often incur high computational overheads.

\subsection{Multi-Modal Learning}
\label{sec:mml}
Weather data comes from heterogeneous sources, encompassing observational data (e.g., sensors, radar, satellite imagery), reanalysis data, and supplementary text descriptions~\cite{li2024cllmate}. 
\grayshade{\texttt{Opportunities:}} These modalities can complement each other, offering a more comprehensive understanding of weather and climate phenomena. Therefore, a promising direction is to leverage such multi-modal data to learn joint representations of weather and climate events. However, a key challenge lies in effectively ``aligning'' these multi-modal data.
Mapping numerical data to textual descriptions presents an additional layer of complexity. One possibility involves leveraging large language models (LLMs) to construct knowledge graphs that extract information about weather and climate events from corpora of environment-focused news articles.
These extracted events can then be linked with meteorological raster data to enrich the model's understanding and predictive capabilities~\cite{li2024cllmate}.

\subsection{Data Processing and Management}
\label{sec:data}
%
\paragraph{Data Storage:} The volume of weather and climate data is increasing daily - European Centre for Medium-Range Weather Forecasts (ECMWF) archives contain about 450 PB of data to which 300 TB are added daily~\cite{mukkavilli2023ai}.
\grayshade{\texttt{Opportunities:}} Variational Autoencoder (VAE) approaches have emerged as powerful tools for data compression~\cite{liu2024compressing,han2024cra5}, converting the high-dimensional data from the original space to a lower latent space. \citeauthor{liu2024compressing} reduce the data size from 8.61 TB to a compact 204 GB and \citeauthor{han2024cra5} compress the ERA5 dataset (226 TB) into a CRA5 dataset (0.7 TB). More importantly, they demonstrate that downstream experiments of global weather forecasting models trained on the compact CRA5 dataset achieve accuracy comparable to the models trained on the original dataset. This approach significantly reduces storage requirements for massive weather datasets.

\paragraph{Data Quality:} Massive gridded reanalysis data are computed using mechanical or statistics models, which are still based on empirical assumptions. Thus, the quality of the reanalysis data is of concern. 
\grayshade{\texttt{Opportunities:}} Data assimilation~\cite{manshausen2024generative} is a promising method to increase data quality by calibrating model outputs with observational data, which could be remote sensing imagery and ground station measurements. For example, \texttt{SLAMS} proposes a conditional diffusion model to assimilate \emph{in situ} weather station data and \emph{ex situ} satellite imagery to effectively calibrate the vertical temperature profiles~\cite{qu2024deep}, \texttt{ADAF} employs Swin Transformer to achieve effective data assimilation using real-world observations from different locations and multiple sources, including sparse surface weather observations and satellite imagery~\cite{xiang2024adaf}. 
Furthermore, \texttt{EarthNet} ia a multi-modal foundation model for global data assimilation of Earth observations utilizing masked autoencoders~\cite{vandal2024global}.
In summary, DL methods have become increasingly popular for integrating weather data from various sources to provide more precise representations.

\section{Discussion}
\label{sec:disc}
We have introduced three categories of models in Section \ref{sec:overview}. Each approach offers unique strengths and trade-offs, making them suitable for different scenarios depending on the nature of the task, data availability, and computational resources.
Below, we provide a detailed comparison and analysis of what works best in different scenarios, exploring why certain models excel in specific contexts.

\paragraph{Deterministic Predictive Models.} These models have demonstrated exceptional performance for short-, medium- and long-range weather predictions. While Transformer-based models work well on temporal predictions, GNN-based models excel at modeling spatial relations, and hybrid models capture spatiotemporal dependencies with greater accuracies, but may require a longer time for training.
\texttt{WeatherBench 2}~\cite{rasp2023weatherbench} has benchmarked data-driven global medium-range (10 days) weather models and provides a detailed headline scorecard\footnote{https://sites.research.google/weatherbench/}. 
In summary, \texttt{NeuralGCM} outperforms other state-of-the-art DL models, and it is comparable with the physics-based ECMWF's \texttt{IFS} regarding geopotential, temperature, and wind variables. Models like \texttt{GraphCast}, \texttt{Pangu}, and \texttt{Fuxi} have shown competitive or better performance compared with ECMWF’s High-RESolution forecast (\texttt{HRES}).
However, three challenges remain. 
1) Their output is usually blurry because they are typically trained to minimize a deterministic loss function that uses mean squared error (MSE). This becomes worse for extreme weather events.
2) They lack aleatoric and epistemic uncertainty quantification. Even though there have been attempts to use traditional initial condition perturbation methods to produce ensemble forecasts, modeling the uncertainty of weather evolution has not been addressed.
3) These models need architectural changes and re-training when applied to other specific tasks.

\paragraph{Probabilistic generative models.} 
These models have shown great promise for accurate weather prediction.
More importantly, probabilistic generative models such as \texttt{GenCast}, \texttt{CoDiCast}, and \texttt{CasCast} (see Figure \ref{fig:taxonomy}) have brought unique strengths by modeling aleatoric and epistemic uncertainty due to the probabilistic noise sampling.
These are particularly valuable for predicting extreme weather events, where probabilistic outputs can facilitate informed decision-making. 
\texttt{GenCast} has reported greater skill than \texttt{IFS ENS} on 97.4\% of 1320 targets they evaluated.
However, these models require more computational resources for training and inference than deterministic predictive models, though they are faster than physics-based models.

\paragraph{Foundation models.}
Foundation models like \texttt{Aurora}, \texttt{ClimaX} and \texttt{Prithvi WxC} represent a significant leap in adaptability and transfer learning, offering robust performance across diverse tasks after fine-tuning. 
Furthermore, current foundation models are primarily based on deterministic predictive learning for pre-training, where latent embeddings are often obtained with predictive learning.
We have not identified any that utilize probabilistic generative architectures.
However, their large parameter size and pre-training requirements can create barriers for research groups with limited computational resources. 
Furthermore, fine-tuning techniques in weather forecasting are still in their early stages and could benefit from insights and advancements in the natural language processing domain~\cite{zheng2023learn,sun2022recent}.

\begin{table}[ht]
\centering
\caption{Comparison of Predictive Learning, Generative Learning, and Pre-training \& Fine-tuning Models for global medium-range (10 days) weather prediction.}
\vspace{-2mm}
\resizebox{.999\columnwidth}{!}{
    \begin{tabular}{l|l|l|l}
    \toprule
    \textbf{ }        & \textbf{Predictive}             & \textbf{Generative}       & \textbf{Pre-training \&} \\
    \textbf{ }        & \textbf{Learning}               & \textbf{Learning}         & \textbf{Fine-tuning}  \\
    \midrule
    \multirow{2}{*}{Accuracy}   & \texttt{NeuralGCM} and \texttt{FuXi} are  & \texttt{GenCast}: 97.4\% targets & \texttt{Aurora} vastly   \\ 
                                & comparable with \texttt{IFS ENS}         & better than \texttt{IFS ENS}     & better than \texttt{IFS HERS} \\ 
    \midrule
    \multirow{2}{*}{Efficiency} & Fast training;        & Slow training;            & Slow training;  \\
                      & Fast inference                  & Slow inference            & Fast inference   \\
    \midrule
    Uncertainty       & Need perturbation               & Inherent                & -                \\ 
    \midrule
    Adaptability      & Need re-training                & Need re-training          & Fine-tuning       \\ 
    \bottomrule
    \end{tabular}
}
\label{tab:learning_models_comparison}
\end{table}

\section{Conclusions}
In this work, we present a comprehensive and up-to-date survey of data-driven deep learning models and foundation models for weather prediction. 
We introduce a novel categorization of these models based on their training paradigms and provide an in-depth review, analysis, and comparison of key methodologies within each category. 
Additionally, we summarize available datasets, open-source codebases, and diverse real-world applications in a GitHub repository. 
More importantly, we outline ten critical research directions across five primary avenues for advancing AI-driven weather prediction, offering a roadmap for future research. 

\paragraph{Limitations.} In this survey, we are particularly targeting the topic of weather prediction. 
Due to the limited space, other research topics in weather and climate domains are out of the scope of this survey, including climate downscaling~\cite{ling2024diffusion}, climate emulation~\cite{yu2024climsim}, and climate trend prediction~\cite{cael2023global}.

\section*{Acknowledgments}
This work is supported by the Institute for Geospatial Understanding through an Integrative Discovery Environment (I-GUIDE), which is funded by the National Science Foundation (NSF) under award number: 2118329. Any opinions, findings, conclusions, or recommendations expressed herein are those of the authors and do not necessarily represent the views of NSF.

\bibliography{references}

\begin{thebibliography}{340}
\providecommand{\natexlab}[1]{#1}

\bibitem[{Abbass et~al.(2022)Abbass, Qasim, Song, Murshed, Mahmood, and Younis}]{abbass2022review}
Kashif Abbass, Muhammad~Zeeshan Qasim, Huaming Song, Muntasir Murshed, Haider Mahmood, and Ijaz Younis. 2022.
\newblock A review of the global climate change impacts, adaptation, and sustainable mitigation measures.
\newblock \emph{Environmental Science and Pollution Research}, 29(28):42539--42559.

\bibitem[{Adikari et~al.(2021)Adikari, Shrestha, Ratnayake, Budhathoki, Mohanasundaram, and Dailey}]{adikari2021evaluation}
Kasuni~E Adikari, Sangam Shrestha, Dhanika~T Ratnayake, Aakanchya Budhathoki, S~Mohanasundaram, and Matthew~N Dailey. 2021.
\newblock Evaluation of artificial intelligence models for flood and drought forecasting in arid and tropical regions.
\newblock \emph{Environmental Modelling \& Software}, 144:105136.

\bibitem[{Ahmed et~al.(2022)Ahmed, Deo, Feng, Ghahramani, Raj, Yin, and Yang}]{ahmed2022hybrid}
AA~Masrur Ahmed, Ravinesh~C Deo, Qi~Feng, Afshin Ghahramani, Nawin Raj, Zhenliang Yin, and Linshan Yang. 2022.
\newblock Hybrid deep learning method for a week-ahead evapotranspiration forecasting.
\newblock \emph{Stochastic Environmental Research and Risk Assessment}, pages 1--19.

\bibitem[{Alexe et~al.(2024)Alexe, Boucher, Lean, Pinnington, Laloyaux, McNally, Lang, Chantry, Burrows, Chrust et~al.}]{alexe2024graphdop}
Mihai Alexe, Eulalie Boucher, Peter Lean, Ewan Pinnington, Patrick Laloyaux, Anthony McNally, Simon Lang, Matthew Chantry, Chris Burrows, Marcin Chrust, et~al. 2024.
\newblock Graphdop: Towards skilful data-driven medium-range weather forecasts learnt and initialised directly from observations.
\newblock \emph{arXiv preprint arXiv:2412.15687}.

\bibitem[{Alijoyo et~al.(2024)Alijoyo, Gongada, Kaur, Mageswari, Sekhar, Ramesh, El-Ebiary, and Ulmas}]{alijoyo2024advanced}
Franciskus~Antonius Alijoyo, Taviti~Naidu Gongada, Chamandeep Kaur, N~Mageswari, JC~Sekhar, Janjhyam Venkata~Naga Ramesh, Yousef A~Baker El-Ebiary, and Zoirov Ulmas. 2024.
\newblock Advanced hybrid cnn-bi-lstm model augmented with ga and ffo for enhanced cyclone intensity forecasting.
\newblock \emph{Alexandria Engineering Journal}, 92:346--357.

\bibitem[{An et~al.(2024)An, Oh, Sohn, and Kim}]{an2024deep}
Sojung An, Tae-Jin Oh, Eunha Sohn, and Donghyun Kim. 2024.
\newblock Deep learning for precipitation nowcasting: A survey from the perspective of time series forecasting.
\newblock \emph{arXiv preprint arXiv:2406.04867}.

\bibitem[{Andrae et~al.(2024)Andrae, Landelius, Oskarsson, and Lindsten}]{andrae2024continuous}
Martin Andrae, Tomas Landelius, Joel Oskarsson, and Fredrik Lindsten. 2024.
\newblock Continuous ensemble weather forecasting with diffusion models.
\newblock \emph{arXiv preprint arXiv:2410.05431}.

\bibitem[{Andrychowicz et~al.(2023)Andrychowicz, Espeholt, Li, Merchant, Merose, Zyda, Agrawal, and Kalchbrenner}]{andrychowicz2023deep}
Marcin Andrychowicz, Lasse Espeholt, Di~Li, Samier Merchant, Alexander Merose, Fred Zyda, Shreya Agrawal, and Nal Kalchbrenner. 2023.
\newblock \href {https://arxiv.org/abs/2306.06079} {Deep learning for day forecasts from sparse observations}.
\newblock \emph{Preprint}, arXiv:2306.06079.

\bibitem[{Ansari et~al.(2024)Ansari, Stella, Turkmen, Zhang, Mercado, Shen, Shchur, Rangapuram, Arango, Kapoor et~al.}]{ansari2024chronos}
Abdul~Fatir Ansari, Lorenzo Stella, Caner Turkmen, Xiyuan Zhang, Pedro Mercado, Huibin Shen, Oleksandr Shchur, Syama~Sundar Rangapuram, Sebastian~Pineda Arango, Shubham Kapoor, et~al. 2024.
\newblock Chronos: Learning the language of time series.
\newblock \emph{arXiv preprint arXiv:2403.07815}.

\bibitem[{Arjovsky et~al.(2017)Arjovsky, Chintala, and Bottou}]{arjovsky2017wasserstein}
Martin Arjovsky, Soumith Chintala, and L{\'e}on Bottou. 2017.
\newblock Wasserstein generative adversarial networks.
\newblock In \emph{International conference on machine learning}, pages 214--223. PMLR.

\bibitem[{Arora et~al.(2022)Arora, Sawaran~Singh, Singh, Rakesh~Shrivastava, Mathur, Tiwari, and Agarwal}]{arora2022air}
Sugandha Arora, Narinderjit~Singh Sawaran~Singh, Divyanshu Singh, Rishi Rakesh~Shrivastava, Trilok Mathur, Kamlesh Tiwari, and Shivi Agarwal. 2022.
\newblock Air quality prediction using the fractional gradient-based recurrent neural network.
\newblock \emph{Computational Intelligence and Neuroscience}, 2022(1):9755422.

\bibitem[{Ashok and Pekkat(2022)}]{ashok2022systematic}
Shejule~Priya Ashok and Sreeja Pekkat. 2022.
\newblock A systematic quantitative review on the performance of some of the recent short-term rainfall forecasting techniques.
\newblock \emph{Journal of Water and Climate Change}, 13(8):3004--3029.

\bibitem[{Asperti et~al.(2023{\natexlab{a}})Asperti, Merizzi, Paparella, Pedrazzi, Angelinelli, and Colamonaco}]{asperti2308precipitation}
A~Asperti, F~Merizzi, A~Paparella, G~Pedrazzi, M~Angelinelli, and S~Colamonaco. 2023{\natexlab{a}}.
\newblock Precipitation nowcasting with generative diffusion models.
\newblock \emph{arXiv preprint arXiv:2308.06733}.

\bibitem[{Asperti et~al.(2023{\natexlab{b}})Asperti, Merizzi, Paparella, Pedrazzi, Angelinelli, and Colamonaco}]{asperti2023precipitation}
Andrea Asperti, Fabio Merizzi, Alberto Paparella, Giorgio Pedrazzi, Matteo Angelinelli, and Stefano Colamonaco. 2023{\natexlab{b}}.
\newblock \href {https://arxiv.org/abs/2308.06733} {Precipitation nowcasting with generative diffusion models}.
\newblock \emph{Preprint}, arXiv:2308.06733.

\bibitem[{Ayzel et~al.(2020{\natexlab{a}})Ayzel, Scheffer, and Heistermann}]{gmd-13-2631-2020}
G.~Ayzel, T.~Scheffer, and M.~Heistermann. 2020{\natexlab{a}}.
\newblock \href {https://doi.org/10.5194/gmd-13-2631-2020} {Rainnet v1.0: a~convolutional neural network for radar-based precipitation nowcasting}.
\newblock \emph{Geoscientific Model Development}, 13(6):2631--2644.

\bibitem[{Ayzel et~al.(2020{\natexlab{b}})Ayzel, Scheffer, and Heistermann}]{ayzel2020rainnet}
Georgy Ayzel, Tobias Scheffer, and Maik Heistermann. 2020{\natexlab{b}}.
\newblock Rainnet v1. 0: a convolutional neural network for radar-based precipitation nowcasting.
\newblock \emph{Geoscientific Model Development}, 13(6):2631--2644.

\bibitem[{Bai et~al.(2022)Bai, Sun, Zhang, Song, and Chen}]{bai2022rainformer}
Cong Bai, Feng Sun, Jinglin Zhang, Yi~Song, and Shengyong Chen. 2022.
\newblock Rainformer: Features extraction balanced network for radar-based precipitation nowcasting.
\newblock \emph{IEEE Geoscience and Remote Sensing Letters}, 19:1--5.

\bibitem[{Berry and Harlim(2017)}]{berry2017correcting}
Tyrus Berry and John Harlim. 2017.
\newblock Correcting biased observation model error in data assimilation.
\newblock \emph{Monthly Weather Review}, 145(7):2833--2853.

\bibitem[{Bi et~al.(2023)Bi, Xie, Zhang, Chen, Gu, and Tian}]{bi2023accurate}
Kaifeng Bi, Lingxi Xie, Hengheng Zhang, Xin Chen, Xiaotao Gu, and Qi~Tian. 2023.
\newblock Accurate medium-range global weather forecasting with 3d neural networks.
\newblock \emph{Nature}, 619(7970):533--538.

\bibitem[{Bodnar et~al.(2024)Bodnar, Bruinsma, Lucic, Stanley, Brandstetter, Garvan, Riechert, Weyn, Dong, Vaughan et~al.}]{bodnar2024aurora}
Cristian Bodnar, Wessel~P Bruinsma, Ana Lucic, Megan Stanley, Johannes Brandstetter, Patrick Garvan, Maik Riechert, Jonathan Weyn, Haiyu Dong, Anna Vaughan, et~al. 2024.
\newblock Aurora: A foundation model of the atmosphere.
\newblock \emph{arXiv preprint arXiv:2405.13063}.

\bibitem[{Bojesomo et~al.(2021)Bojesomo, Al-Marzouqi, and Liatsis}]{bojesomo2021spatiotemporal}
Alabi Bojesomo, Hasan Al-Marzouqi, and Panos Liatsis. 2021.
\newblock Spatiotemporal vision transformer for short time weather forecasting.
\newblock In \emph{2021 IEEE International Conference on Big Data (Big Data)}, pages 5741--5746. IEEE.

\bibitem[{Box et~al.(2015)Box, Jenkins, Reinsel, and Ljung}]{box2015time}
George~EP Box, Gwilym~M Jenkins, Gregory~C Reinsel, and Greta~M Ljung. 2015.
\newblock \emph{Time series analysis: forecasting and control}.
\newblock John Wiley \& Sons.

\bibitem[{Broom{\'e} and Ridenour(2014)}]{broome2014pde}
Sofia Broom{\'e} and Jonathan Ridenour. 2014.
\newblock A pde perspective on climate modeling.

\bibitem[{B{\"u}lte et~al.(2024)B{\"u}lte, Horat, Quinting, and Lerch}]{bulte2024uncertainty}
Christopher B{\"u}lte, Nina Horat, Julian Quinting, and Sebastian Lerch. 2024.
\newblock Uncertainty quantification for data-driven weather models.
\newblock \emph{arXiv preprint arXiv:2403.13458}.

\bibitem[{Burge et~al.(2023)Burge, Bonanni, Hu, and Ihme}]{burge2023recurrent}
John Burge, Matthew~R Bonanni, R~Lily Hu, and Matthias Ihme. 2023.
\newblock Recurrent convolutional deep neural networks for modeling time-resolved wildfire spread behavior.
\newblock \emph{Fire Technology}, 59(6):3327--3354.

\bibitem[{Cachay et~al.(2020)Cachay, Erickson, Bucker, Pokropek, Potosnak, Osei, and L{\"u}tjens}]{cachay2020graph}
Salva~R{\"u}hling Cachay, Emma Erickson, Arthur Fender~C Bucker, Ernest Pokropek, Willa Potosnak, Salomey Osei, and Bj{\"o}rn L{\"u}tjens. 2020.
\newblock Graph neural networks for improved el ni$\backslash$\~{} no forecasting.
\newblock \emph{arXiv preprint arXiv:2012.01598}.

\bibitem[{Cael et~al.(2023)Cael, Bisson, Boss, Dutkiewicz, and Henson}]{cael2023global}
BB~Cael, Kelsey Bisson, Emmanuel Boss, Stephanie Dutkiewicz, and Stephanie Henson. 2023.
\newblock Global climate-change trends detected in indicators of ocean ecology.
\newblock \emph{Nature}, 619(7970):551--554.

\bibitem[{Cao et~al.(2024{\natexlab{a}})Cao, Zhang, Lall, Holsclaw, and Shao}]{cao2024predictability}
Qing Cao, Hanchen Zhang, Upmanu Lall, Tracy Holsclaw, and Quanxi Shao. 2024{\natexlab{a}}.
\newblock The predictability of daily rainfall during rainy season over east asia by a bayesian nonhomogeneous hidden markov model.
\newblock \emph{Journal of Flood Risk Management}, 17(1):e12942.

\bibitem[{Cao et~al.(2024{\natexlab{b}})Cao, Zhou, Yu, Rao, Wu, Li, and Zhu}]{cao2024forest}
Yue Cao, Xuanyu Zhou, Yanqi Yu, Shuyu Rao, Yihui Wu, Chunpeng Li, and Zhengli Zhu. 2024{\natexlab{b}}.
\newblock Forest fire prediction based on time series networks and remote sensing images.
\newblock \emph{Forests}, 15(7):1221.

\bibitem[{Carleton and Hsiang(2016)}]{carleton2016social}
Tamma~A Carleton and Solomon~M Hsiang. 2016.
\newblock Social and economic impacts of climate.
\newblock \emph{Science}, 353(6304):aad9837.

\bibitem[{Chattopadhyay et~al.(2022)Chattopadhyay, Mustafa, Hassanzadeh, Bach, and Kashinath}]{chattopadhyay2022towards}
Ashesh Chattopadhyay, Mustafa Mustafa, Pedram Hassanzadeh, Eviatar Bach, and Karthik Kashinath. 2022.
\newblock Towards physics-inspired data-driven weather forecasting: integrating data assimilation with a deep spatial-transformer-based u-net in a case study with era5.
\newblock \emph{Geoscientific Model Development}, 15(5):2221--2237.

\bibitem[{Chen et~al.(2023{\natexlab{a}})Chen, Han, Gong, Bai, Ling, Luo, Chen, Ma, Zhang, Su et~al.}]{chen2023fengwu}
Kang Chen, Tao Han, Junchao Gong, Lei Bai, Fenghua Ling, Jing-Jia Luo, Xi~Chen, Leiming Ma, Tianning Zhang, Rui Su, et~al. 2023{\natexlab{a}}.
\newblock Fengwu: Pushing the skillful global medium-range weather forecast beyond 10 days lead.
\newblock \emph{arXiv preprint arXiv:2304.02948}.

\bibitem[{Chen et~al.(2020)Chen, Cao, Ma, and Zhang}]{chen2020deep}
Lei Chen, Yuan Cao, Leiming Ma, and Junping Zhang. 2020.
\newblock A deep learning-based methodology for precipitation nowcasting with radar.
\newblock \emph{Earth and Space Science}, 7(2):e2019EA000812.

\bibitem[{Chen et~al.(2023{\natexlab{b}})Chen, Du, Hu, Wang, and Wang}]{chen2023swinrdm}
Lei Chen, Fei Du, Yuan Hu, Zhibin Wang, and Fan Wang. 2023{\natexlab{b}}.
\newblock Swinrdm: integrate swinrnn with diffusion model towards high-resolution and high-quality weather forecasting.
\newblock In \emph{Proceedings of the AAAI Conference on Artificial Intelligence}, pages 322--330.

\bibitem[{Chen et~al.(2023{\natexlab{c}})Chen, Zhong, Zhang, Cheng, Xu, Qi, and Li}]{chen2023fuxi}
Lei Chen, Xiaohui Zhong, Feng Zhang, Yuan Cheng, Yinghui Xu, Yuan Qi, and Hao Li. 2023{\natexlab{c}}.
\newblock Fuxi: A cascade machine learning forecasting system for 15-day global weather forecast.
\newblock \emph{npj Climate and Atmospheric Science}, 6(1):190.

\bibitem[{Chen et~al.(2023{\natexlab{d}})Chen, Chen, Zhang, Liu, Osman, Farghali, Hua, Al-Fatesh, Ihara, Rooney et~al.}]{chen2023artificial}
Lin Chen, Zhonghao Chen, Yubing Zhang, Yunfei Liu, Ahmed~I Osman, Mohamed Farghali, Jianmin Hua, Ahmed Al-Fatesh, Ikko Ihara, David~W Rooney, et~al. 2023{\natexlab{d}}.
\newblock Artificial intelligence-based solutions for climate change: a review.
\newblock \emph{Environmental Chemistry Letters}, 21(5):2525--2557.

\bibitem[{Chen et~al.(2023{\natexlab{e}})Chen, Xu, Wu, and Huang}]{chen2023group}
Ling Chen, Jiahui Xu, Binqing Wu, and Jianlong Huang. 2023{\natexlab{e}}.
\newblock Group-aware graph neural network for nationwide city air quality forecasting.
\newblock \emph{ACM Transactions on Knowledge Discovery from Data}, 18(3):1--20.

\bibitem[{Chen et~al.(2021{\natexlab{a}})Chen, Peng, Fu, and Ling}]{chen2021autoformer}
Minghao Chen, Houwen Peng, Jianlong Fu, and Haibin Ling. 2021{\natexlab{a}}.
\newblock Autoformer: Searching transformers for visual recognition.
\newblock In \emph{Proceedings of the IEEE/CVF international conference on computer vision}, pages 12270--12280.

\bibitem[{Chen et~al.(2023{\natexlab{f}})Chen, Long, Jiang, Liu, and Zhang}]{chen2023foundation}
Shengchao Chen, Guodong Long, Jing Jiang, Dikai Liu, and Chengqi Zhang. 2023{\natexlab{f}}.
\newblock Foundation models for weather and climate data understanding: A comprehensive survey.
\newblock \emph{arXiv preprint arXiv:2312.03014}.

\bibitem[{Chen et~al.(2022)Chen, Shu, Zhao, Wan, Huang, and Li}]{chen2022dynamic}
Shengchao Chen, Ting Shu, Huan Zhao, Qilin Wan, Jincan Huang, and Cailing Li. 2022.
\newblock Dynamic multiscale fusion generative adversarial network for radar image extrapolation.
\newblock \emph{IEEE Transactions on Geoscience and Remote Sensing}, 60:1--11.

\bibitem[{Chen(2017)}]{bjAirQuality1}
Song Chen. 2017.
\newblock \href {https://archive.ics.uci.edu/dataset/381/beijing+pm2+5+data} {Beijing air quality data set 1}.
\newblock UCI Machine Learning Repository.

\bibitem[{Chen(2019)}]{bjAirQuality2}
Song Chen. 2019.
\newblock \href {https://archive.ics.uci.edu/dataset/501/beijing+multi+site+air+quality+data} {Beijing air quality data set}.
\newblock UCI Machine Learning Repository.

\bibitem[{Chen et~al.(2024)Chen, Wang, Huang, and Tian}]{chen2024coupling}
Yutong Chen, Ya~Wang, Gang Huang, and Qun Tian. 2024.
\newblock Coupling physical factors for precipitation forecast in china with graph neural network.
\newblock \emph{Geophysical Research Letters}, 51(2):e2023GL106676.

\bibitem[{Chen et~al.(2021{\natexlab{b}})Chen, Gao, Wang, and Yan}]{chen2021physics}
Zhihao Chen, Jie Gao, Weikai Wang, and Zheng Yan. 2021{\natexlab{b}}.
\newblock Physics-informed generative neural network: an application to troposphere temperature prediction.
\newblock \emph{Environmental Research Letters}, 16(6):065003.

\bibitem[{Cheng et~al.(2023)Cheng, Zhou, Song, and Zhao}]{cheng2023highway}
Xin Cheng, Jingmei Zhou, Jiachun Song, and Xiangmo Zhao. 2023.
\newblock A highway traffic image enhancement algorithm based on improved gan in complex weather conditions.
\newblock \emph{IEEE Transactions on Intelligent Transportation Systems}.

\bibitem[{Choi et~al.(2023)Choi, Kim, Kim, Jung, and Cho}]{choi2023pct}
Jaeho Choi, Yura Kim, Kwang-Ho Kim, Sung-Hwa Jung, and Ikhyun Cho. 2023.
\newblock Pct-cyclegan: Paired complementary temporal cycle-consistent adversarial networks for radar-based precipitation nowcasting.
\newblock In \emph{Proceedings of the 32nd ACM International Conference on Information and Knowledge Management}, pages 348--358.

\bibitem[{Chowdhury et~al.(2021)Chowdhury, Zhu, and Zhang}]{chowdhury2021mitigating}
Sifat Chowdhury, Kai Zhu, and Yu~Zhang. 2021.
\newblock Mitigating greenhouse gas emissions through generative adversarial networks based wildfire prediction.
\newblock \emph{arXiv preprint arXiv:2108.08952}.

\bibitem[{Chung et~al.(2014)Chung, Gulcehre, Cho, and Bengio}]{chung2014empirical}
Junyoung Chung, Caglar Gulcehre, KyungHyun Cho, and Yoshua Bengio. 2014.
\newblock Empirical evaluation of gated recurrent neural networks on sequence modeling.
\newblock \emph{arXiv preprint arXiv:1412.3555}.

\bibitem[{Coiffier(2011)}]{coiffier2011fundamentals}
Jean Coiffier. 2011.
\newblock \emph{Fundamentals of numerical weather prediction}.
\newblock Cambridge University Press.

\bibitem[{Colverd et~al.(2023)Colverd, Darm, Silverberg, and Kasmanoff}]{colverd2023floodbrain}
Grace Colverd, Paul Darm, Leonard Silverberg, and Noah Kasmanoff. 2023.
\newblock Floodbrain: Flood disaster reporting by web-based retrieval augmented generation with an llm.
\newblock \emph{arXiv preprint arXiv:2311.02597}.

\bibitem[{Croitoru et~al.(2023)Croitoru, Hondru, Ionescu, and Shah}]{croitoru2023diffusion}
Florinel-Alin Croitoru, Vlad Hondru, Radu~Tudor Ionescu, and Mubarak Shah. 2023.
\newblock Diffusion models in vision: A survey.
\newblock \emph{IEEE Transactions on Pattern Analysis and Machine Intelligence}.

\bibitem[{Dai et~al.(2023)Dai, Yang, Liu, Liu, and Liu}]{dai2023timeddpm}
Yun Dai, Chao Yang, Kaixin Liu, Angpeng Liu, and Yi~Liu. 2023.
\newblock Timeddpm: Time series augmentation strategy for industrial soft sensing.
\newblock \emph{IEEE Sensors Journal}.

\bibitem[{Danandeh~Mehr et~al.(2023)Danandeh~Mehr, Rikhtehgar~Ghiasi, Yaseen, Sorman, and Abualigah}]{danandeh2023novel}
Ali Danandeh~Mehr, Amir Rikhtehgar~Ghiasi, Zaher~Mundher Yaseen, Ali~Unal Sorman, and Laith Abualigah. 2023.
\newblock A novel intelligent deep learning predictive model for meteorological drought forecasting.
\newblock \emph{Journal of Ambient Intelligence and Humanized Computing}, 14(8):10441--10455.

\bibitem[{Das et~al.(2023)Das, Kong, Sen, and Zhou}]{das2023decoder}
Abhimanyu Das, Weihao Kong, Rajat Sen, and Yichen Zhou. 2023.
\newblock A decoder-only foundation model for time-series forecasting.
\newblock \emph{arXiv preprint arXiv:2310.10688}.

\bibitem[{Das et~al.(2024)Das, Posch, Barber, Hicks, Vandal, Duffy, Singh, van Werkhoven, and Ganguly}]{das2024hybrid}
Puja Das, August Posch, Nathan Barber, Michael Hicks, Thomas~J Vandal, Kate Duffy, Debjani Singh, Katie van Werkhoven, and Auroop~R Ganguly. 2024.
\newblock Hybrid physics-ai outperforms numerical weather prediction for extreme precipitation nowcasting.
\newblock \emph{arXiv preprint arXiv:2407.11317}.

\bibitem[{De~B{\'e}zenac et~al.(2019)De~B{\'e}zenac, Pajot, and Gallinari}]{de2019deep}
Emmanuel De~B{\'e}zenac, Arthur Pajot, and Patrick Gallinari. 2019.
\newblock Deep learning for physical processes: Incorporating prior scientific knowledge.
\newblock \emph{Journal of Statistical Mechanics: Theory and Experiment}, 2019(12):124009.

\bibitem[{de~Witt et~al.(2021)de~Witt, Tong, Zantedeschi, De~Martini, Kalaitzis, Chantry, Watson-Parris, and Bilinski}]{de2021rainbench}
Christian~Schroeder de~Witt, Catherine Tong, Valentina Zantedeschi, Daniele De~Martini, Alfredo Kalaitzis, Matthew Chantry, Duncan Watson-Parris, and Piotr Bilinski. 2021.
\newblock Rainbench: Towards data-driven global precipitation forecasting from satellite imagery.
\newblock In \emph{Proceedings of the AAAI Conference on Artificial Intelligence}, pages 14902--14910.

\bibitem[{Descombes et~al.(2020)Descombes, Pitteloud, Glauser, Defossez, Kergunteuil, Allard, Rasmann, and Pellissier}]{descombes2020novel}
Patrice Descombes, Camille Pitteloud, Ga{\"e}tan Glauser, Emmanuel Defossez, Alan Kergunteuil, Pierre-Marie Allard, Sergio Rasmann, and Lo{\"\i}c Pellissier. 2020.
\newblock Novel trophic interactions under climate change promote alpine plant coexistence.
\newblock \emph{Science}, 370(6523):1469--1473.

\bibitem[{Dhyani and Pandya(2021)}]{dhyani2021deep}
Yogesh Dhyani and Rahul~Jashvantbhai Pandya. 2021.
\newblock Deep learning oriented satellite remote sensing for drought and prediction in agriculture.
\newblock In \emph{2021 IEEE 18th India Council International Conference (INDICON)}, pages 1--5. IEEE.

\bibitem[{Dikshit and Pradhan(2021)}]{dikshit2021explainable}
Abhirup Dikshit and Biswajeet Pradhan. 2021.
\newblock Explainable ai in drought forecasting.
\newblock \emph{Machine Learning with Applications}, 6:100192.

\bibitem[{Dikshit et~al.(2021)Dikshit, Pradhan, and Alamri}]{dikshit2021long}
Abhirup Dikshit, Biswajeet Pradhan, and Abdullah~M Alamri. 2021.
\newblock Long lead time drought forecasting using lagged climate variables and a stacked long short-term memory model.
\newblock \emph{Science of The Total Environment}, 755:142638.

\bibitem[{do~Lago et~al.(2023)do~Lago, Giacomoni, Bentivoglio, Taormina, Junior, and Mendiondo}]{do2023generalizing}
Cesar~AF do~Lago, Marcio~H Giacomoni, Roberto Bentivoglio, Riccardo Taormina, Marcus N~Gomes Junior, and Eduardo~M Mendiondo. 2023.
\newblock Generalizing rapid flood predictions to unseen urban catchments with conditional generative adversarial networks.
\newblock \emph{Journal of Hydrology}, 618:129276.

\bibitem[{Dosovitskiy et~al.(2020)Dosovitskiy, Beyer, Kolesnikov, Weissenborn, Zhai, Unterthiner, Dehghani, Minderer, Heigold, Gelly et~al.}]{dosovitskiy2020image}
Alexey Dosovitskiy, Lucas Beyer, Alexander Kolesnikov, Dirk Weissenborn, Xiaohua Zhai, Thomas Unterthiner, Mostafa Dehghani, Matthias Minderer, Georg Heigold, Sylvain Gelly, et~al. 2020.
\newblock An image is worth 16x16 words: Transformers for image recognition at scale.
\newblock \emph{arXiv preprint arXiv:2010.11929}.

\bibitem[{Du and Li(2023)}]{du2023diffusion}
Zhekai Du and Jingjing Li. 2023.
\newblock Diffusion-based probabilistic uncertainty estimation for active domain adaptation.
\newblock \emph{Advances in Neural Information Processing Systems}, 36:17129--17155.

\bibitem[{Durai and Bhradwaj(2014)}]{durai2014evaluation}
VR~Durai and Rashmi Bhradwaj. 2014.
\newblock Evaluation of statistical bias correction methods for numerical weather prediction model forecasts of maximum and minimum temperatures.
\newblock \emph{Natural Hazards}, 73:1229--1254.

\bibitem[{Dzulhijjah et~al.(2023)Dzulhijjah, Majid, Alwanda, Kusuma, Zakaria, Kusrini, and Kusnawi}]{dzulhijjah2023comparative}
Dwi~Ahmad Dzulhijjah, Muhammad~Nurkholis Majid, Almi~Yulistia Alwanda, Dimas~Candra Kusuma, Fariz Zakaria, Kusrini Kusrini, and Kusnawi Kusnawi. 2023.
\newblock Comparative analysis of hybrid long short-term memory models for fire danger index forecasting with weather data.
\newblock In \emph{2023 6th International Conference on Information and Communications Technology (ICOIACT)}, pages 165--170. IEEE.

\bibitem[{Ehsani et~al.(2022)Ehsani, Zarei, Gupta, Barnard, Lyons, and Behrangi}]{ehsani2022nowcasting}
Mohammad~Reza Ehsani, Ariyan Zarei, Hoshin~Vijai Gupta, Kobus Barnard, Eric Lyons, and Ali Behrangi. 2022.
\newblock Nowcasting-nets: Representation learning to mitigate latency gap of satellite precipitation products using convolutional and recurrent neural networks.
\newblock \emph{IEEE Transactions on Geoscience and Remote Sensing}, 60:1--21.

\bibitem[{Elbeltagi et~al.(2024)Elbeltagi, Srivastava, Ehsan, Sharma, Yu, Khadke, Gautam, Awad, and Jinsong}]{elbeltagi2024advanced}
Ahmed Elbeltagi, Aman Srivastava, Muhsan Ehsan, Gitika Sharma, Jiawen Yu, Leena Khadke, Vinay~Kumar Gautam, Ahmed Awad, and Deng Jinsong. 2024.
\newblock Advanced stacked integration method for forecasting long-term drought severity: Cnn with machine learning models.
\newblock \emph{Journal of Hydrology: Regional Studies}, 53:101759.

\bibitem[{Espeholt et~al.(2022)Espeholt, Agrawal, S{\o}nderby, Kumar, Heek, Bromberg, Gazen, Carver, Andrychowicz, Hickey et~al.}]{espeholt2022deep}
Lasse Espeholt, Shreya Agrawal, Casper S{\o}nderby, Manoj Kumar, Jonathan Heek, Carla Bromberg, Cenk Gazen, Rob Carver, Marcin Andrychowicz, Jason Hickey, et~al. 2022.
\newblock Deep learning for twelve hour precipitation forecasts.
\newblock \emph{Nature communications}, 13(1):1--10.

\bibitem[{Evans(2022)}]{evans2022partial}
Lawrence~C Evans. 2022.
\newblock \emph{Partial differential equations}, volume~19.
\newblock American Mathematical Society.

\bibitem[{Fang et~al.(2021)Fang, Xue, Shen, and Sheng}]{fang2021survey}
Wei Fang, Qiongying Xue, Liang Shen, and Victor~S Sheng. 2021.
\newblock Survey on the application of deep learning in extreme weather prediction.
\newblock \emph{Atmosphere}, 12(6):661.

\bibitem[{Feng et~al.(2023)Feng, Tan, and He}]{feng2023physics}
Dongyu Feng, Zeli Tan, and QiZhi He. 2023.
\newblock Physics-informed neural networks of the saint-venant equations for downscaling a large-scale river model.
\newblock \emph{Water Resources Research}, 59(2):e2022WR033168.

\bibitem[{Feng et~al.(2024{\natexlab{a}})Feng, Lai, Yang, Zhou, Yin, and Zhao}]{feng2024timesieve}
Ninghui Feng, Songning Lai, Jiayu Yang, Fobao Zhou, Zhenxiao Yin, and Hang Zhao. 2024{\natexlab{a}}.
\newblock Timesieve: Extracting temporal dynamics through information bottlenecks.
\newblock \emph{arXiv preprint arXiv:2406.05036}.

\bibitem[{Feng et~al.(2024{\natexlab{b}})Feng, Miao, Zhang, and Zhao}]{feng2024latent}
Shibo Feng, Chunyan Miao, Zhong Zhang, and Peilin Zhao. 2024{\natexlab{b}}.
\newblock Latent diffusion transformer for probabilistic time series forecasting.
\newblock In \emph{Proceedings of the AAAI Conference on Artificial Intelligence}, volume~38, pages 11979--11987.

\bibitem[{Ferchichi et~al.(2024)Ferchichi, Chihaoui, and Ferchichi}]{ferchichi2024spatio}
Ahlem Ferchichi, Mejda Chihaoui, and Aya Ferchichi. 2024.
\newblock Spatio-temporal modeling of climate change impacts on drought forecast using generative adversarial network: A case study in africa.
\newblock \emph{Expert Systems with Applications}, 238:122211.

\bibitem[{Flandroy et~al.(2018)Flandroy, Poutahidis, Berg, Clarke, Dao, Decaestecker, Furman, Haahtela, Massart, Plovier et~al.}]{flandroy2018impact}
Lucette Flandroy, Theofilos Poutahidis, Gabriele Berg, Gerard Clarke, Maria-Carlota Dao, Ellen Decaestecker, Eeva Furman, Tari Haahtela, S{\'e}bastien Massart, Hubert Plovier, et~al. 2018.
\newblock The impact of human activities and lifestyles on the interlinked microbiota and health of humans and of ecosystems.
\newblock \emph{Science of the total environment}, 627:1018--1038.

\bibitem[{Franch et~al.(2024)Franch, Tomasi, Wanjari, Poli, Cardinali, Alberoni, and Cristoforetti}]{franch2024gptcast}
Gabriele Franch, Elena Tomasi, Rishabh Wanjari, Virginia Poli, Chiara Cardinali, Pier~Paolo Alberoni, and Marco Cristoforetti. 2024.
\newblock Gptcast: a weather language model for precipitation nowcasting.
\newblock \emph{arXiv preprint arXiv:2407.02089}.

\bibitem[{Gao et~al.(2023{\natexlab{a}})Gao, Xiong, Gao, Jia, Pan, Bi, Dai, Sun, and Wang}]{gao2023retrieval}
Yunfan Gao, Yun Xiong, Xinyu Gao, Kangxiang Jia, Jinliu Pan, Yuxi Bi, Yi~Dai, Jiawei Sun, and Haofen Wang. 2023{\natexlab{a}}.
\newblock Retrieval-augmented generation for large language models: A survey.
\newblock \emph{arXiv preprint arXiv:2312.10997}.

\bibitem[{Gao et~al.(2023{\natexlab{b}})Gao, Shi, Han, Wang, Jin, Maddix, Zhu, Li, and Wang}]{gao2023prediff}
Zhihan Gao, Xingjian Shi, Boran Han, Hao Wang, Xiaoyong Jin, Danielle Maddix, Yi~Zhu, Mu~Li, and Yuyang Wang. 2023{\natexlab{b}}.
\newblock Prediff: Precipitation nowcasting with latent diffusion models.
\newblock \emph{arXiv preprint arXiv:2307.10422}.

\bibitem[{Gao et~al.(2024)Gao, Shi, Han, Wang, Jin, Maddix, Zhu, Li, and Wang}]{gao2024prediff}
Zhihan Gao, Xingjian Shi, Boran Han, Hao Wang, Xiaoyong Jin, Danielle Maddix, Yi~Zhu, Mu~Li, and Yuyang~Bernie Wang. 2024.
\newblock Prediff: Precipitation nowcasting with latent diffusion models.
\newblock \emph{Advances in Neural Information Processing Systems}, 36.

\bibitem[{Gao et~al.(2020)Gao, Shi, Wang, Yeung, chun Woo, and Wong}]{Gao2020}
Zhihan Gao, Xingjian Shi, Hao Wang, Dit-Yan Yeung, Wang chun Woo, and Wai-Kin Wong. 2020.
\newblock \href {https://www.amazon.science/publications/deep-learning-and-the-weather-forecasting-problem-precipitation-nowcasting} {Deep learning and the weather forecasting problem: Precipitation nowcasting}.
\newblock \emph{Deep Learning for the Earth Sciences}.

\bibitem[{Gao et~al.(2021)Gao, Shi, Wang, Yeung, Woo, and Wong}]{gao2021deep}
Zhihan Gao, Xingjian Shi, Hao Wang, Dit-Yan Yeung, Wang-chun Woo, and Wai-Kin Wong. 2021.
\newblock Deep learning and the weather forecasting problem: Precipitation nowcasting.
\newblock \emph{Deep Learning for the Earth Sciences: A Comprehensive Approach to Remote Sensing, Climate Science, and Geosciences}, pages 218--239.

\bibitem[{Gao et~al.(2022)Gao, Shi, Wang, Zhu, Wang, Li, and Yeung}]{gao2022earthformer}
Zhihan Gao, Xingjian Shi, Hao Wang, Yi~Zhu, Yuyang~Bernie Wang, Mu~Li, and Dit-Yan Yeung. 2022.
\newblock Earthformer: Exploring space-time transformers for earth system forecasting.
\newblock \emph{Advances in Neural Information Processing Systems}, 35:25390--25403.

\bibitem[{Gao et~al.(2023{\natexlab{c}})Gao, Li, Yu, and Xu}]{gao2023global}
Ziheng Gao, Zhuolin Li, Jie Yu, and Lingyu Xu. 2023{\natexlab{c}}.
\newblock Global spatiotemporal graph attention network for sea surface temperature prediction.
\newblock \emph{IEEE Geoscience and Remote Sensing Letters}, 20:1--5.

\bibitem[{Gauthier(2014)}]{gauthier2014conditional}
Jon Gauthier. 2014.
\newblock Conditional generative adversarial nets for convolutional face generation.
\newblock \emph{Class project for Stanford CS231N: convolutional neural networks for visual recognition, Winter semester}, 2014(5):2.

\bibitem[{Geng and Wang(2021)}]{geng2021spatiotemporal}
Huantong Geng and Tianlei Wang. 2021.
\newblock Spatiotemporal model based on deep learning for enso forecasts.
\newblock \emph{Atmosphere}, 12(7):810.

\bibitem[{Geng et~al.(2024)Geng, Wu, Zhuang, Geng, Xie, and Shi}]{geng2024ms}
Huantong Geng, Fangli Wu, Xiaoran Zhuang, Liangchao Geng, Boyang Xie, and Zhanpeng Shi. 2024.
\newblock The ms-radarformer: A transformer-based multi-scale deep learning model for radar echo extrapolation.
\newblock \emph{Remote Sensing}, 16(2):274.

\bibitem[{Gibson et~al.(2021)Gibson, Chapman, Altinok, Delle~Monache, DeFlorio, and Waliser}]{gibson2021training}
Peter~B Gibson, William~E Chapman, Alphan Altinok, Luca Delle~Monache, Michael~J DeFlorio, and Duane~E Waliser. 2021.
\newblock Training machine learning models on climate model output yields skillful interpretable seasonal precipitation forecasts.
\newblock \emph{Communications Earth \& Environment}, 2(1):159.

\bibitem[{Giffard-Roisin et~al.(2020)Giffard-Roisin, Yang, Charpiat, Kumler~Bonfanti, K{\'e}gl, and Monteleoni}]{giffard2020tropical}
Sophie Giffard-Roisin, Mo~Yang, Guillaume Charpiat, Christina Kumler~Bonfanti, Bal{\'a}zs K{\'e}gl, and Claire Monteleoni. 2020.
\newblock Tropical cyclone track forecasting using fused deep learning from aligned reanalysis data.
\newblock \emph{Frontiers in big Data}, 3:1.

\bibitem[{Gilik et~al.(2022)Gilik, Ogrenci, and Ozmen}]{gilik2022air}
Aysenur Gilik, Arif~Selcuk Ogrenci, and Atilla Ozmen. 2022.
\newblock Air quality prediction using cnn+ lstm-based hybrid deep learning architecture.
\newblock \emph{Environmental science and pollution research}, pages 1--19.

\bibitem[{Gilmer et~al.(2017)Gilmer, Schoenholz, Riley, Vinyals, and Dahl}]{gilmer2017neural}
Justin Gilmer, Samuel~S Schoenholz, Patrick~F Riley, Oriol Vinyals, and George~E Dahl. 2017.
\newblock Neural message passing for quantum chemistry.
\newblock In \emph{International conference on machine learning}, pages 1263--1272. PMLR.

\bibitem[{Gong et~al.(2024)Gong, Bai, Ye, Xu, Liu, Dai, Yang, and Ouyang}]{gong2024cascast}
Junchao Gong, Lei Bai, Peng Ye, Wanghan Xu, Na~Liu, Jianhua Dai, Xiaokang Yang, and Wanli Ouyang. 2024.
\newblock Cascast: Skillful high-resolution precipitation nowcasting via cascaded modelling.
\newblock \emph{arXiv preprint arXiv:2402.04290}.

\bibitem[{Goodfellow et~al.(2014)Goodfellow, Pouget-Abadie, Mirza, Xu, Warde-Farley, Ozair, Courville, and Bengio}]{goodfellow2014generative}
Ian Goodfellow, Jean Pouget-Abadie, Mehdi Mirza, Bing Xu, David Warde-Farley, Sherjil Ozair, Aaron Courville, and Yoshua Bengio. 2014.
\newblock Generative adversarial nets.
\newblock \emph{Advances in neural information processing systems}, 27.

\bibitem[{Gopu et~al.(2023)Gopu, Ramakrishnan, Balasubramanian, and Srinidhi}]{gopu2023comparative}
Arunkumar Gopu, Anjali Ramakrishnan, Ganesan Balasubramanian, and Kuna Srinidhi. 2023.
\newblock A comparative study on forest fire prediction using arima, sarima, lstm, and gru methods.
\newblock In \emph{2023 IEEE International Conference on Contemporary Computing and Communications (InC4)}, volume~1, pages 1--5. IEEE.

\bibitem[{Goswami et~al.(2024)Goswami, Szafer, Choudhry, Cai, Li, and Dubrawski}]{goswami2024moment}
Mononito Goswami, Konrad Szafer, Arjun Choudhry, Yifu Cai, Shuo Li, and Artur Dubrawski. 2024.
\newblock Moment: A family of open time-series foundation models.
\newblock \emph{arXiv preprint arXiv:2402.03885}.

\bibitem[{Gu and Dao(2023)}]{gu2023mamba}
Albert Gu and Tri Dao. 2023.
\newblock Mamba: Linear-time sequence modeling with selective state spaces.
\newblock \emph{arXiv preprint arXiv:2312.00752}.

\bibitem[{Gu et~al.(2020)Gu, Dao, Ermon, Rudra, and R{\'e}}]{gu2020hippo}
Albert Gu, Tri Dao, Stefano Ermon, Atri Rudra, and Christopher R{\'e}. 2020.
\newblock Hippo: Recurrent memory with optimal polynomial projections.
\newblock \emph{Advances in neural information processing systems}, 33:1474--1487.

\bibitem[{Guen and Thome(2020)}]{guen2020disentangling}
Vincent~Le Guen and Nicolas Thome. 2020.
\newblock Disentangling physical dynamics from unknown factors for unsupervised video prediction.
\newblock In \emph{Proceedings of the IEEE/CVF conference on computer vision and pattern recognition}, pages 11474--11484.

\bibitem[{Guibas et~al.(2021)Guibas, Mardani, Li, Tao, Anandkumar, and Catanzaro}]{guibas2021adaptive}
John Guibas, Morteza Mardani, Zongyi Li, Andrew Tao, Anima Anandkumar, and Bryan Catanzaro. 2021.
\newblock Adaptive fourier neural operators: Efficient token mixers for transformers.
\newblock \emph{arXiv preprint arXiv:2111.13587}.

\bibitem[{Guidotti et~al.(2018)Guidotti, Monreale, Ruggieri, Turini, Giannotti, and Pedreschi}]{guidotti2018survey}
Riccardo Guidotti, Anna Monreale, Salvatore Ruggieri, Franco Turini, Fosca Giannotti, and Dino Pedreschi. 2018.
\newblock A survey of methods for explaining black box models.
\newblock \emph{ACM computing surveys (CSUR)}, 51(5):1--42.

\bibitem[{Gyaneshwar et~al.(2023)Gyaneshwar, Mishra, Chadha, Raj~Vincent, Rajinikanth, Pattukandan~Ganapathy, and Srinivasan}]{gyaneshwar2023contemporary}
Amogh Gyaneshwar, Anirudh Mishra, Utkarsh Chadha, PM~Durai Raj~Vincent, Venkatesan Rajinikanth, Ganapathy Pattukandan~Ganapathy, and Kathiravan Srinivasan. 2023.
\newblock A contemporary review on deep learning models for drought prediction.
\newblock \emph{Sustainability}, 15(7):6160.

\bibitem[{Ham et~al.(2019)Ham, Kim, and Luo}]{ham2019deep}
Yoo-Geun Ham, Jeong-Hwan Kim, and Jing-Jia Luo. 2019.
\newblock Deep learning for multi-year enso forecasts.
\newblock \emph{Nature}, 573(7775):568--572.

\bibitem[{Han et~al.(2021)Han, Liang, Chen, Zhang, and Ge}]{han2021convective}
Lei Han, He~Liang, Haonan Chen, Wei Zhang, and Yurong Ge. 2021.
\newblock Convective precipitation nowcasting using u-net model.
\newblock \emph{IEEE Transactions on Geoscience and Remote Sensing}, 60:1--8.

\bibitem[{Han et~al.(2024{\natexlab{a}})Han, Chen, Guo, Xu, and Bai}]{han2024cra5}
Tao Han, Zhenghao Chen, Song Guo, Wanghan Xu, and Lei Bai. 2024{\natexlab{a}}.
\newblock Cra5: Extreme compression of era5 for portable global climate and weather research via an efficient variational transformer.
\newblock \emph{arXiv preprint arXiv:2405.03376}.

\bibitem[{Han et~al.(2024{\natexlab{b}})Han, Guo, Chen, Xu, and Bai}]{han2024weather}
Tao Han, Song Guo, Zhenghao Chen, Wanghan Xu, and Lei Bai. 2024{\natexlab{b}}.
\newblock Weather-5k: A large-scale global station weather dataset towards comprehensive time-series forecasting benchmark.
\newblock \emph{arXiv preprint arXiv:2406.14399}.

\bibitem[{Hao et~al.(2023)Hao, Yan, and Chiang}]{hao2023forecasting}
Ruonan Hao, Huaxiang Yan, and Yen-Ming Chiang. 2023.
\newblock Forecasting the propagation from meteorological to hydrological and agricultural drought in the huaihe river basin with machine learning methods.
\newblock \emph{Remote Sensing}, 15(23):5524.

\bibitem[{Harder et~al.(2022)Harder, Yang, Ramesh, Sattigeri, Hernandez-Garcia, Watson, Szwarcman, and Rolnick}]{harder2022generating}
Paula Harder, Qidong Yang, Venkatesh Ramesh, Prasanna Sattigeri, Alex Hernandez-Garcia, Campbell Watson, Daniela Szwarcman, and David Rolnick. 2022.
\newblock Generating physically-consistent high-resolution climate data with hard-constrained neural networks.
\newblock \emph{arXiv preprint arXiv:2208.05424}, 18:109--122.

\bibitem[{He et~al.(2019)He, Lin, Liu, Ding, and Jiang}]{he2019dlenso}
Dandan He, Pengfei Lin, Hailong Liu, Lei Ding, and Jinrong Jiang. 2019.
\newblock Dlenso: A deep learning enso forecasting model.
\newblock In \emph{PRICAI 2019: Trends in Artificial Intelligence: 16th Pacific Rim International Conference on Artificial Intelligence, Cuvu, Yanuca Island, Fiji, August 26--30, 2019, Proceedings, Part II 16}, pages 12--23. Springer.

\bibitem[{He et~al.(2022)He, Chen, Xie, Li, Doll{\'a}r, and Girshick}]{he2022masked}
Kaiming He, Xinlei Chen, Saining Xie, Yanghao Li, Piotr Doll{\'a}r, and Ross Girshick. 2022.
\newblock Masked autoencoders are scalable vision learners.
\newblock In \emph{Proceedings of the IEEE/CVF conference on computer vision and pattern recognition}, pages 16000--16009.

\bibitem[{He et~al.(2017)He, Gkioxari, Doll{\'a}r, and Girshick}]{he2017mask}
Kaiming He, Georgia Gkioxari, Piotr Doll{\'a}r, and Ross Girshick. 2017.
\newblock Mask r-cnn.
\newblock In \emph{Proceedings of the IEEE international conference on computer vision}, pages 2961--2969.

\bibitem[{He et~al.(2016)He, Zhang, Ren, and Sun}]{he2016deep}
Kaiming He, Xiangyu Zhang, Shaoqing Ren, and Jian Sun. 2016.
\newblock Deep residual learning for image recognition.
\newblock In \emph{Proceedings of the IEEE conference on computer vision and pattern recognition}, pages 770--778.

\bibitem[{He et~al.(2024{\natexlab{a}})He, Zhu, Zhao, Song, and Huang}]{he2024interpretable}
Qi~He, Zihang Zhu, Danfeng Zhao, Wei Song, and Dongmei Huang. 2024{\natexlab{a}}.
\newblock An interpretable deep learning approach for detecting marine heatwaves patterns.
\newblock \emph{Applied Sciences}, 14(2):601.

\bibitem[{He et~al.(2024{\natexlab{b}})He, Zhang, and Chew}]{he2024data}
Renfei He, Limao Zhang, and Alvin Wei~Ze Chew. 2024{\natexlab{b}}.
\newblock Data-driven multi-step prediction and analysis of monthly rainfall using explainable deep learning.
\newblock \emph{Expert Systems with Applications}, 235:121160.

\bibitem[{He et~al.(2024{\natexlab{c}})He, Huang, Jiang, Nie, Wang, Wang, and Chen}]{he2024foundation}
Yuting He, Fuxiang Huang, Xinrui Jiang, Yuxiang Nie, Minghao Wang, Jiguang Wang, and Hao Chen. 2024{\natexlab{c}}.
\newblock Foundation model for advancing healthcare: Challenges, opportunities, and future directions.
\newblock \emph{arXiv preprint arXiv:2404.03264}.

\bibitem[{Herruzo et~al.(2021)Herruzo, Gruca, Lliso, Calbet, R{\'\i}podas, Hochreiter, Kopp, and Kreil}]{herruzo2021high}
Pedro Herruzo, Aleksandra Gruca, Lloren{\c{c}} Lliso, Xavier Calbet, Pilar R{\'\i}podas, Sepp Hochreiter, Michael Kopp, and David~P Kreil. 2021.
\newblock High-resolution multi-channel weather forecasting--first insights on transfer learning from the weather4cast competitions 2021.
\newblock In \emph{2021 IEEE International Conference on Big Data (Big Data)}, pages 5750--5757. IEEE.

\bibitem[{Hertz et~al.(2022)Hertz, Mokady, Tenenbaum, Aberman, Pritch, and Cohen-Or}]{hertz2022prompt}
Amir Hertz, Ron Mokady, Jay Tenenbaum, Kfir Aberman, Yael Pritch, and Daniel Cohen-Or. 2022.
\newblock Prompt-to-prompt image editing with cross attention control.
\newblock \emph{arXiv preprint arXiv:2208.01626}.

\bibitem[{Ho et~al.(2020)Ho, Jain, and Abbeel}]{ho2020denoising}
Jonathan Ho, Ajay Jain, and Pieter Abbeel. 2020.
\newblock Denoising diffusion probabilistic models.
\newblock \emph{Advances in neural information processing systems}, 33:6840--6851.

\bibitem[{Hochreiter and Schmidhuber(1997)}]{hochreiter1997long}
Sepp Hochreiter and J{\"u}rgen Schmidhuber. 1997.
\newblock Long short-term memory.
\newblock \emph{Neural computation}, 9(8):1735--1780.

\bibitem[{Hu et~al.(2021{\natexlab{a}})Hu, Shen, Wallis, Allen-Zhu, Li, Wang, Wang, and Chen}]{hu2021lora}
Edward~J Hu, Yelong Shen, Phillip Wallis, Zeyuan Allen-Zhu, Yuanzhi Li, Shean Wang, Lu~Wang, and Weizhu Chen. 2021{\natexlab{a}}.
\newblock Lora: Low-rank adaptation of large language models.
\newblock \emph{arXiv preprint arXiv:2106.09685}.

\bibitem[{Hu et~al.(2021{\natexlab{b}})Hu, Weng, Huang, Gao, Ye, and You}]{hu2021deep}
Jie Hu, Bin Weng, Tianqiang Huang, Jianyun Gao, Feng Ye, and Lijun You. 2021{\natexlab{b}}.
\newblock Deep residual convolutional neural network combining dropout and transfer learning for enso forecasting.
\newblock \emph{Geophysical Research Letters}, 48(24):e2021GL093531.

\bibitem[{Hu et~al.(2023)Hu, Chen, Wang, and Li}]{hu2023swinvrnn}
Yuan Hu, Lei Chen, Zhibin Wang, and Hao Li. 2023.
\newblock Swinvrnn: A data-driven ensemble forecasting model via learned distribution perturbation.
\newblock \emph{Journal of Advances in Modeling Earth Systems}, 15(2):e2022MS003211.

\bibitem[{Huang et~al.(2019)Huang, Vega-Westhoff, and Sriver}]{huang2019analyzing}
Andrew Huang, Ben Vega-Westhoff, and Ryan~L Sriver. 2019.
\newblock Analyzing el ni{\~n}o--southern oscillation predictability using long-short-term-memory models.
\newblock \emph{Earth and Space Science}, 6(2):212--221.

\bibitem[{Huffman et~al.(2020)Huffman, Bolvin, Braithwaite, Hsu, Joyce, Kidd, Nelkin, Sorooshian, Stocker, Tan et~al.}]{huffman2020integrated}
George~J Huffman, David~T Bolvin, Dan Braithwaite, Kuo-Lin Hsu, Robert~J Joyce, Christopher Kidd, Eric~J Nelkin, Soroosh Sorooshian, Erich~F Stocker, Jackson Tan, et~al. 2020.
\newblock Integrated multi-satellite retrievals for the global precipitation measurement (gpm) mission (imerg).
\newblock \emph{Satellite precipitation measurement: Volume 1}, pages 343--353.

\bibitem[{Huot et~al.(2022)Huot, Hu, Goyal, Sankar, Ihme, and Chen}]{huot2022next}
Fantine Huot, R~Lily Hu, Nita Goyal, Tharun Sankar, Matthias Ihme, and Yi-Fan Chen. 2022.
\newblock Next day wildfire spread: A machine learning dataset to predict wildfire spreading from remote-sensing data.
\newblock \emph{IEEE Transactions on Geoscience and Remote Sensing}, 60:1--13.

\bibitem[{Inoue and Misumi(2022)}]{inoue2022learning}
Tsuyoshi Inoue and Ryohei Misumi. 2022.
\newblock Learning from precipitation events in the wider domain to improve the performance of a deep learning--based precipitation nowcasting model.
\newblock \emph{Weather and Forecasting}, 37(6):1013--1026.

\bibitem[{Jiang et~al.(2024)Jiang, Zhang, Li, Zhang, Hu, Gao, and Duan}]{jiang2024advancing}
Wenjun Jiang, Jize Zhang, Yuerong Li, Dongqin Zhang, Gang Hu, Huanxiang Gao, and Zhongdong Duan. 2024.
\newblock Advancing storm surge forecasting from scarce observation data: A causal-inference based spatio-temporal graph neural network approach.
\newblock \emph{Coastal Engineering}, 190:104512.

\bibitem[{Jiang et~al.(2022)Jiang, Wang, Su, Li, Wang, Zheng, Wang, and Meng}]{jiang2022modeling}
Wenyu Jiang, Fei Wang, Guofeng Su, Xin Li, Guanning Wang, Xinxin Zheng, Ting Wang, and Qingxiang Meng. 2022.
\newblock Modeling wildfire spread with an irregular graph network.
\newblock \emph{Fire}, 5(6):185.

\bibitem[{Jing et~al.(2019)Jing, Li, Ding, Sun, Tang, and Cai}]{jing2019aenn}
JR~Jing, Qian Li, XY~Ding, NL~Sun, Rong Tang, and YL~Cai. 2019.
\newblock Aenn: A generative adversarial neural network for weather radar echo extrapolation.
\newblock \emph{The International Archives of the Photogrammetry, Remote Sensing and Spatial Information Sciences}, 42:89--94.

\bibitem[{Juhasz et~al.(2024)Juhasz, Dutia, Franks, Delahunty, Mills, and Pim}]{juhasz2024responsible}
Matyas Juhasz, Kalyan Dutia, Henry Franks, Conor Delahunty, Patrick~Fawbert Mills, and Harrison Pim. 2024.
\newblock Responsible retrieval augmented generation for climate decision making from documents.
\newblock \emph{arXiv preprint arXiv:2410.23902}.

\bibitem[{Kabir et~al.(2020)Kabir, Patidar, Xia, Liang, Neal, and Pender}]{kabir2020deep}
Syed Kabir, Sandhya Patidar, Xilin Xia, Qiuhua Liang, Jeffrey Neal, and Gareth Pender. 2020.
\newblock A deep convolutional neural network model for rapid prediction of fluvial flood inundation.
\newblock \emph{Journal of Hydrology}, 590:125481.

\bibitem[{Kashinath et~al.(2021)Kashinath, Mudigonda, Kim, Kapp-Schwoerer, Graubner, Karaismailoglu, Von~Kleist, Kurth, Greiner, Mahesh et~al.}]{kashinath2021climatenet}
Karthik Kashinath, Mayur Mudigonda, Sol Kim, Lukas Kapp-Schwoerer, Andre Graubner, Ege Karaismailoglu, Leo Von~Kleist, Thorsten Kurth, Annette Greiner, Ankur Mahesh, et~al. 2021.
\newblock Climatenet: An expert-labeled open dataset and deep learning architecture for enabling high-precision analyses of extreme weather.
\newblock \emph{Geoscientific Model Development}, 14(1):107--124.

\bibitem[{Kazadi et~al.(2024)Kazadi, Doss-Gollin, Sebastian, and Silva}]{kazadi2024floodgnn}
Arnold Kazadi, James Doss-Gollin, Antonia Sebastian, and Arlei Silva. 2024.
\newblock Floodgnn-gru: a spatio-temporal graph neural network for flood prediction.
\newblock \emph{Environmental Data Science}, 3:e21.

\bibitem[{Keisler(2022)}]{keisler2022forecasting}
Ryan Keisler. 2022.
\newblock Forecasting global weather with graph neural networks.
\newblock \emph{arXiv preprint arXiv:2202.07575}.

\bibitem[{Khan et~al.(2023)Khan, Mustafa, Hossain, Shams, and Julius}]{khan2023short}
MMH Khan, MRU Mustafa, MS~Hossain, S~Shams, and AD~Julius. 2023.
\newblock Short-term and long-term rainfall forecasting using arima model.
\newblock \emph{International Journal of Environmental Science and Development}, 14(5):292--298.

\bibitem[{Khennou et~al.(2021)Khennou, Ghaoui, and Akhloufi}]{khennou2021forest}
Fadoua Khennou, Jade Ghaoui, and Moulay~A Akhloufi. 2021.
\newblock Forest fire spread prediction using deep learning.
\newblock In \emph{Geospatial informatics XI}, volume 11733, pages 106--117. SPIE.

\bibitem[{Kim et~al.(2024)Kim, Ham, Kim, Li, and Ma}]{kim2024improvement}
Jeong-Hwan Kim, Yoo-Geun Ham, Daehyun Kim, Tim Li, and Chen Ma. 2024.
\newblock Improvement in forecasting short-term tropical cyclone intensity change and their rapid intensification using deep learning.
\newblock \emph{Artificial Intelligence for the Earth Systems}, 3(2):e230052.

\bibitem[{Kim et~al.(2022{\natexlab{a}})Kim, Kang, Shin, Yoon, Eom, Shin, and Yun}]{kim2022region}
Taehyeon Kim, Shinhwan Kang, Hyeonjeong Shin, Deukryeol Yoon, Seongha Eom, Kijung Shin, and Se-Young Yun. 2022{\natexlab{a}}.
\newblock Region-conditioned orthogonal 3d u-net for weather4cast competition.
\newblock \emph{arXiv preprint arXiv:2212.02059}.

\bibitem[{Kim et~al.(2022{\natexlab{b}})Kim, Yang, Zhang, and Hong}]{kim2022near}
Taereem Kim, Tiantian Yang, Lujun Zhang, and Yang Hong. 2022{\natexlab{b}}.
\newblock Near real-time hurricane rainfall forecasting using convolutional neural network models with integrated multi-satellite retrievals for gpm (imerg) product.
\newblock \emph{Atmospheric Research}, 270:106037.

\bibitem[{Kiranyaz et~al.(2021)Kiranyaz, Avci, Abdeljaber, Ince, Gabbouj, and Inman}]{kiranyaz20211d}
Serkan Kiranyaz, Onur Avci, Osama Abdeljaber, Turker Ince, Moncef Gabbouj, and Daniel~J Inman. 2021.
\newblock 1d convolutional neural networks and applications: A survey.
\newblock \emph{Mechanical systems and signal processing}, 151:107398.

\bibitem[{Kirschstein and Sun(2024)}]{kirschstein2024merit}
Nikolas Kirschstein and Yixuan Sun. 2024.
\newblock The merit of river network topology for neural flood forecasting.
\newblock \emph{arXiv preprint arXiv:2405.19836}.

\bibitem[{Kitamoto et~al.(2023)Kitamoto, Hwang, Vuillod, Gautier, Tian, and Clanuwat}]{kitamoto2023digital}
Asanobu Kitamoto, Jared Hwang, Bastien Vuillod, Lucas Gautier, Yingtao Tian, and Tarin Clanuwat. 2023.
\newblock Digital typhoon: Long-term satellite image dataset for the spatio-temporal modeling of tropical cyclones.
\newblock \emph{arXiv preprint arXiv:2311.02665}.

\bibitem[{Kochkov et~al.(2024)Kochkov, Yuval, Langmore, Norgaard, Smith, Mooers, Kl{\"o}wer, Lottes, Rasp, D{\"u}ben et~al.}]{kochkov2024neural}
Dmitrii Kochkov, Janni Yuval, Ian Langmore, Peter Norgaard, Jamie Smith, Griffin Mooers, Milan Kl{\"o}wer, James Lottes, Stephan Rasp, Peter D{\"u}ben, et~al. 2024.
\newblock Neural general circulation models for weather and climate.
\newblock \emph{Nature}, 632(8027):1060--1066.

\bibitem[{Kumar et~al.(2021)Kumar, Biswas, and Pandey}]{kumar2021predicting}
Sandeep Kumar, Koushik Biswas, and Ashish~Kumar Pandey. 2021.
\newblock Predicting landfall’s location and time of a tropical cyclone using reanalysis data.
\newblock In \emph{Artificial Neural Networks and Machine Learning--ICANN 2021: 30th International Conference on Artificial Neural Networks, Bratislava, Slovakia, September 14--17, 2021, Proceedings, Part IV 30}, pages 372--383. Springer.

\bibitem[{Kumar et~al.(2022)Kumar, Biswas, and Pandey}]{kumar2022forecastingformationtropicalcyclone}
Sandeep Kumar, Koushik Biswas, and Ashish~Kumar Pandey. 2022.
\newblock \href {https://arxiv.org/abs/2212.06149} {Forecasting formation of a tropical cyclone using reanalysis data}.
\newblock \emph{Preprint}, arXiv:2212.06149.

\bibitem[{Lai and Dzombak(2020)}]{lai2020use}
Yuchuan Lai and David~A Dzombak. 2020.
\newblock Use of the autoregressive integrated moving average (arima) model to forecast near-term regional temperature and precipitation.
\newblock \emph{Weather and forecasting}, 35(3):959--976.

\bibitem[{Lam et~al.(2022)Lam, Sanchez-Gonzalez, Willson, Wirnsberger, Fortunato, Pritzel, Ravuri, Ewalds, Alet, Eaton-Rosen et~al.}]{lam2022graphcast}
Remi Lam, Alvaro Sanchez-Gonzalez, Matthew Willson, Peter Wirnsberger, Meire Fortunato, Alexander Pritzel, Suman Ravuri, Timo Ewalds, Ferran Alet, Zach Eaton-Rosen, et~al. 2022.
\newblock Graphcast: Learning skillful medium-range global weather forecasting.
\newblock \emph{arXiv preprint arXiv:2212.12794}.

\bibitem[{Lang et~al.(2024)Lang, Alexe, Chantry, Dramsch, Pinault, Raoult, Clare, Lessig, Maier-Gerber, Magnusson et~al.}]{lang2024aifs}
Simon Lang, Mihai Alexe, Matthew Chantry, Jesper Dramsch, Florian Pinault, Baudouin Raoult, Mariana~CA Clare, Christian Lessig, Michael Maier-Gerber, Linus Magnusson, et~al. 2024.
\newblock Aifs-ecmwf's data-driven forecasting system.
\newblock \emph{arXiv preprint arXiv:2406.01465}.

\bibitem[{Larvor et~al.(2020)Larvor, Berthomier, Chabot, Le~Pape, Pradel, and Perez}]{larvormeteonet}
Gwenna{\"e}lle Larvor, L{\'e}a Berthomier, Vincent Chabot, Brice Le~Pape, Bruno Pradel, and Lior Perez. 2020.
\newblock Meteonet, an open reference weather dataset by meteo-france. 2020.

\bibitem[{Lebedev et~al.(2019)Lebedev, Ivashkin, Rudenko, Ganshin, Molchanov, Ovcharenko, Grokhovetskiy, Bushmarinov, and Solomentsev}]{lebedev2019precipitation}
Vadim Lebedev, Vladimir Ivashkin, Irina Rudenko, Alexander Ganshin, Alexander Molchanov, Sergey Ovcharenko, Ruslan Grokhovetskiy, Ivan Bushmarinov, and Dmitry Solomentsev. 2019.
\newblock Precipitation nowcasting with satellite imagery.
\newblock In \emph{Proceedings of the 25th ACM SIGKDD international conference on knowledge discovery \& data mining}, pages 2680--2688.

\bibitem[{LeCun et~al.(1995)LeCun, Bengio et~al.}]{lecun1995convolutional}
Yann LeCun, Yoshua Bengio, et~al. 1995.
\newblock Convolutional networks for images, speech, and time series.
\newblock \emph{The handbook of brain theory and neural networks}, 3361(10):1995.

\bibitem[{Lee and Oh(2024)}]{lee2024inducing}
Seungjun Lee and Taeil Oh. 2024.
\newblock Inducing point operator transformer: A flexible and scalable architecture for solving pdes.
\newblock In \emph{Proceedings of the AAAI Conference on Artificial Intelligence}, volume~38, pages 153--161.

\bibitem[{Leinonen et~al.(2023)Leinonen, Hamann, Nerini, Germann, and Franch}]{leinonen2023latent}
Jussi Leinonen, Ulrich Hamann, Daniele Nerini, Urs Germann, and Gabriele Franch. 2023.
\newblock Latent diffusion models for generative precipitation nowcasting with accurate uncertainty quantification.
\newblock \emph{arXiv preprint arXiv:2304.12891}.

\bibitem[{Li et~al.(2024{\natexlab{a}})Li, Wang, Wang, Lau, and Qu}]{li2024cllmate}
Haobo Li, Zhaowei Wang, Jiachen Wang, Alexis Kai~Hon Lau, and Huamin Qu. 2024{\natexlab{a}}.
\newblock Cllmate: A multimodal llm for weather and climate events forecasting.
\newblock \emph{arXiv preprint arXiv:2409.19058}.

\bibitem[{Li et~al.(2023{\natexlab{a}})Li, Carver, Lopez-Gomez, Sha, and Anderson}]{li2023seeds}
Lizao Li, Rob Carver, Ignacio Lopez-Gomez, Fei Sha, and John Anderson. 2023{\natexlab{a}}.
\newblock Seeds: Emulation of weather forecast ensembles with diffusion models.
\newblock \emph{arXiv preprint arXiv:2306.14066}.

\bibitem[{Li et~al.(2022)Li, Zhang, and Krebs}]{li2022prediction}
Peifeng Li, Jin Zhang, and Peter Krebs. 2022.
\newblock Prediction of flow based on a cnn-lstm combined deep learning approach.
\newblock \emph{Water}, 14(6):993.

\bibitem[{Li et~al.(2024{\natexlab{b}})Li, Liu, Chen, Chen, Liang, Zou, and Shi}]{li2024deepphysinet}
Wenyuan Li, Zili Liu, Keyan Chen, Hao Chen, Shunlin Liang, Zhengxia Zou, and Zhenwei Shi. 2024{\natexlab{b}}.
\newblock Deepphysinet: Bridging deep learning and atmospheric physics for accurate and continuous weather modeling.
\newblock \emph{arXiv preprint arXiv:2401.04125}.

\bibitem[{Li et~al.(2023{\natexlab{b}})Li, Zhou, Zhao, and Wen}]{li2023diffusion}
Yifan Li, Kun Zhou, Wayne~Xin Zhao, and Ji-Rong Wen. 2023{\natexlab{b}}.
\newblock Diffusion models for non-autoregressive text generation: A survey.
\newblock \emph{arXiv preprint arXiv:2303.06574}.

\bibitem[{Li et~al.(2021)Li, Liu, Yang, Peng, and Zhou}]{li2021survey}
Zewen Li, Fan Liu, Wenjie Yang, Shouheng Peng, and Jun Zhou. 2021.
\newblock A survey of convolutional neural networks: analysis, applications, and prospects.
\newblock \emph{IEEE transactions on neural networks and learning systems}, 33(12):6999--7019.

\bibitem[{Lian et~al.(2020)Lian, Dong, Zhang, Pan, and Liu}]{lian2020novel}
Jie Lian, Pingping Dong, Yuping Zhang, Jianguo Pan, and Kehao Liu. 2020.
\newblock A novel data-driven tropical cyclone track prediction model based on cnn and gru with multi-dimensional feature selection.
\newblock \emph{Ieee Access}, 8:97114--97128.

\bibitem[{Liang et~al.(2024)Liang, Sun, Shu, Li, Liu, Wei, and Yin}]{liang2024adaptive}
Chengyu Liang, Zhengya Sun, Gaojin Shu, Wenhui Li, An-An Liu, Zhiqiang Wei, and Bo~Yin. 2024.
\newblock Adaptive graph spatial-temporal attention networks for long lead enso prediction.
\newblock \emph{Expert Systems with Applications}, page 124492.

\bibitem[{Liang et~al.(2023)Liang, Xia, Ke, Wang, Wen, Zhang, Zheng, and Zimmermann}]{liang2023airformer}
Yuxuan Liang, Yutong Xia, Songyu Ke, Yiwei Wang, Qingsong Wen, Junbo Zhang, Yu~Zheng, and Roger Zimmermann. 2023.
\newblock Airformer: Predicting nationwide air quality in china with transformers.
\newblock In \emph{Proceedings of the AAAI Conference on Artificial Intelligence}, volume~37, pages 14329--14337.

\bibitem[{Ling et~al.(2024{\natexlab{a}})Ling, Lu, Luo, Bai, Behera, Jin, Pan, Jiang, and Yamagata}]{ling2024diffusion}
Fenghua Ling, Zeyu Lu, Jing-Jia Luo, Lei Bai, Swadhin~K Behera, Dachao Jin, Baoxiang Pan, Huidong Jiang, and Toshio Yamagata. 2024{\natexlab{a}}.
\newblock Diffusion model-based probabilistic downscaling for 180-year east asian climate reconstruction.
\newblock \emph{npj Climate and Atmospheric Science}, 7(1):131.

\bibitem[{Ling et~al.(2024{\natexlab{b}})Ling, Li, Qin, Yang, and Huang}]{ling2024srndiff}
Xudong Ling, Chaorong Li, Fengqing Qin, Peng Yang, and Yuanyuan Huang. 2024{\natexlab{b}}.
\newblock Srndiff: Short-term rainfall nowcasting with condition diffusion model.
\newblock \emph{arXiv preprint arXiv:2402.13737}.

\bibitem[{Liu and Lee(2020)}]{liu2020mpl}
Hong-Bin Liu and Ickjai Lee. 2020.
\newblock Mpl-gan: Toward realistic meteorological predictive learning using conditional gan.
\newblock \emph{IEEE Access}, 8:93179--93186.

\bibitem[{Liu et~al.(2024{\natexlab{a}})Liu, Yang, Li, and Hong}]{liu2024retrieval}
Jingwei Liu, Ling Yang, Hongyan Li, and Shenda Hong. 2024{\natexlab{a}}.
\newblock Retrieval-augmented diffusion models for time series forecasting.
\newblock \emph{arXiv preprint arXiv:2410.18712}.

\bibitem[{Liu et~al.(2021)Liu, Tang, Wu, Li, Wang, and Chen}]{liu2021forecasting}
Jun Liu, Youmin Tang, Yanling Wu, Tang Li, Qiang Wang, and Dake Chen. 2021.
\newblock Forecasting the indian ocean dipole with deep learning techniques.
\newblock \emph{Geophysical Research Letters}, 48(20):e2021GL094407.

\bibitem[{Liu et~al.(2024{\natexlab{b}})Liu, Zhou, Sun, and Jin}]{liu2024mitigating}
Peiyuan Liu, Tian Zhou, Liang Sun, and Rong Jin. 2024{\natexlab{b}}.
\newblock Mitigating time discretization challenges with weatherode: A sandwich physics-driven neural ode for weather forecasting.
\newblock \emph{arXiv preprint arXiv:2410.06560}.

\bibitem[{Liu et~al.(2024{\natexlab{c}})Liu, Gong, Zhuang, Zhong, Kang, and Li}]{liu2024compressing}
Qian Liu, Bing Gong, Xiaoran Zhuang, Xiaohui Zhong, Zhiming Kang, and Hao Li. 2024{\natexlab{c}}.
\newblock Compressing high-resolution data through latent representation encoding for downscaling large-scale ai weather forecast model.
\newblock \emph{arXiv preprint arXiv:2410.09109}.

\bibitem[{Liu et~al.(2023{\natexlab{a}})Liu, Hu, Zhang, Wu, Wang, Ma, and Long}]{liu2023itransformer}
Yong Liu, Tengge Hu, Haoran Zhang, Haixu Wu, Shiyu Wang, Lintao Ma, and Mingsheng Long. 2023{\natexlab{a}}.
\newblock itransformer: Inverted transformers are effective for time series forecasting.
\newblock \emph{arXiv preprint arXiv:2310.06625}.

\bibitem[{Liu et~al.(2024{\natexlab{d}})Liu, Zhang, Li, Huang, Wang, and Long}]{liu2024timer}
Yong Liu, Haoran Zhang, Chenyu Li, Xiangdong Huang, Jianmin Wang, and Mingsheng Long. 2024{\natexlab{d}}.
\newblock Timer: Generative pre-trained transformers are large time series models.
\newblock In \emph{Forty-first International Conference on Machine Learning}.

\bibitem[{Liu et~al.(2023{\natexlab{b}})Liu, Duffy, Dy, and Ganguly}]{liu2023explainable}
Yumin Liu, Kate Duffy, Jennifer~G Dy, and Auroop~R Ganguly. 2023{\natexlab{b}}.
\newblock Explainable deep learning for insights in el ni{\~n}o and river flows.
\newblock \emph{Nature Communications}, 14(1):339.

\bibitem[{Liu et~al.(2022)Liu, Hu, Lin, Yao, Xie, Wei, Ning, Cao, Zhang, Dong et~al.}]{liu2022swin}
Ze~Liu, Han Hu, Yutong Lin, Zhuliang Yao, Zhenda Xie, Yixuan Wei, Jia Ning, Yue Cao, Zheng Zhang, Li~Dong, et~al. 2022.
\newblock Swin transformer v2: Scaling up capacity and resolution.
\newblock In \emph{Proceedings of the IEEE/CVF conference on computer vision and pattern recognition}, pages 12009--12019.

\bibitem[{Liu et~al.(2024{\natexlab{e}})Liu, Wang, Shi, Zheng, Chen, Song, Dong, Obeysekera, Shirani, and Luo}]{liu2024timex}
Zichuan Liu, Tianchun Wang, Jimeng Shi, Xu~Zheng, Zhuomin Chen, Lei Song, Wenqian Dong, Jayantha Obeysekera, Farhad Shirani, and Dongsheng Luo. 2024{\natexlab{e}}.
\newblock Timex++: Learning time-series explanations with information bottleneck.
\newblock \emph{arXiv preprint arXiv:2405.09308}.

\bibitem[{Liu et~al.(2024{\natexlab{f}})Liu, Chen, Bai, Li, Ouyang, Zou, and Shi}]{liu2024mambads}
Zili Liu, Hao Chen, Lei Bai, Wenyuan Li, Wanli Ouyang, Zhengxia Zou, and Zhenwei Shi. 2024{\natexlab{f}}.
\newblock Mambads: Near-surface meteorological field downscaling with topography constrained selective state space modeling.
\newblock \emph{arXiv preprint arXiv:2408.10854}.

\bibitem[{Lu et~al.(2022)Lu, Wang, Zhang, Yu, He, Zhang, and Li}]{lu2022vision}
Mingyue Lu, Menglong Wang, Qian Zhang, Manzhu Yu, Caifen He, Yadong Zhang, and Yuchen Li. 2022.
\newblock A vision transformer for lightning intensity estimation using 3d weather radar.
\newblock \emph{Science of the total environment}, 853:158496.

\bibitem[{Lundberg(2017)}]{lundberg2017unified}
Scott Lundberg. 2017.
\newblock A unified approach to interpreting model predictions.
\newblock \emph{arXiv preprint arXiv:1705.07874}.

\bibitem[{Luo et~al.(2022)Luo, Li, Ye, Feng, and Ng}]{luo2022experimental}
Chuyao Luo, Xutao Li, Yunming Ye, Shanshan Feng, and Michael~K Ng. 2022.
\newblock Experimental study on generative adversarial network for precipitation nowcasting.
\newblock \emph{IEEE Transactions on Geoscience and Remote Sensing}, 60:1--20.

\bibitem[{Ma et~al.(2023{\natexlab{a}})Ma, Xie, Teng, Wang, Ji, Zhang, and Li}]{ma2023histgnn}
Minbo Ma, Peng Xie, Fei Teng, Bin Wang, Shenggong Ji, Junbo Zhang, and Tianrui Li. 2023{\natexlab{a}}.
\newblock Histgnn: Hierarchical spatio-temporal graph neural network for weather forecasting.
\newblock \emph{Information Sciences}, 648:119580.

\bibitem[{Ma et~al.(2023{\natexlab{b}})Ma, Zhang, and Liu}]{ma2023mm}
Zhifeng Ma, Hao Zhang, and Jie Liu. 2023{\natexlab{b}}.
\newblock Mm-rnn: A multimodal rnn for precipitation nowcasting.
\newblock \emph{IEEE Transactions on Geoscience and Remote Sensing}.

\bibitem[{Man et~al.(2023)Man, Zhang, Li, and Shao}]{man2023w}
Xin Man, Chenghong Zhang, Changyu Li, and Jie Shao. 2023.
\newblock W-mae: Pre-trained weather model with masked autoencoder for multi-variable weather forecasting.
\newblock \emph{arXiv preprint arXiv:2304.08754}.

\bibitem[{Manshausen et~al.(2024)Manshausen, Cohen, Pathak, Pritchard, Garg, Mardani, Kashinath, Byrne, and Brenowitz}]{manshausen2024generative}
Peter Manshausen, Yair Cohen, Jaideep Pathak, Mike Pritchard, Piyush Garg, Morteza Mardani, Karthik Kashinath, Simon Byrne, and Noah Brenowitz. 2024.
\newblock Generative data assimilation of sparse weather station observations at kilometer scales.
\newblock \emph{arXiv preprint arXiv:2406.16947}.

\bibitem[{Marchuk(2012)}]{marchuk2012numerical}
Gurii Marchuk. 2012.
\newblock \emph{Numerical methods in weather prediction}.
\newblock Elsevier.

\bibitem[{Mardani et~al.(2023)Mardani, Brenowitz, Cohen, Pathak, Chen, Liu, Vahdat, Kashinath, Kautz, and Pritchard}]{mardani2023generative}
Morteza Mardani, Noah Brenowitz, Yair Cohen, Jaideep Pathak, Chieh-Yu Chen, Cheng-Chin Liu, Arash Vahdat, Karthik Kashinath, Jan Kautz, and Mike Pritchard. 2023.
\newblock Generative residual diffusion modeling for km-scale atmospheric downscaling.
\newblock \emph{arXiv preprint arXiv:2309.15214}.

\bibitem[{Marjani and Mesgari(2023)}]{marjani2023large}
M~Marjani and MS~Mesgari. 2023.
\newblock The large-scale wildfire spread prediction using a multi-kernel convolutional neural network.
\newblock \emph{ISPRS Annals of the Photogrammetry, Remote Sensing and Spatial Information Sciences}, 10:483--488.

\bibitem[{Marjani et~al.(2023)Marjani, Ahmadi, and Mahdianpari}]{marjani2023firepred}
Mohammad Marjani, Seyed~Ali Ahmadi, and Masoud Mahdianpari. 2023.
\newblock Firepred: A hybrid multi-temporal convolutional neural network model for wildfire spread prediction.
\newblock \emph{Ecological Informatics}, 78:102282.

\bibitem[{Marjani et~al.(2024)Marjani, Mahdianpari, and Mohammadimanesh}]{marjani2024cnn}
Mohammad Marjani, Masoud Mahdianpari, and Fariba Mohammadimanesh. 2024.
\newblock Cnn-bilstm: A novel deep learning model for near-real-time daily wildfire spread prediction.
\newblock \emph{Remote Sensing}, 16(8):1467.

\bibitem[{Masrur and Yu(2023)}]{masrur2023spatiotemporal}
Arif Masrur and Manzhu Yu. 2023.
\newblock Spatiotemporal attention convlstm networks for predicting and physically interpreting wildfire spread.
\newblock In \emph{Artificial intelligence in earth science}, pages 119--156. Elsevier.

\bibitem[{Masrur et~al.(2024)Masrur, Yu, and Taylor}]{masrur2024capturing}
Arif Masrur, Manzhu Yu, and Alan Taylor. 2024.
\newblock Capturing and interpreting wildfire spread dynamics: attention-based spatiotemporal models using convlstm networks.
\newblock \emph{Ecological Informatics}, 82:102760.

\bibitem[{Materia et~al.(2023)Materia, Garc{\'\i}a, van Straaten, Mamalakis, Cavicchia, Coumou, De~Luca, Kretschmer, Donat et~al.}]{materia2023artificial}
Stefano Materia, Llu{\'\i}s~Palma Garc{\'\i}a, Chiem van Straaten, Antonios Mamalakis, Leone Cavicchia, Dim Coumou, Paolo De~Luca, Marlene Kretschmer, Markus~G Donat, et~al. 2023.
\newblock Artificial intelligence for prediction of climate extremes: State of the art, challenges and future perspectives.
\newblock \emph{arXiv preprint arXiv:2310.01944}.

\bibitem[{Medsker and Jain(2001)}]{medsker2001recurrent}
Larry~R Medsker and LC~Jain. 2001.
\newblock Recurrent neural networks.
\newblock \emph{Design and Applications}, 5(64-67):2.

\bibitem[{Meng et~al.(2023)Meng, Gao, Rigall, Dong, Dong, and Du}]{meng2023physical}
Yuxin Meng, Feng Gao, Eric Rigall, Ran Dong, Junyu Dong, and Qian Du. 2023.
\newblock Physical knowledge-enhanced deep neural network for sea surface temperature prediction.
\newblock \emph{IEEE Transactions on Geoscience and Remote Sensing}, 61:1--13.

\bibitem[{Meng et~al.(2021)Meng, Rigall, Chen, Gao, Dong, and Chen}]{meng2021physics}
Yuxin Meng, Eric Rigall, Xueen Chen, Feng Gao, Junyu Dong, and Sheng Chen. 2021.
\newblock Physics-guided generative adversarial networks for sea subsurface temperature prediction.
\newblock \emph{IEEE transactions on neural networks and learning systems}.

\bibitem[{Meo et~al.(2024)Meo, Roy, Lic{\u{a}}, Yin, Che, Wang, Imhoff, Uijlenhoet, and Dauwels}]{meo2024extreme}
Cristian Meo, Ankush Roy, Mircea Lic{\u{a}}, Junzhe Yin, Zeineb~Bou Che, Yanbo Wang, Ruben Imhoff, Remko Uijlenhoet, and Justin Dauwels. 2024.
\newblock Extreme precipitation nowcasting using transformer-based generative models.
\newblock \emph{arXiv preprint arXiv:2403.03929}.

\bibitem[{Miao et~al.(2023)Miao, Li, Mu, He, Ma, Chen, Wei, and Gao}]{miao2023time}
Xinyu Miao, Jian Li, Yunjie Mu, Cheng He, Yunfei Ma, Jie Chen, Wentao Wei, and Demin Gao. 2023.
\newblock Time series forest fire prediction based on improved transformer.
\newblock \emph{Forests}, 14(8):1596.

\bibitem[{Mikolov et~al.(2011)Mikolov, Kombrink, Burget, {\v{C}}ernock{\`y}, and Khudanpur}]{mikolov2011extensions}
Tom{\'a}{\v{s}} Mikolov, Stefan Kombrink, Luk{\'a}{\v{s}} Burget, Jan {\v{C}}ernock{\`y}, and Sanjeev Khudanpur. 2011.
\newblock Extensions of recurrent neural network language model.
\newblock In \emph{2011 IEEE international conference on acoustics, speech and signal processing (ICASSP)}, pages 5528--5531. IEEE.

\bibitem[{Miller et~al.(2024)Miller, Aldosari, Saeed, Barna, Rana, Arpinar, and Liu}]{miller2024survey}
John~A Miller, Mohammed Aldosari, Farah Saeed, Nasid~Habib Barna, Subas Rana, I~Budak Arpinar, and Ninghao Liu. 2024.
\newblock A survey of deep learning and foundation models for time series forecasting.
\newblock \emph{arXiv preprint arXiv:2401.13912}.

\bibitem[{Mir et~al.(2024)Mir, Arbab et~al.}]{mir2024enso}
Shabana Mir, Masood~Ahmad Arbab, et~al. 2024.
\newblock Enso dataset \& comparison of deep learning models for enso forecasting.
\newblock \emph{Earth Science Informatics}, 17(3):2623--2628.

\bibitem[{Mirza(2014)}]{mirza2014conditional}
Mehdi Mirza. 2014.
\newblock Conditional generative adversarial nets.
\newblock \emph{arXiv preprint arXiv:1411.1784}.

\bibitem[{Moishin et~al.(2021)Moishin, Deo, Prasad, Raj, and Abdulla}]{moishin2021designing}
Mohammed Moishin, Ravinesh~C Deo, Ramendra Prasad, Nawin Raj, and Shahab Abdulla. 2021.
\newblock Designing deep-based learning flood forecast model with convlstm hybrid algorithm.
\newblock \emph{IEEE Access}, 9:50982--50993.

\bibitem[{Molina et~al.(2023)Molina, O’Brien, Anderson, Ashfaq, Bennett, Collins, Dagon, Restrepo, and Ullrich}]{molina2023review}
Maria~J Molina, Travis~A O’Brien, Gemma Anderson, Moetasim Ashfaq, Katrina~E Bennett, William~D Collins, Katherine Dagon, Juan~M Restrepo, and Paul~A Ullrich. 2023.
\newblock A review of recent and emerging machine learning applications for climate variability and weather phenomena.
\newblock \emph{Artificial Intelligence for the Earth Systems}, pages 1--46.

\bibitem[{Mu et~al.(2021)Mu, Qin, and Yuan}]{mu2021enso}
Bin Mu, Bo~Qin, and Shijin Yuan. 2021.
\newblock Enso-asc 1.0. 0: Enso deep learning forecast model with a multivariate air--sea coupler.
\newblock \emph{Geoscientific Model Development}, 14(11):6977--6999.

\bibitem[{Mukkavilli et~al.(2023)Mukkavilli, Civitarese, Schmude, Jakubik, Jones, Nguyen, Phillips, Roy, Singh, Watson et~al.}]{mukkavilli2023ai}
S~Karthik Mukkavilli, Daniel~Salles Civitarese, Johannes Schmude, Johannes Jakubik, Anne Jones, Nam Nguyen, Christopher Phillips, Sujit Roy, Shraddha Singh, Campbell Watson, et~al. 2023.
\newblock Ai foundation models for weather and climate: Applications, design, and implementation.
\newblock \emph{arXiv preprint arXiv:2309.10808}.

\bibitem[{Nath et~al.(2023)Nath, Shukla, Wang, and Quilodr{\'a}n-Casas}]{nath2023forecasting}
Pritthijit Nath, Pancham Shukla, Shuai Wang, and C{\'e}sar Quilodr{\'a}n-Casas. 2023.
\newblock Forecasting tropical cyclones with cascaded diffusion models.
\newblock \emph{arXiv preprint arXiv:2310.01690}.

\bibitem[{Nevo et~al.(2022)Nevo, Morin, Gerzi~Rosenthal, Metzger, Barshai, Weitzner, Voloshin, Kratzert, Elidan, Dror et~al.}]{nevo2022flood}
Sella Nevo, Efrat Morin, Adi Gerzi~Rosenthal, Asher Metzger, Chen Barshai, Dana Weitzner, Dafi Voloshin, Frederik Kratzert, Gal Elidan, Gideon Dror, et~al. 2022.
\newblock Flood forecasting with machine learning models in an operational framework.
\newblock \emph{Hydrology and Earth System Sciences}, 26(15):4013--4032.

\bibitem[{Nguyen and Kieu(2024)}]{nguyen2024predicting}
Quan Nguyen and Chanh Kieu. 2024.
\newblock Predicting tropical cyclone formation with deep learning.
\newblock \emph{Weather and Forecasting}, 39(1):241--258.

\bibitem[{Nguyen et~al.(2023{\natexlab{a}})Nguyen, Brandstetter, Kapoor, Gupta, and Grover}]{nguyen2023climax}
Tung Nguyen, Johannes Brandstetter, Ashish Kapoor, Jayesh~K Gupta, and Aditya Grover. 2023{\natexlab{a}}.
\newblock Climax: A foundation model for weather and climate.
\newblock \emph{arXiv preprint arXiv:2301.10343}.

\bibitem[{Nguyen et~al.(2023{\natexlab{b}})Nguyen, Jewik, Bansal, Sharma, and Grover}]{nguyen2023climatelearn}
Tung Nguyen, Jason Jewik, Hritik Bansal, Prakhar Sharma, and Aditya Grover. 2023{\natexlab{b}}.
\newblock Climatelearn: Benchmarking machine learning for weather and climate modeling.
\newblock \emph{arXiv preprint arXiv:2307.01909}.

\bibitem[{Nguyen et~al.(2023{\natexlab{c}})Nguyen, Shah, Bansal, Arcomano, Maulik, Kotamarthi, Foster, Madireddy, and Grover}]{nguyen2023scaling}
Tung Nguyen, Rohan Shah, Hritik Bansal, Troy Arcomano, Romit Maulik, Veerabhadra Kotamarthi, Ian Foster, Sandeep Madireddy, and Aditya Grover. 2023{\natexlab{c}}.
\newblock Scaling transformer neural networks for skillful and reliable medium-range weather forecasting.
\newblock \emph{arXiv preprint arXiv:2312.03876}.

\bibitem[{Nguyen et~al.(2024)Nguyen, Sinha, Deepak, McKinnon, and Grover}]{nguyen2024atmosarena}
Tung Nguyen, Prateik Sinha, Advit Deepak, Karen~A. McKinnon, and Aditya Grover. 2024.
\newblock \href {https://www.climatechange.ai/papers/neurips2024/19} {Atmosarena: Benchmarking foundation models for atmospheric sciences}.
\newblock In \emph{NeurIPS 2024 Workshop on Tackling Climate Change with Machine Learning}.

\bibitem[{Nie et~al.(2022)Nie, Nguyen, Sinthong, and Kalagnanam}]{nie2022time}
Yuqi Nie, Nam~H Nguyen, Phanwadee Sinthong, and Jayant Kalagnanam. 2022.
\newblock A time series is worth 64 words: Long-term forecasting with transformers.
\newblock \emph{arXiv preprint arXiv:2211.14730}.

\bibitem[{Nyamane et~al.(2024)Nyamane, Abd~Elbasit, and Obagbuwa}]{nyamane2024harnessing}
Seipati Nyamane, Mohamed~AM Abd~Elbasit, and Ibidun~Christiana Obagbuwa. 2024.
\newblock Harnessing deep learning for meteorological drought forecasts in the northern cape, south africa.
\newblock \emph{International Journal of Intelligent Systems}, 2024(1):7562587.

\bibitem[{Ou et~al.(2024)Ou, Nai, Pan, Pan, Shen, Jiang, Liu, Tang, Li, and Zheng}]{ou2024drum}
Zhigang Ou, Congyi Nai, Baoxiang Pan, Ming Pan, Chaopeng Shen, Peishi Jiang, Xingcai Liu, Qiuhong Tang, Wenqing Li, and Yi~Zheng. 2024.
\newblock Drum: Diffusion-based runoff model for probabilistic flood forecasting.
\newblock \emph{arXiv preprint arXiv:2412.11942}.

\bibitem[{Palmer(2019)}]{palmer2019stochastic}
TN~Palmer. 2019.
\newblock Stochastic weather and climate models.
\newblock \emph{Nature Reviews Physics}, 1(7):463--471.

\bibitem[{Palmer et~al.(2005)Palmer, Shutts, Hagedorn, Doblas-Reyes, Jung, and Leutbecher}]{palmer2005representing}
TN~Palmer, GJ~Shutts, R~Hagedorn, FJ~Doblas-Reyes, Thomas Jung, and M~Leutbecher. 2005.
\newblock Representing model uncertainty in weather and climate prediction.
\newblock \emph{Annu. Rev. Earth Planet. Sci.}, 33(1):163--193.

\bibitem[{Park et~al.(2022)Park, Lee, Son, Cho, and Kim}]{park2022nowformer}
Jinyoung Park, Inyoung Lee, Minseok Son, Seungju Cho, and Changick Kim. 2022.
\newblock Nowformer: A locally enhanced temporal learner for precipitation nowcasting.
\newblock In \emph{Proceedings of the NeurIPS 2022 Workshop on Tackling Climate Change with Machine Learning}.

\bibitem[{Park et~al.(2020)Park, Im, Han, and Rhee}]{park2020short}
Sumin Park, Jungho Im, Daehyeon Han, and Jinyoung Rhee. 2020.
\newblock Short-term forecasting of satellite-based drought indices using their temporal patterns and numerical model output.
\newblock \emph{Remote Sensing}, 12(21):3499.

\bibitem[{Pathak et~al.(2022)Pathak, Subramanian, Harrington, Raja, Chattopadhyay, Mardani, Kurth, Hall, Li, Azizzadenesheli et~al.}]{pathak2022fourcastnet}
Jaideep Pathak, Shashank Subramanian, Peter Harrington, Sanjeev Raja, Ashesh Chattopadhyay, Morteza Mardani, Thorsten Kurth, David Hall, Zongyi Li, Kamyar Azizzadenesheli, et~al. 2022.
\newblock Fourcastnet: A global data-driven high-resolution weather model using adaptive fourier neural operators.
\newblock \emph{arXiv preprint arXiv:2202.11214}.

\bibitem[{Perumal and Van~Zyl(2020)}]{perumal2020comparison}
Rylan Perumal and Terence~L Van~Zyl. 2020.
\newblock Comparison of recurrent neural network architectures for wildfire spread modelling.
\newblock In \emph{2020 International SAUPEC/RobMech/PRASA Conference}, pages 1--6. IEEE.

\bibitem[{Poornima and Pushpalatha(2019)}]{poornima2019drought}
S~Poornima and M~Pushpalatha. 2019.
\newblock Drought prediction based on spi and spei with varying timescales using lstm recurrent neural network.
\newblock \emph{Soft Computing}, 23(18):8399--8412.

\bibitem[{Price et~al.(2023)Price, Sanchez-Gonzalez, Alet, Andersson, El-Kadi, Masters, Ewalds, Stott, Mohamed, Battaglia et~al.}]{price2023gencast}
Ilan Price, Alvaro Sanchez-Gonzalez, Ferran Alet, Tom~R Andersson, Andrew El-Kadi, Dominic Masters, Timo Ewalds, Jacklynn Stott, Shakir Mohamed, Peter Battaglia, et~al. 2023.
\newblock Gencast: Diffusion-based ensemble forecasting for medium-range weather.
\newblock \emph{arXiv preprint arXiv:2312.15796}.

\bibitem[{Qin et~al.(2024)Qin, Chen, Jiang, Sun, Ye, and Lin}]{qin2024metmamba}
Haoyu Qin, Yungang Chen, Qianchuan Jiang, Pengchao Sun, Xiancai Ye, and Chao Lin. 2024.
\newblock Metmamba: Regional weather forecasting with spatial-temporal mamba model.
\newblock \emph{arXiv preprint arXiv:2408.06400}.

\bibitem[{Qu et~al.(2024)Qu, Nathaniel, Li, and Gentine}]{qu2024deep}
Yongquan Qu, Juan Nathaniel, Shuolin Li, and Pierre Gentine. 2024.
\newblock Deep generative data assimilation in multimodal setting.
\newblock In \emph{Proceedings of the IEEE/CVF Conference on Computer Vision and Pattern Recognition}, pages 449--459.

\bibitem[{Rabier et~al.(1998)Rabier, Th{\'e}paut, and Courtier}]{rabier1998extended}
Florence Rabier, Jean-Noel Th{\'e}paut, and Philippe Courtier. 1998.
\newblock Extended assimilation and forecast experiments with a four-dimensional variational assimilation system.
\newblock \emph{Quarterly Journal of the Royal Meteorological Society}, 124(550):1861--1887.

\bibitem[{Racah et~al.(2017)Racah, Beckham, Maharaj, Ebrahimi~Kahou, Prabhat, and Pal}]{racah2017extremeweather}
Evan Racah, Christopher Beckham, Tegan Maharaj, Samira Ebrahimi~Kahou, Mr~Prabhat, and Chris Pal. 2017.
\newblock Extremeweather: A large-scale climate dataset for semi-supervised detection, localization, and understanding of extreme weather events.
\newblock \emph{Advances in neural information processing systems}, 30.

\bibitem[{Ramavajjala(2024)}]{ramavajjala2024heal}
Vivek Ramavajjala. 2024.
\newblock Heal-vit: Vision transformers on a spherical mesh for medium-range weather forecasting.
\newblock \emph{arXiv preprint arXiv:2403.17016}.

\bibitem[{Ran et~al.(2024)Ran, Xiao, Wang, Shi, Lin, Meng, and Allmendinger}]{ran2024hr}
Nian Ran, Peng Xiao, Yue Wang, Wesley Shi, Jianxin Lin, Qi~Meng, and Richard Allmendinger. 2024.
\newblock Hr-extreme: A high-resolution dataset for extreme weather forecasting.
\newblock \emph{arXiv preprint arXiv:2409.18885}.

\bibitem[{Rasp et~al.(2020)Rasp, Dueben, Scher, Weyn, Mouatadid, and Thuerey}]{rasp2020weatherbench}
Stephan Rasp, Peter~D Dueben, Sebastian Scher, Jonathan~A Weyn, Soukayna Mouatadid, and Nils Thuerey. 2020.
\newblock Weatherbench: a benchmark data set for data-driven weather forecasting.
\newblock \emph{Journal of Advances in Modeling Earth Systems}, 12(11):e2020MS002203.

\bibitem[{Rasp et~al.(2023)Rasp, Hoyer, Merose, Langmore, Battaglia, Russel, Sanchez-Gonzalez, Yang, Carver, Agrawal et~al.}]{rasp2023weatherbench}
Stephan Rasp, Stephan Hoyer, Alexander Merose, Ian Langmore, Peter Battaglia, Tyler Russel, Alvaro Sanchez-Gonzalez, Vivian Yang, Rob Carver, Shreya Agrawal, et~al. 2023.
\newblock Weatherbench 2: A benchmark for the next generation of data-driven global weather models.
\newblock \emph{arXiv preprint arXiv:2308.15560}.

\bibitem[{Rasul et~al.(2021)Rasul, Seward, Schuster, and Vollgraf}]{rasul2021autoregressive}
Kashif Rasul, Calvin Seward, Ingmar Schuster, and Roland Vollgraf. 2021.
\newblock Autoregressive denoising diffusion models for multivariate probabilistic time series forecasting.
\newblock In \emph{International Conference on Machine Learning}, pages 8857--8868. PMLR.

\bibitem[{Ravindra et~al.(2019)Ravindra, Rattan, Mor, and Aggarwal}]{ravindra2019generalized}
Khaiwal Ravindra, Preety Rattan, Suman Mor, and Ashutosh~Nath Aggarwal. 2019.
\newblock Generalized additive models: Building evidence of air pollution, climate change and human health.
\newblock \emph{Environment international}, 132:104987.

\bibitem[{Ravuri et~al.(2021)Ravuri, Lenc, Willson, Kangin, Lam, Mirowski, Fitzsimons, Athanassiadou, Kashem, Madge et~al.}]{ravuri2021skilful}
Suman Ravuri, Karel Lenc, Matthew Willson, Dmitry Kangin, Remi Lam, Piotr Mirowski, Megan Fitzsimons, Maria Athanassiadou, Sheleem Kashem, Sam Madge, et~al. 2021.
\newblock Skilful precipitation nowcasting using deep generative models of radar.
\newblock \emph{Nature}, 597(7878):672--677.

\bibitem[{Ravuru et~al.(2024)Ravuru, Sakhinana, and Runkana}]{ravuru2024agentic}
Chidaksh Ravuru, Sagar~Srinivas Sakhinana, and Venkataramana Runkana. 2024.
\newblock Agentic retrieval-augmented generation for time series analysis.
\newblock \emph{arXiv preprint arXiv:2408.14484}.

\bibitem[{Ren et~al.(2016)Ren, He, Girshick, and Sun}]{ren2016faster}
Shaoqing Ren, Kaiming He, Ross Girshick, and Jian Sun. 2016.
\newblock Faster r-cnn: Towards real-time object detection with region proposal networks.
\newblock \emph{IEEE transactions on pattern analysis and machine intelligence}, 39(6):1137--1149.

\bibitem[{Ren et~al.(2021)Ren, Li, Ren, Song, Xu, Deng, and Wang}]{ren2021deep}
Xiaoli Ren, Xiaoyong Li, Kaijun Ren, Junqiang Song, Zichen Xu, Kefeng Deng, and Xiang Wang. 2021.
\newblock Deep learning-based weather prediction: a survey.
\newblock \emph{Big Data Research}, 23:100178.

\bibitem[{Ribeiro et~al.(2016)Ribeiro, Singh, and Guestrin}]{ribeiro2016should}
Marco~Tulio Ribeiro, Sameer Singh, and Carlos Guestrin. 2016.
\newblock " why should i trust you?" explaining the predictions of any classifier.
\newblock In \emph{Proceedings of the 22nd ACM SIGKDD international conference on knowledge discovery and data mining}, pages 1135--1144.

\bibitem[{Rombach et~al.(2022)Rombach, Blattmann, Lorenz, Esser, and Ommer}]{rombach2022high}
Robin Rombach, Andreas Blattmann, Dominik Lorenz, Patrick Esser, and Bj{\"o}rn Ommer. 2022.
\newblock High-resolution image synthesis with latent diffusion models.
\newblock In \emph{Proceedings of the IEEE/CVF conference on computer vision and pattern recognition}, pages 10684--10695.

\bibitem[{Ronneberger et~al.(2015)Ronneberger, Fischer, and Brox}]{ronneberger2015u}
Olaf Ronneberger, Philipp Fischer, and Thomas Brox. 2015.
\newblock U-net: Convolutional networks for biomedical image segmentation.
\newblock In \emph{Medical image computing and computer-assisted intervention--MICCAI 2015: 18th international conference, Munich, Germany, October 5-9, 2015, proceedings, part III 18}, pages 234--241. Springer.

\bibitem[{R{\"u}hling~Cachay et~al.(2024)R{\"u}hling~Cachay, Zhao, Joren, and Yu}]{ruhling2024dyffusion}
Salva R{\"u}hling~Cachay, Bo~Zhao, Hailey Joren, and Rose Yu. 2024.
\newblock Dyffusion: A dynamics-informed diffusion model for spatiotemporal forecasting.
\newblock \emph{Advances in Neural Information Processing Systems}, 36.

\bibitem[{Ruma et~al.(2023)Ruma, Adnan, Dewan, and Rahman}]{ruma2023particle}
Jannatul~Ferdous Ruma, Mohammed Sarfaraz~Gani Adnan, Ashraf Dewan, and Rashedur~M Rahman. 2023.
\newblock Particle swarm optimization based lstm networks for water level forecasting: A case study on bangladesh river network.
\newblock \emph{Results in Engineering}, 17:100951.

\bibitem[{Saharia et~al.(2022)Saharia, Chan, Chang, Lee, Ho, Salimans, Fleet, and Norouzi}]{saharia2022palette}
Chitwan Saharia, William Chan, Huiwen Chang, Chris Lee, Jonathan Ho, Tim Salimans, David Fleet, and Mohammad Norouzi. 2022.
\newblock Palette: Image-to-image diffusion models.
\newblock In \emph{ACM SIGGRAPH 2022 Conference Proceedings}, pages 1--10.

\bibitem[{Salcedo-Sanz et~al.(2024)Salcedo-Sanz, P{\'e}rez-Aracil, Ascenso, Del~Ser, Casillas-P{\'e}rez, Kadow, Fister, Barriopedro, Garc{\'\i}a-Herrera, Giuliani et~al.}]{salcedo2024analysis}
Sancho Salcedo-Sanz, Jorge P{\'e}rez-Aracil, Guido Ascenso, Javier Del~Ser, David Casillas-P{\'e}rez, Christopher Kadow, Du{\v{s}}an Fister, David Barriopedro, Ricardo Garc{\'\i}a-Herrera, Matteo Giuliani, et~al. 2024.
\newblock Analysis, characterization, prediction, and attribution of extreme atmospheric events with machine learning and deep learning techniques: a review.
\newblock \emph{Theoretical and Applied Climatology}, 155(1):1--44.

\bibitem[{Saleem et~al.(2024)Saleem, Salim, and Purcell}]{saleem2024conformer}
Hira Saleem, Flora Salim, and Cormac Purcell. 2024.
\newblock Conformer: Embedding continuous attention in vision transformer for weather forecasting.
\newblock \emph{arXiv preprint arXiv:2402.17966}.

\bibitem[{Sbrana et~al.(2020)Sbrana, Rossi, and Naldi}]{sbrana2020n}
Attilio Sbrana, Andr{\'e} Luis~Debiaso Rossi, and Murilo~Coelho Naldi. 2020.
\newblock N-beats-rnn: deep learning for time series forecasting.
\newblock In \emph{2020 19th IEEE International Conference on Machine Learning and Applications (ICMLA)}, pages 765--768. IEEE.

\bibitem[{Scarselli et~al.(2008)Scarselli, Gori, Tsoi, Hagenbuchner, and Monfardini}]{scarselli2008graph}
Franco Scarselli, Marco Gori, Ah~Chung Tsoi, Markus Hagenbuchner, and Gabriele Monfardini. 2008.
\newblock The graph neural network model.
\newblock \emph{IEEE transactions on neural networks}, 20(1):61--80.

\bibitem[{Schmalfuss et~al.(2023)Schmalfuss, Mehl, and Bruhn}]{schmalfuss2023distracting}
Jenny Schmalfuss, Lukas Mehl, and Andr{\'e}s Bruhn. 2023.
\newblock Distracting downpour: Adversarial weather attacks for motion estimation.
\newblock In \emph{Proceedings of the IEEE/CVF International Conference on Computer Vision}, pages 10106--10116.

\bibitem[{Schmude et~al.(2024)Schmude, Roy, Trojak, Jakubik, Civitarese, Singh, Kuehnert, Ankur, Gupta, Phillips et~al.}]{schmude2024prithvi}
Johannes Schmude, Sujit Roy, Will Trojak, Johannes Jakubik, Daniel~Salles Civitarese, Shraddha Singh, Julian Kuehnert, Kumar Ankur, Aman Gupta, Christopher~E Phillips, et~al. 2024.
\newblock Prithvi wxc: Foundation model for weather and climate.
\newblock \emph{arXiv preprint arXiv:2409.13598}.

\bibitem[{Selvaraju et~al.(2017)Selvaraju, Cogswell, Das, Vedantam, Parikh, and Batra}]{selvaraju2017grad}
Ramprasaath~R Selvaraju, Michael Cogswell, Abhishek Das, Ramakrishna Vedantam, Devi Parikh, and Dhruv Batra. 2017.
\newblock Grad-cam: Visual explanations from deep networks via gradient-based localization.
\newblock In \emph{Proceedings of the IEEE international conference on computer vision}, pages 618--626.

\bibitem[{Seo et~al.(2022)Seo, Kim, Shin, Kim, Ahn, and Choi}]{seo2022domain}
Minseok Seo, Doyi Kim, Seungheon Shin, Eunbin Kim, Sewoong Ahn, and Yeji Choi. 2022.
\newblock Domain generalization strategy to train classifiers robust to spatial-temporal shift.
\newblock \emph{arXiv preprint arXiv:2212.02968}.

\bibitem[{Shadrin et~al.(2024)Shadrin, Illarionova, Gubanov, Evteeva, Mironenko, Levchunets, Belousov, and Burnaev}]{shadrin2024wildfire}
Dmitrii Shadrin, Svetlana Illarionova, Fedor Gubanov, Ksenia Evteeva, Maksim Mironenko, Ivan Levchunets, Roman Belousov, and Evgeny Burnaev. 2024.
\newblock Wildfire spreading prediction using multimodal data and deep neural network approach.
\newblock \emph{Scientific Reports}, 14(1):2606.

\bibitem[{Shao et~al.(2024)Shao, Feng, Lu, Zhang, and Zou}]{shao2024data}
Pingping Shao, Jun Feng, Jiamin Lu, Pengcheng Zhang, and Chenxin Zou. 2024.
\newblock Data-driven and knowledge-guided denoising diffusion model for flood forecasting.
\newblock \emph{Expert Systems with Applications}, 244:122908.

\bibitem[{She et~al.(2023)She, Zhang, Man, Luo, and Shao}]{she2023self}
Lei She, Chenghong Zhang, Xin Man, Xuewei Luo, and Jie Shao. 2023.
\newblock A self-attention causal lstm model for precipitation nowcasting.
\newblock In \emph{2023 IEEE International Conference on Multimedia and Expo Workshops (ICMEW)}, pages 470--473. IEEE.

\bibitem[{Shen and Kwok(2023)}]{shen2023non}
Lifeng Shen and James Kwok. 2023.
\newblock Non-autoregressive conditional diffusion models for time series prediction.
\newblock In \emph{International Conference on Machine Learning}, pages 31016--31029. PMLR.

\bibitem[{Shi et~al.(2022)Shi, Jain, and Narasimhan}]{shi2022time}
Jimeng Shi, Mahek Jain, and Giri Narasimhan. 2022.
\newblock Time series forecasting (tsf) using various deep learning models.
\newblock \emph{arXiv preprint arXiv:2204.11115}.

\bibitem[{Shi et~al.(2024{\natexlab{a}})Shi, Jin, Han, and Narasimhan}]{shi2024codicast}
Jimeng Shi, Bowen Jin, Jiawei Han, and Giri Narasimhan. 2024{\natexlab{a}}.
\newblock Codicast: Conditional diffusion model for weather prediction with uncertainty quantification.
\newblock \emph{arXiv preprint arXiv:2409.05975}.

\bibitem[{Shi et~al.(2023)Shi, Stebliankin, Wang, Wang, and Narasimhan}]{shi2023graph}
Jimeng Shi, Vitalii Stebliankin, Zhaonan Wang, Shaowen Wang, and Giri Narasimhan. 2023.
\newblock Graph transformer network for flood forecasting with heterogeneous covariates.
\newblock \emph{arXiv preprint arXiv:2310.07631}.

\bibitem[{Shi et~al.(2024{\natexlab{b}})Shi, Yin, Leon, Obeysekera, and Narasimhan}]{shi2024fidlar}
Jimeng Shi, Zeda Yin, Arturo Leon, Jayantha Obeysekera, and Giri Narasimhan. 2024{\natexlab{b}}.
\newblock Fidlar: Forecast-informed deep learning architecture for flood mitigation.
\newblock \emph{arXiv preprint arXiv:2402.13371}.

\bibitem[{Shi et~al.(2015)Shi, Chen, Wang, Yeung, Wong, and Woo}]{shi2015convolutional}
Xingjian Shi, Zhourong Chen, Hao Wang, Dit-Yan Yeung, Wai-Kin Wong, and Wang-chun Woo. 2015.
\newblock Convolutional lstm network: A machine learning approach for precipitation nowcasting.
\newblock \emph{Advances in neural information processing systems}, 28.

\bibitem[{Shi et~al.(2017)Shi, Gao, Lausen, Wang, Yeung, Wong, and Woo}]{shi2017deep}
Xingjian Shi, Zhihan Gao, Leonard Lausen, Hao Wang, Dit-Yan Yeung, Wai-kin Wong, and Wang-chun Woo. 2017.
\newblock Deep learning for precipitation nowcasting: A benchmark and a new model.
\newblock \emph{Advances in neural information processing systems}, 30.

\bibitem[{Shukla and Pandya(2023)}]{shukla2023deep}
Jyoti~S Shukla and Rahul~Jashvantbhai Pandya. 2023.
\newblock Deep learning-oriented c-gan models for vegetative drought prediction on peninsular india.
\newblock \emph{IEEE Journal of Selected Topics in Applied Earth Observations and Remote Sensing}.

\bibitem[{Sit et~al.(2021)Sit, Seo, and Demir}]{sit2021iowarain}
Muhammed Sit, Bong-Chul Seo, and Ibrahim Demir. 2021.
\newblock Iowarain: A statewide rain event dataset based on weather radars and quantitative precipitation estimation.
\newblock \emph{arXiv preprint arXiv:2107.03432}.

\bibitem[{Situ et~al.(2024{\natexlab{a}})Situ, Wang, Teng, Feng, Chen, Zhou, and Fu}]{situ2024improving}
Zuxiang Situ, Qi~Wang, Shuai Teng, Wanen Feng, Gongfa Chen, Qianqian Zhou, and Guangtao Fu. 2024{\natexlab{a}}.
\newblock Improving urban flood prediction using lstm-deeplabv3+ and bayesian optimization with spatiotemporal feature fusion.
\newblock \emph{Journal of Hydrology}, 630:130743.

\bibitem[{Situ et~al.(2024{\natexlab{b}})Situ, Zhong, Zhang, Teng, Ge, Zhou, and Zhao}]{situ2024attention}
Zuxiang Situ, Qisheng Zhong, Jianliang Zhang, Shuai Teng, Xiaoguang Ge, Qianqian Zhou, and Zhiwei Zhao. 2024{\natexlab{b}}.
\newblock Attention-based deep learning framework for urban flood damage and risk assessment with improved flood prediction and land use segmentation.
\newblock \emph{International Journal of Disaster Risk Reduction}, page 105165.

\bibitem[{Smith et~al.(2016)Smith, Lakshmanan, Stumpf, Ortega, Hondl, Cooper, Calhoun, Kingfield, Manross, Toomey et~al.}]{smith2016multi}
Travis~M Smith, Valliappa Lakshmanan, Gregory~J Stumpf, Kiel~L Ortega, Kurt Hondl, Karen Cooper, Kristin~M Calhoun, Darrel~M Kingfield, Kevin~L Manross, Robert Toomey, et~al. 2016.
\newblock Multi-radar multi-sensor (mrms) severe weather and aviation products: Initial operating capabilities.
\newblock \emph{Bulletin of the American Meteorological Society}, 97(9):1617--1630.

\bibitem[{Son et~al.(2022)Son, Ma, Wang, Rasch, Wang, Kim, Jeong, Lim, and Yoon}]{son2022deep}
Rackhun Son, Po-Lun Ma, Hailong Wang, Philp~J Rasch, Shih-Yu Wang, Hyungjun Kim, Jee-Hoon Jeong, Kyo-Sun~Sunny Lim, and Jin-Ho Yoon. 2022.
\newblock Deep learning provides substantial improvements to county-level fire weather forecasting over the western united states.
\newblock \emph{Journal of Advances in Modeling Earth Systems}, 14(10):e2022MS002995.

\bibitem[{S{\o}nderby et~al.(2020)S{\o}nderby, Espeholt, Heek, Dehghani, Oliver, Salimans, Agrawal, Hickey, and Kalchbrenner}]{sonderby2020metnet}
Casper~Kaae S{\o}nderby, Lasse Espeholt, Jonathan Heek, Mostafa Dehghani, Avital Oliver, Tim Salimans, Shreya Agrawal, Jason Hickey, and Nal Kalchbrenner. 2020.
\newblock Metnet: A neural weather model for precipitation forecasting.
\newblock \emph{arXiv preprint arXiv:2003.12140}.

\bibitem[{Song et~al.(2023)Song, Su, Li, Sun, Ren, Liu, and Liu}]{song2023spatial}
Dan Song, Xinqi Su, Wenhui Li, Zhengya Sun, Tongwei Ren, Wen Liu, and An-An Liu. 2023.
\newblock Spatial-temporal transformer network for multi-year enso prediction.
\newblock \emph{Frontiers in Marine Science}, 10:1143499.

\bibitem[{Song et~al.(2020)Song, Meng, and Ermon}]{song2020denoising}
Jiaming Song, Chenlin Meng, and Stefano Ermon. 2020.
\newblock Denoising diffusion implicit models.
\newblock \emph{arXiv preprint arXiv:2010.02502}.

\bibitem[{Sun et~al.(2022)Sun, Yang, Liu, Yin, Li, and Xu}]{sun2022recent}
Zehua Sun, Huanqi Yang, Kai Liu, Zhimeng Yin, Zhenjiang Li, and Weitao Xu. 2022.
\newblock Recent advances in lora: A comprehensive survey.
\newblock \emph{ACM Transactions on Sensor Networks}, 18:1--44.

\bibitem[{Tang et~al.(2024)Tang, Dong, Tang, Chu, and Liang}]{tang2024vmrnn}
Yujin Tang, Peijie Dong, Zhenheng Tang, Xiaowen Chu, and Junwei Liang. 2024.
\newblock Vmrnn: Integrating vision mamba and lstm for efficient and accurate spatiotemporal forecasting.
\newblock In \emph{Proceedings of the IEEE/CVF Conference on Computer Vision and Pattern Recognition}, pages 5663--5673.

\bibitem[{Tang et~al.(2023)Tang, Zhou, Pan, Gong, and Liang}]{tang2023postrainbench}
Yujin Tang, Jiaming Zhou, Xiang Pan, Zeying Gong, and Junwei Liang. 2023.
\newblock \href {https://arxiv.org/abs/2310.02676} {Postrainbench: A comprehensive benchmark and a new model for precipitation forecasting}.
\newblock \emph{Preprint}, arXiv:2310.02676.

\bibitem[{Tian et~al.(2021)Tian, Wu, Cui, and Hu}]{tian2021drought}
Wan Tian, Jiujing Wu, Hengjian Cui, and Tao Hu. 2021.
\newblock Drought prediction based on feature-based transfer learning and time series imaging.
\newblock \emph{IEEE Access}, 9:101454--101468.

\bibitem[{Tobler(1999)}]{tobler1999linear}
Waldo Tobler. 1999.
\newblock Linear pycnophylactic reallocation comment on a paper by d. martin.
\newblock \emph{International Journal of Geographical Information Science}, 13(1):85--90.

\bibitem[{Tobler(2004)}]{tobler2004first}
Waldo Tobler. 2004.
\newblock On the first law of geography: A reply.
\newblock \emph{Annals of the association of American geographers}, 94(2):304--310.

\bibitem[{Trabucco et~al.(2023)Trabucco, Doherty, Gurinas, and Salakhutdinov}]{trabucco2023effective}
Brandon Trabucco, Kyle Doherty, Max Gurinas, and Ruslan Salakhutdinov. 2023.
\newblock Effective data augmentation with diffusion models.
\newblock \emph{arXiv preprint arXiv:2302.07944}.

\bibitem[{Vandal et~al.(2024)Vandal, Duffy, McDuff, Nachmany, and Hartshorn}]{vandal2024global}
Thomas~J Vandal, Kate Duffy, Daniel McDuff, Yoni Nachmany, and Chris Hartshorn. 2024.
\newblock Global atmospheric data assimilation with multi-modal masked autoencoders.
\newblock \emph{arXiv preprint arXiv:2407.11696}.

\bibitem[{Vaswani(2017)}]{vaswani2017attention}
A~Vaswani. 2017.
\newblock Attention is all you need.
\newblock \emph{Advances in Neural Information Processing Systems}.

\bibitem[{Veillette et~al.(2020)Veillette, Samsi, and Mattioli}]{veillette2020sevir}
Mark Veillette, Siddharth Samsi, and Chris Mattioli. 2020.
\newblock Sevir: A storm event imagery dataset for deep learning applications in radar and satellite meteorology.
\newblock \emph{Advances in Neural Information Processing Systems}, 33:22009--22019.

\bibitem[{Verma et~al.(2023)Verma, Srivastava, Tiwari, and Verma}]{verma2023deep}
Shikha Verma, Kuldeep Srivastava, Akhilesh Tiwari, and Shekhar Verma. 2023.
\newblock Deep learning techniques in extreme weather events: A review.
\newblock \emph{arXiv preprint arXiv:2308.10995}.

\bibitem[{Verma et~al.(2024)Verma, Heinonen, and Garg}]{verma2024climode}
Yogesh Verma, Markus Heinonen, and Vikas Garg. 2024.
\newblock Climode: Climate and weather forecasting with physics-informed neural odes.
\newblock \emph{arXiv preprint arXiv:2404.10024}.

\bibitem[{Vito(2016)}]{italyAirQuality}
Saverio Vito. 2016.
\newblock \href {https://archive.ics.uci.edu/dataset/360/air+quality} {Italy air quality data set}.
\newblock UCI Machine Learning Repository.

\bibitem[{von Bortkiewicz(1921)}]{von1921variationsbreite}
Ladislaus von Bortkiewicz. 1921.
\newblock \emph{Variationsbreite und mittlerer Fehler}.
\newblock Berliner Mathematische Gesellschaft.

\bibitem[{Vosper et~al.(2023)Vosper, Watson, Harris, McRae, Santos-Rodriguez, Aitchison, and Mitchell}]{vosper2023deep}
Emily Vosper, Peter Watson, Lucy Harris, Andrew McRae, Raul Santos-Rodriguez, Laurence Aitchison, and Dann Mitchell. 2023.
\newblock Deep learning for downscaling tropical cyclone rainfall to hazard-relevant spatial scales.
\newblock \emph{Journal of Geophysical Research: Atmospheres}, 128(10):e2022JD038163.

\bibitem[{Wang et~al.(2023{\natexlab{a}})Wang, Cheng, Zhang, and Yu}]{wang2023enso}
Gai-Ge Wang, Honglei Cheng, Yiming Zhang, and Hui Yu. 2023{\natexlab{a}}.
\newblock Enso analysis and prediction using deep learning: a review.
\newblock \emph{Neurocomputing}, 520:216--229.

\bibitem[{Wang et~al.(2023{\natexlab{b}})Wang, Fung, and Lau}]{wang2023physical}
Rui Wang, Jimmy~CH Fung, and Alexis~KH Lau. 2023{\natexlab{b}}.
\newblock Physical-dynamic-driven ai-synthetic precipitation nowcasting using task-segmented generative model.
\newblock \emph{Geophysical Research Letters}, 50(21):e2023GL106084.

\bibitem[{Wang et~al.(2023{\natexlab{c}})Wang, Su, Wong, Lau, and Fung}]{wang2023skillful}
Rui Wang, Lin Su, Wai~Kin Wong, Alexis~KH Lau, and Jimmy~CH Fung. 2023{\natexlab{c}}.
\newblock Skillful radar-based heavy rainfall nowcasting using task-segmented generative adversarial network.
\newblock \emph{IEEE Transactions on Geoscience and Remote Sensing}.

\bibitem[{Wang et~al.(2020)Wang, Li, Zhang, Meng, Meng, and Gao}]{wang2020pm2}
Shuo Wang, Yanran Li, Jiang Zhang, Qingye Meng, Lingwei Meng, and Fei Gao. 2020.
\newblock Pm2.5-gnn: A domain knowledge enhanced graph neural network for pm2. 5 forecasting.
\newblock In \emph{Proceedings of the 28th international conference on advances in geographic information systems}, pages 163--166.

\bibitem[{Wang et~al.(2024)Wang, Ba, Zhang, Zhang, Kadambi, Soatto, Wong, and Hsieh}]{wang2024evaluating}
Yihan Wang, Yunhao Ba, Howard~Chenyang Zhang, Huan Zhang, Achuta Kadambi, Stefano Soatto, Alex Wong, and Cho-Jui Hsieh. 2024.
\newblock Evaluating worst case adversarial weather perturbations robustness.
\newblock In \emph{NeurIPS ML Safety Workshop}.

\bibitem[{Wang et~al.(2018)Wang, Gao, Long, Wang, and Philip}]{wang2018predrnn}
Yunbo Wang, Zhifeng Gao, Mingsheng Long, Jianmin Wang, and S~Yu Philip. 2018.
\newblock Predrnn++: Towards a resolution of the deep-in-time dilemma in spatiotemporal predictive learning.
\newblock In \emph{International conference on machine learning}, pages 5123--5132. PMLR.

\bibitem[{Wang et~al.(2017)Wang, Long, Wang, Gao, and Yu}]{wang2017predrnn}
Yunbo Wang, Mingsheng Long, Jianmin Wang, Zhifeng Gao, and Philip~S Yu. 2017.
\newblock Predrnn: Recurrent neural networks for predictive learning using spatiotemporal lstms.
\newblock \emph{Advances in neural information processing systems}, 30.

\bibitem[{Wang et~al.(2022)Wang, Wu, Zhang, Gao, Wang, Philip, and Long}]{wang2022predrnn}
Yunbo Wang, Haixu Wu, Jianjin Zhang, Zhifeng Gao, Jianmin Wang, S~Yu Philip, and Mingsheng Long. 2022.
\newblock Predrnn: A recurrent neural network for spatiotemporal predictive learning.
\newblock \emph{IEEE Transactions on Pattern Analysis and Machine Intelligence}, 45(2):2208--2225.

\bibitem[{Waseem et~al.(2022)Waseem, Mushtaq, Abid, Abu-Mahfouz, Shaikh, Turan, and Rasheed}]{waseem2022forecasting}
Khawaja~Hassan Waseem, Hammad Mushtaq, Fazeel Abid, Adnan~M Abu-Mahfouz, Asadullah Shaikh, Mehmet Turan, and Jawad Rasheed. 2022.
\newblock Forecasting of air quality using an optimized recurrent neural network.
\newblock \emph{Processes}, 10(10):2117.

\bibitem[{Washburn and Wood(1995)}]{washburn1995two}
Alan Washburn and Kevin Wood. 1995.
\newblock Two-person zero-sum games for network interdiction.
\newblock \emph{Operations research}, 43(2):243--251.

\bibitem[{Wen et~al.(2022)Wen, Zhou, Zhang, Chen, Ma, Yan, and Sun}]{wen2022transformers}
Qingsong Wen, Tian Zhou, Chaoli Zhang, Weiqi Chen, Ziqing Ma, Junchi Yan, and Liang Sun. 2022.
\newblock Transformers in time series: A survey.
\newblock \emph{arXiv preprint arXiv:2202.07125}.

\bibitem[{Wilhite(2016)}]{wilhite2016drought}
Donald~A Wilhite. 2016.
\newblock Drought as a natural hazard: concepts and definitions.
\newblock In \emph{Droughts}, pages 3--18. Routledge.

\bibitem[{Woo et~al.(2024)Woo, Liu, Kumar, Xiong, Savarese, and Sahoo}]{woo2024unified}
Gerald Woo, Chenghao Liu, Akshat Kumar, Caiming Xiong, Silvio Savarese, and Doyen Sahoo. 2024.
\newblock Unified training of universal time series forecasting transformers.
\newblock \emph{arXiv preprint arXiv:2402.02592}.

\bibitem[{Wu et~al.(2024)Wu, Chen, Wang, Peng, Sun, and Chen}]{wu2024weathergnn}
Binqing Wu, Weiqi Chen, Wengwei Wang, Bingqing Peng, Liang Sun, and Ling Chen. 2024.
\newblock Weathergnn: Exploiting meteo-and spatial-dependencies for local numerical weather prediction bias-correction.
\newblock In \emph{Proceedings of the International Joint Conference on Artificial Intelligence}, pages 2433--2441.

\bibitem[{Wu et~al.(2023)Wu, Zhou, Long, and Wang}]{wu2023interpretable}
Haixu Wu, Hang Zhou, Mingsheng Long, and Jianmin Wang. 2023.
\newblock Interpretable weather forecasting for worldwide stations with a unified deep model.
\newblock \emph{Nature Machine Intelligence}, 5(6):602--611.

\bibitem[{Wu et~al.(2021)Wu, Geng, Liu, and Shi}]{wu2021tropical}
Yuqiao Wu, Xiaoyi Geng, Zili Liu, and Zhenwei Shi. 2021.
\newblock Tropical cyclone forecast using multitask deep learning framework.
\newblock \emph{IEEE Geoscience and Remote Sensing Letters}, 19:1--5.

\bibitem[{Wu et~al.(2020)Wu, Pan, Chen, Long, Zhang, and Philip}]{wu2020comprehensive}
Zonghan Wu, Shirui Pan, Fengwen Chen, Guodong Long, Chengqi Zhang, and S~Yu Philip. 2020.
\newblock A comprehensive survey on graph neural networks.
\newblock \emph{IEEE transactions on neural networks and learning systems}, 32(1):4--24.

\bibitem[{Xiang et~al.(2024)Xiang, Jin, Dong, Bai, Fang, Zhao, Sun, Thambiratnam, Zhang, and Huang}]{xiang2024adaf}
Yanfei Xiang, Weixin Jin, Haiyu Dong, Mingliang Bai, Zuliang Fang, Pengcheng Zhao, Hongyu Sun, Kit Thambiratnam, Qi~Zhang, and Xiaomeng Huang. 2024.
\newblock Adaf: An artificial intelligence data assimilation framework for weather forecasting.
\newblock \emph{arXiv preprint arXiv:2411.16807}.

\bibitem[{Xiao et~al.(2023)Xiao, Bai, Xue, Chen, Han, and Ouyang}]{xiao2023fengwu}
Yi~Xiao, Lei Bai, Wei Xue, Kang Chen, Tao Han, and Wanli Ouyang. 2023.
\newblock Fengwu-4dvar: Coupling the data-driven weather forecasting model with 4d variational assimilation.
\newblock \emph{arXiv preprint arXiv:2312.12455}.

\bibitem[{Xiong et~al.(2024)Xiong, Jin, Lu, and Zhang}]{xiong2024benchmarking}
Guangzhi Xiong, Qiao Jin, Zhiyong Lu, and Aidong Zhang. 2024.
\newblock Benchmarking retrieval-augmented generation for medicine.
\newblock \emph{arXiv preprint arXiv:2402.13178}.

\bibitem[{Xu et~al.(2022)Xu, Zhang, Ding, and Zhang}]{xu2022application}
Dehe Xu, Qi~Zhang, Yan Ding, and De~Zhang. 2022.
\newblock Application of a hybrid arima-lstm model based on the spei for drought forecasting.
\newblock \emph{Environmental Science and Pollution Research}, 29(3):4128--4144.

\bibitem[{Xu et~al.(2024)Xu, Qin, Sun, Liao, and Zheng}]{xu2024pfformer}
Luwen Xu, Jiwei Qin, Dezhi Sun, Yuanyuan Liao, and Jiong Zheng. 2024.
\newblock Pfformer: A time-series forecasting model for short-term precipitation forecasting.
\newblock \emph{IEEE Access}.

\bibitem[{Xu et~al.(2018)Xu, Zhang, Huang, Zhang, Gan, Huang, and He}]{xu2018attngan}
Tao Xu, Pengchuan Zhang, Qiuyuan Huang, Han Zhang, Zhe Gan, Xiaolei Huang, and Xiaodong He. 2018.
\newblock Attngan: Fine-grained text to image generation with attentional generative adversarial networks.
\newblock In \emph{Proceedings of the IEEE conference on computer vision and pattern recognition}, pages 1316--1324.

\bibitem[{Yadav et~al.(2022)Yadav, Zaidi, Mishra, and Yadav}]{yadav2022survey}
Satya~Prakash Yadav, Subiya Zaidi, Annu Mishra, and Vibhash Yadav. 2022.
\newblock Survey on machine learning in speech emotion recognition and vision systems using a recurrent neural network (rnn).
\newblock \emph{Archives of Computational Methods in Engineering}, 29(3):1753--1770.

\bibitem[{Yang et~al.(2023)Yang, Zhang, Song, Hong, Xu, Zhao, Zhang, Cui, and Yang}]{yang2023diffusion}
Ling Yang, Zhilong Zhang, Yang Song, Shenda Hong, Runsheng Xu, Yue Zhao, Wentao Zhang, Bin Cui, and Ming-Hsuan Yang. 2023.
\newblock Diffusion models: A comprehensive survey of methods and applications.
\newblock \emph{ACM Computing Surveys}, 56(4):1--39.

\bibitem[{Yang et~al.(2024{\natexlab{a}})Yang, Giezendanner, Civitarese, Jakubik, Schmitt, Chandra, Vila, Hohl, Hill, Watson et~al.}]{yang2024multi}
Qidong Yang, Jonathan Giezendanner, Daniel~Salles Civitarese, Johannes Jakubik, Eric Schmitt, Anirban Chandra, Jeremy Vila, Detlef Hohl, Chris Hill, Campbell Watson, et~al. 2024{\natexlab{a}}.
\newblock Multi-modal graph neural networks for localized off-grid weather forecasting.
\newblock \emph{arXiv preprint arXiv:2410.12938}.

\bibitem[{Yang et~al.(2024{\natexlab{b}})Yang, Jin, Wen, Zhang, Liang, Ma, Wang, Liu, Yang, Xu et~al.}]{yang2024survey}
Yiyuan Yang, Ming Jin, Haomin Wen, Chaoli Zhang, Yuxuan Liang, Lintao Ma, Yi~Wang, Chenghao Liu, Bin Yang, Zenglin Xu, et~al. 2024{\natexlab{b}}.
\newblock A survey on diffusion models for time series and spatio-temporal data.
\newblock \emph{arXiv preprint arXiv:2404.18886}.

\bibitem[{Ye et~al.(2021)Ye, Hu, Huang, You, Weng, and Gao}]{ye2021transformer}
Feng Ye, Jie Hu, Tian-Qiang Huang, Li-Jun You, Bin Weng, and Jian-Yun Gao. 2021.
\newblock Transformer for ei ni{\~n}o-southern oscillation prediction.
\newblock \emph{IEEE Geoscience and Remote Sensing Letters}, 19:1--5.

\bibitem[{Yi et~al.(2018)Yi, Zhang, Wang, Li, and Zheng}]{yi2018deep}
Xiuwen Yi, Junbo Zhang, Zhaoyuan Wang, Tianrui Li, and Yu~Zheng. 2018.
\newblock Deep distributed fusion network for air quality prediction.
\newblock In \emph{Proceedings of the 24th ACM SIGKDD international conference on knowledge discovery \& data mining}, pages 965--973.

\bibitem[{Yin et~al.(2024)Yin, Meo, Roy, Cher, Lic{\u{a}}, Wang, Imhoff, Uijlenhoet, and Dauwels}]{yin2024precipitation}
Junzhe Yin, Cristian Meo, Ankush Roy, Zeineh~Bou Cher, Mircea Lic{\u{a}}, Yanbo Wang, Ruben Imhoff, Remko Uijlenhoet, and Justin Dauwels. 2024.
\newblock Precipitation nowcasting using physics informed discriminator generative models.
\newblock In \emph{2024 32nd European Signal Processing Conference (EUSIPCO)}, pages 967--971. IEEE.

\bibitem[{Yin et~al.(2023)Yin, Bian, Hu, Shi, and Leon}]{yin2023physic}
Zeda Yin, Linglong Bian, Beichao Hu, Jimeng Shi, and Arturo~S Leon. 2023.
\newblock Physic-informed neural network approach coupled with boundary conditions for solving 1d steady shallow water equations for riverine system.
\newblock In \emph{World Environmental and Water Resources Congress 2023}, pages 280--288.

\bibitem[{Yu et~al.(2017)Yu, Yin, and Zhu}]{yu2017spatio}
Bing Yu, Haoteng Yin, and Zhanxing Zhu. 2017.
\newblock Spatio-temporal graph convolutional networks: A deep learning framework for traffic forecasting.
\newblock \emph{arXiv preprint arXiv:1709.04875}.

\bibitem[{Yu et~al.(2025)Yu, Wang, Wang, Shao, Sun, Yao, and Xu}]{yu2025mgsfformer}
Chengqing Yu, Fei Wang, Yilun Wang, Zezhi Shao, Tao Sun, Di~Yao, and Yongjun Xu. 2025.
\newblock Mgsfformer: A multi-granularity spatiotemporal fusion transformer for air quality prediction.
\newblock \emph{Information Fusion}, 113:102607.

\bibitem[{Yu et~al.(2024{\natexlab{a}})Yu, Li, Ye, Zhang, Luo, Dai, Wang, and Chen}]{yu2024diffcast}
Demin Yu, Xutao Li, Yunming Ye, Baoquan Zhang, Chuyao Luo, Kuai Dai, Rui Wang, and Xunlai Chen. 2024{\natexlab{a}}.
\newblock Diffcast: A unified framework via residual diffusion for precipitation nowcasting.
\newblock In \emph{Proceedings of the IEEE/CVF Conference on Computer Vision and Pattern Recognition}, pages 27758--27767.

\bibitem[{Yu et~al.(2024{\natexlab{b}})Yu, Hu, Subramaniam, Hannah, Peng, Lin, Bhouri, Gupta, L{\"u}tjens, Will et~al.}]{yu2024climsim}
Sungduk Yu, Zeyuan Hu, Akshay Subramaniam, Walter Hannah, Liran Peng, Jerry Lin, Mohamed~Aziz Bhouri, Ritwik Gupta, Bj{\"o}rn L{\"u}tjens, Justus~C Will, et~al. 2024{\natexlab{b}}.
\newblock Climsim-online: A large multi-scale dataset and framework for hybrid ml-physics climate emulation.
\newblock \emph{arXiv preprint arXiv:2306.08754}.

\bibitem[{Yuan et~al.(2025)Yuan, Wang, Mu, and Zhou}]{yuan2025tianxing}
Shijin Yuan, Guansong Wang, Bin Mu, and Feifan Zhou. 2025.
\newblock Tianxing: A linear complexity transformer model with explicit attention decay for global weather forecasting.
\newblock \emph{Advances in Atmospheric Sciences}, 42(1):9--25.

\bibitem[{Zeng et~al.(2023)Zeng, Chen, Zhang, and Xu}]{zeng2023transformers}
Ailing Zeng, Muxi Chen, Lei Zhang, and Qiang Xu. 2023.
\newblock Are transformers effective for time series forecasting?
\newblock In \emph{Proceedings of the AAAI conference on artificial intelligence}, volume~37, pages 11121--11128.

\bibitem[{Zhang et~al.(2017)Zhang, Xiong, Su, and Duan}]{zhang2017context}
Biao Zhang, Deyi Xiong, Jinsong Su, and Hong Duan. 2017.
\newblock A context-aware recurrent encoder for neural machine translation.
\newblock \emph{IEEE/ACM Transactions on Audio, Speech, and Language Processing}, 25(12):2424--2432.

\bibitem[{Zhang et~al.(2011)Zhang, Lee, Wang, Li, Pei, Zhang, and An}]{zhang2011causality}
David~D Zhang, Harry~F Lee, Cong Wang, Baosheng Li, Qing Pei, Jane Zhang, and Yulun An. 2011.
\newblock The causality analysis of climate change and large-scale human crisis.
\newblock \emph{Proceedings of the National Academy of Sciences}, 108(42):17296--17301.

\bibitem[{Zhang et~al.(2024{\natexlab{a}})Zhang, Huang, and Sun}]{zhang2024multiscale}
Jia-Li Zhang, Xiao-Meng Huang, and Yu-Ze Sun. 2024{\natexlab{a}}.
\newblock Multiscale spatiotemporal meteorological drought prediction: A deep learning approach.
\newblock \emph{Advances in Climate Change Research}, 15(2):211--221.

\bibitem[{Zhang et~al.(2024{\natexlab{b}})Zhang, Yan, Amonkar, Nayak, and Lall}]{zhang2024potential}
Mengjie Zhang, Lei Yan, Yash Amonkar, Adam Nayak, and Upmanu Lall. 2024{\natexlab{b}}.
\newblock Potential climate predictability of renewable energy supply and demand for texas given the enso hidden state.
\newblock \emph{Science Advances}, 10(44):eado3517.

\bibitem[{Zhang et~al.(2024{\natexlab{c}})Zhang, Li, Huang, Wang, Li, and Shen}]{zhang2024multivariate}
Q~Zhang, YP~Li, GH~Huang, H~Wang, YF~Li, and ZY~Shen. 2024{\natexlab{c}}.
\newblock Multivariate time series convolutional neural networks for long-term agricultural drought prediction under global warming.
\newblock \emph{Agricultural Water Management}, 292:108683.

\bibitem[{Zhang et~al.(2021)Zhang, Liu, Hang, and Liu}]{zhang2021predicting}
Rui Zhang, Qingshan Liu, Renlong Hang, and Guangcan Liu. 2021.
\newblock Predicting tropical cyclogenesis using a deep learning method from gridded satellite and era5 reanalysis data in the western north pacific basin.
\newblock \emph{IEEE Transactions on Geoscience and Remote Sensing}, 60:1--10.

\bibitem[{Zhang et~al.(2023{\natexlab{a}})Zhang, Xie, Tian, Zhao, Geng, Lu, Ma, Huang, and Choy Lim Kam~Sian}]{zhang2023construction}
Yonghong Zhang, Donglin Xie, Wei Tian, Huajun Zhao, Sutong Geng, Huanyu Lu, Guangyi Ma, Jie Huang, and Kenny~Thiam Choy Lim Kam~Sian. 2023{\natexlab{a}}.
\newblock Construction of an integrated drought monitoring model based on deep learning algorithms.
\newblock \emph{Remote Sensing}, 15(3):667.

\bibitem[{Zhang et~al.(2023{\natexlab{b}})Zhang, Long, Chen, Xing, Jin, Jordan, and Wang}]{zhang2023skilful}
Yuchen Zhang, Mingsheng Long, Kaiyuan Chen, Lanxiang Xing, Ronghua Jin, Michael~I Jordan, and Jianmin Wang. 2023{\natexlab{b}}.
\newblock Skilful nowcasting of extreme precipitation with nowcastnet.
\newblock \emph{Nature}, 619(7970):526--532.

\bibitem[{Zhang and Chong(2007)}]{zhang2007comparison}
Zheng Zhang and Kil~To Chong. 2007.
\newblock Comparison between first-order hold with zero-order hold in discretization of input-delay nonlinear systems.
\newblock In \emph{2007 International Conference on Control, Automation and Systems}, pages 2892--2896. IEEE.

\bibitem[{Zhao et~al.(2024{\natexlab{a}})Zhao, Bian, Ni, Jin, Weyn, Fang, Xiang, Dong, Zhang, Sun et~al.}]{zhao2024omg}
Pengcheng Zhao, Jiang Bian, Zekun Ni, Weixin Jin, Jonathan Weyn, Zuliang Fang, Siqi Xiang, Haiyu Dong, Bin Zhang, Hongyu Sun, et~al. 2024{\natexlab{a}}.
\newblock Omg-hd: A high-resolution ai weather model for end-to-end forecasts from observations.
\newblock \emph{arXiv preprint arXiv:2412.18239}.

\bibitem[{Zhao et~al.(2023)Zhao, Zhou, Li, Tang, Wang, Hou, Min, Zhang, Zhang, Dong et~al.}]{zhao2023survey}
Wayne~Xin Zhao, Kun Zhou, Junyi Li, Tianyi Tang, Xiaolei Wang, Yupeng Hou, Yingqian Min, Beichen Zhang, Junjie Zhang, Zican Dong, et~al. 2023.
\newblock A survey of large language models.
\newblock \emph{arXiv preprint arXiv:2303.18223}.

\bibitem[{Zhao et~al.(2024{\natexlab{b}})Zhao, Zhou, Zhang, Liu, Chen, Gong, Chen, Fei, Chen, Ouyang et~al.}]{zhao2024weathergfm}
Xiangyu Zhao, Zhiwang Zhou, Wenlong Zhang, Yihao Liu, Xiangyu Chen, Junchao Gong, Hao Chen, Ben Fei, Shiqi Chen, Wanli Ouyang, et~al. 2024{\natexlab{b}}.
\newblock Weathergfm: Learning a weather generalist foundation model via in-context learning.
\newblock \emph{arXiv preprint arXiv:2411.05420}.

\bibitem[{Zheng et~al.(2023)Zheng, Shen, Tang, Luo, Hu, Du, and Tao}]{zheng2023learn}
Hongling Zheng, Li~Shen, Anke Tang, Yong Luo, Han Hu, Bo~Du, and Dacheng Tao. 2023.
\newblock Learn from model beyond fine-tuning: A survey.
\newblock \emph{arXiv preprint arXiv:2310.08184}.

\bibitem[{Zheng et~al.(2013)Zheng, Liu, and Hsieh}]{zheng2013u}
Yu~Zheng, Furui Liu, and Hsun-Ping Hsieh. 2013.
\newblock U-air: When urban air quality inference meets big data.
\newblock In \emph{Proceedings of the 19th ACM SIGKDD international conference on Knowledge discovery and data mining}, pages 1436--1444.

\bibitem[{Zheng et~al.(2022)Zheng, Liu, and Zheng}]{zheng2022p}
Zhentan Zheng, Jianyi Liu, and Nanning Zheng. 2022.
\newblock $p^2$-gan: Efficient stroke style transfer using single style image.
\newblock \emph{IEEE Transactions on Multimedia}.

\bibitem[{Zhou and Zhang(2023)}]{zhou2023self}
Lu~Zhou and Rong-Hua Zhang. 2023.
\newblock A self-attention--based neural network for three-dimensional multivariate modeling and its skillful enso predictions.
\newblock \emph{Science Advances}, 9(10):eadf2827.

\bibitem[{Zhou et~al.(2022)Zhou, Ma, Wen, Wang, Sun, and Jin}]{zhou2022fedformer}
Tian Zhou, Ziqing Ma, Qingsong Wen, Xue Wang, Liang Sun, and Rong Jin. 2022.
\newblock Fedformer: Frequency enhanced decomposed transformer for long-term series forecasting.
\newblock In \emph{International Conference on Machine Learning}, pages 27268--27286. PMLR.

\bibitem[{Zhu et~al.(2017)Zhu, Park, Isola, and Efros}]{zhu2017unpaired}
Jun-Yan Zhu, Taesung Park, Phillip Isola, and Alexei~A Efros. 2017.
\newblock Unpaired image-to-image translation using cycle-consistent adversarial networks.
\newblock In \emph{Proceedings of the IEEE international conference on computer vision}, pages 2223--2232.

\bibitem[{Zhu et~al.(2024{\natexlab{a}})Zhu, Liao, Zhang, Wang, Liu, and Wang}]{zhu2024vision}
Lianghui Zhu, Bencheng Liao, Qian Zhang, Xinlong Wang, Wenyu Liu, and Xinggang Wang. 2024{\natexlab{a}}.
\newblock Vision mamba: Efficient visual representation learning with bidirectional state space model.
\newblock \emph{arXiv preprint arXiv:2401.09417}.

\bibitem[{Zhu et~al.(2024{\natexlab{b}})Zhu, Xiong, Wang, Stewart, Heidler, Wang, Yuan, Dujardin, Xu, and Shi}]{zhu2024foundations}
Xiao~Xiang Zhu, Zhitong Xiong, Yi~Wang, Adam~J Stewart, Konrad Heidler, Yuanyuan Wang, Zhenghang Yuan, Thomas Dujardin, Qingsong Xu, and Yilei Shi. 2024{\natexlab{b}}.
\newblock On the foundations of earth and climate foundation models.
\newblock \emph{arXiv preprint arXiv:2405.04285}.

\bibitem[{Zhu et~al.(2023)Zhu, Xiong, Wu, Nie, Zhang, and Yang}]{zhu2023weather2k}
Xun Zhu, Yutong Xiong, Ming Wu, Gaozhen Nie, Bin Zhang, and Ziheng Yang. 2023.
\newblock Weather2k: A multivariate spatio-temporal benchmark dataset for meteorological forecasting based on real-time observation data from ground weather stations.
\newblock \emph{arXiv preprint arXiv:2302.10493}.

\end{thebibliography}

\newpage
\appendix
\onecolumn

\section*{Appendix}
\section{Datasets}
\label{sec:datasets}
We summarize widely used benchmark datasets, where each data set is presented by domain, name, coverage, collection method, spatial and temporal resolution, time span, and the paper that introduces the dataset.
\begin{table*}[ht]
\small
\centering
\caption{Summary of Publicly Available Data Sets on Weather. CAM5: Community Atmospheric Model v5. }
\vspace{-2mm}
\resizebox{0.999\columnwidth}{!}{
    \begin{tabular}{l|l|lllllll}
        \toprule
            \textbf{Domain} & \textbf{Dataset}  & \textbf{Coverage}  &\textbf{Collect}  & \textbf{Spatial}      & \textbf{Temporal} & \textbf{Time Span}    & \textbf{Paper} \\
        \midrule
            \multirow{5}{*}{General Weather} 
                            & WeatherBench      & Global             & Reanalysis       & $1.40625^\circ,2.8125^\circ,5.625^\circ$  & 6 hours  &1979-2018   & \cite{rasp2020weatherbench} \\
                            & WeatherBench 2    & Global             & Reanalysis       & $0.25^\circ$          & 6 hours           & 1979-2020             & \cite{rasp2023weatherbench} \\
                            & Weather2K         & Region in China    & Observation      & -                     & 1 hour            & 2017.01-2021.08       & \cite{zhu2023weather2k} \\
                            & Weather5K         & Global             & Observation      & -                     & 1 hour            & 2014-2023             & \cite{han2024weather} \\
                            & HR-Extreme        & Region in U.S.     & Radar            & 3 km$\times$3 km      & 1 hour            & 2020-2020             & \cite{ran2024hr} \\
        \midrule
            \multirow{12}{*}{Precipitation} 
                            & SEVIR             & Region in U.S.     & Radar\&Satellite & 1 km$\times$1 km      & 5 mins            &2017-2019              & \cite{veillette2020sevir}   \\
                            & OPERA             & Europe             & Radar\&Satellite & 2 km                  & 15 mins           &2019-2021              & \cite{herruzo2021high} \\
                            & Meteonet          & France             & Radar\&Satellite & 1 km                  & 5-15 mins         &2016-2018              & \cite{larvormeteonet} \\
                            & IMERG             & Global             & Radar\&Satellite & 1 km                  & 30 mins           &2020-2023              & \cite{huffman2020integrated}\\
                            & HKO-7             & Region in Hong Kong& Radar            & 1 km$\times$1 km      & 6 mins            &2009-2015              & \cite{shi2017deep}   \\
                            & Shanghai          & Shanghai           & Radar            & 1 km                  & 6 mins            &2015-2018              & \cite{chen2020deep} \\
                            & JMA               & Japan              & Radar            & 1 km                  & 5 mins            &2015-2017              & \cite{inoue2022learning} \\
                            & MRMS              & CONUS and S. Canada& Radar            & 1 km$\times$1 km      & 2 mins            &2017-2019              & \cite{smith2016multi} \\
                            & RYDL              & Germany            & Radar            & 1 km                  & 5 mins            &2014-2015              & \cite{gmd-13-2631-2020} \\
                            & RainBench         & -                  &                  & $5.625^\circ$         & 1 hour            &2016-2019              & \cite{de2021rainbench}  \\
                            & IowaRain          & Iowa, U.S.         & Radar            & 0.5 km$\times$0.5 km  & 5 mins            &2016-2019              & \cite{sit2021iowarain}  \\
                            & PostRainBench     & Region in China    &                  & 1 km$\times$1 km      & 3 hours           &2010-2021              & \cite{tang2023postrainbench} \\
        \midrule
            \multirow{4}{*}{Wind} 
                            & GlobalWindTemp    & Global             & Observation      & -                     & 1 hour            &2019-2010              & \cite{wu2023interpretable}\\
                            & DigitalTyphoon    & W.N. Pacific basin & Satellite        & 5 km                  & 1 hour            &1978-2022              & \cite{kitamoto2023digital}\\
                            & TropicalCyclone   & Global             & CAM5 simulation  & 25 km                 & 3 hours           &1979-2005              & \cite{racah2017extremeweather}\\
                            & ClimateNet        & Global             & CAM5 simulation  & 25 km                 & 3 hours           &1996-2010              & \cite{kashinath2021climatenet}\\
        \midrule
            \multirow{5}{*}{Air Quality} 
                            & UrbanAir          & Regional, China    & Observation      & -                     & 1 hour            &2014-2015              & \cite{zheng2013u}\\
                            & KnowAir           & Regional, China    & Observation      & -                     & 3 hours           &2015-2018              & \cite{wang2020pm2}\\
                            & ItalianAir        & Italy              & Observation      & -                     & 1 hour            &2004-2005              & \cite{italyAirQuality}\\
                            & BeijingAir1       & Regional, China    & Observation      & -                     & 1 hour            &2010-2014              & \cite{bjAirQuality1}\\
                            & BeijingAir2       & Regional, China    & Observation      & -                     & 1 hour            &2013-2017              & \cite{bjAirQuality2}\\
        \midrule
            \multirow{21}{*}{SST} 
                            & OI SST v2            & Pacific Ocean       & Observation\&Satellite       & $5^\circ$S-$5^\circ$N, $170^\circ$W-$120^\circ$W  & Daily     & 1982–2017 & \cite{huang2019analyzing} \\ 
                            & ZonalWinds           & Pacific Ocean       & Reanalysis                   & $5^\circ$S-$5^\circ$N, $120^\circ$E-$160^\circ$E  & Daily     & 1982–2017 & \cite{huang2019analyzing} \\ 
                            & TropicalOcean        & Pacific Ocean       & Observation                  & $5^\circ$S-$5^\circ$N, $120^\circ$E-$80^\circ$W   & Monthly   & 1982–2017 & \cite{huang2019analyzing} \\ 
                            & SODA SST             & Global              & Reanalysis                   & $5^\circ\times5^\circ$                            & Monthly   & 1871–1973 & \cite{geng2021spatiotemporal} \\ 
                            & GODAS                & Global              & Reanalysis                   & $5^\circ\times5^\circ$                            & Monthly   & 1994–2010 & \cite{geng2021spatiotemporal} \\ 
                            & CMIP5                & Global              & Simulation                   & $5^\circ\times5^\circ$                            & Monthly   & 1861–2004 & \cite{geng2021spatiotemporal} \\ 
                            & ERA-Interim          & Global              & Reanalysis                   & -                                                 & Daily     & 1984–2017 & \cite{ham2019deep} \\ 
                            & CFSv2                & Global              & Reanalysis                   & $5^\circ\times5^\circ$                            & 6 hours   & 1981–2017 & \cite{he2019dlenso} \\ 
                            & NOAA ERSSTv5         & Global              & Observation                  & -                                                 & Monthly   & 1854–2020 & \cite{cachay2020graph} \\ 
                            & CMIP6                & Tropical Pacific    & Simulation                   & $2^\circ\times0.5^\circ$                          & Monthly   & 1850–2014 & \cite{zhou2023self} \\ 
                            & ORAS5                & Tropical Pacific    & Reanalysis                   & -                                                 & Monthly   & 1958–1979 & \cite{zhou2023self} \\ 
                            & NOAA/CIRE            & Global              & Reanalysis                   & $2^\circ\times2^\circ$                            & 6 hours   & 1850–2015 & \citeauthor{mu2021enso} \\ 
                            & REMSS                & Global              & Satellite                    & $0.25^\circ\times0.25^\circ$                      & Daily     & 1997–2020 & \citeauthor{mu2021enso} \\ 
                            & ENSO                 & Tropical Pacific    & NOAA, NCEI, NCAR             & -                                                 & Monthly   & 1950–2023 & \cite{mir2024enso} \\ 
                            & GHRSST               & South China Sea     & Observation                  & $1.20^\circ\times1.20^\circ$                      & Daily     & 2007–2014 & \cite{meng2023physical} \\ 
                            & HYCOM                & South China Sea     & Simulation                   & $1.12^\circ\times1.12^\circ$                      & Daily     & 2007–2014 & \cite{meng2023physical} \\ 
                            & Hadley-OI SST        & Global              & Observation\&Satellite       & $1^\circ\times1^\circ$                            & Monthly   & 1870–2020 & \cite{liu2023explainable} \\ 
                            & COBE SST             & Global              & Observation                  & $1^\circ\times1^\circ$                            & Monthly   & 1891–2020 & \cite{liu2023explainable} \\ 
                            & SILO SST             & Australia           & Observation                  & -                                                 & Monthly   & 1921–2020 & \cite{he2024data} \\ 
                            & OISST                & Global              & Observation\&Reanalysis      & $0.25^\circ\times0.25^\circ$                      & Daily     & 1982–2020 & \cite{he2024interpretable} \\ 
                            & ERA5                 & Global              & Observation\&Reanalysis      & $0.25^\circ\times0.25^\circ$                      & 1 hour    & 1982–2020 & \cite{he2024interpretable} \\ 
        \midrule
           \multirow{10}{*}{Flood} 
                            & DEM                  & Carlisle, UK                             & Observation  & 5 m         & 1 hour    & 2005-2015          & \cite{kabir2020deep} \\ 
                            & AustraliaFlood       & Australia                                & Observation  & -           & Daily     & 1900-2018          & \cite{adikari2021evaluation} \\ 
                            & SekongFlood          & Vietnam, Laos, Cambodia                  & Observation  & -           & Daily     & 1981-2013          & \cite{adikari2021evaluation} \\ 
                            & BangladeshFlood      & Bangladesh (GBM river network)           & Observation  & -           & Daily     & 1979-2014          & \cite{ruma2023particle} \\ 
                            & GermanyFlood         & Germany, Sachsen                         & Radar        & 1 km        & 1 hour    & Different periods  & \cite{li2022prediction} \\ 
                            & ElbeRiverFlow        & Germany, Elbe River in Sachsen           & Observation  & -           & 1 hour    & Different periods  & \cite{li2022prediction} \\ 
                            & FijiFlood            & Fiji Islands                             & Observation  & -           & Daily     & 1990-2019          & \cite{moishin2021designing} \\ 
                            & FloridaFlood         & USA, Coastal South Florida               & Observation  & -           & 1 hour    & 2010-2020          & \cite{shi2024fidlar} \\ 
                            & QijiangRiverBasin    & China, Chongqing, Qijiang River          & Observation  & -           & 1 hour    & 1979-2020          & \cite{shao2024data} \\ 
                            & TunxiRiverBasin      & China, Anhui, Tunxi River                & Observation  & -           & 1 hour    & 1981-2007          & \cite{shao2024data} \\ 
        \midrule
            \multirow{12}{*}{Drought} 
                            & MODIS             & Regional, China   & Satellite         & 500 m                         & Monthly           &2000-2020              & \cite{zhang2023construction}\\
                            & CHIRPS            & Regional, China   & Satellite         & $0.05^\circ$                  & Monthly           &2000-2020              & \cite{zhang2023construction}\\
                            & ChinaDrought      & China             & -                 & -                             & Monthly           &1980-2019              & \cite{xu2022application}\\
                            & IndianDrought     & Peninsular, India & Satellite         & $0.25^\circ\times0.25^\circ$  & Daily             &1981-2021              & \cite{shukla2023deep}\\
                            & AVHRR             & Peninsular, India & Radiometer        & 1 km                          & Daily             &1981-2022              & \cite{shukla2023deep}\\
                            & ERA5              & East Asia         & Reanalysis        & $0.25^\circ\times0.25^\circ$  & 1 hour            &1970-2020              & \cite{zhang2024multiscale}\\
                            & EastAsiaDrought1  & East Asia         & Satellite         & $0.25^\circ$                  & Daily             &2003-2018              & \cite{park2020short}\\
                            & EastAsiaDrought2  & East Asia         & Satellite         & $0.05^\circ$                  & 16 days           &2003-2018              & \cite{park2020short}\\                        
                            & EastAsiaDrought3  & East Asia         & Satellite         & $0.05^\circ $                 & 8 days            &2003-2018              & \cite{park2020short}\\
                            & EastAsiaDrought4  & East Asia         & Simulation        & $0.5^\circ$                   & 3 hours           &2015-2018              & \cite{park2020short}\\
                            & EastAsiaDrought5  & East Asia         & Satellite         & 90 m                          & -                 &-                      & \cite{park2020short}\\       
                            & EastAsiaDrought6  & East Asia         & Satellite         & $0.5^\circ$                   & Yearly            &-                      & \cite{park2020short}\\              
        \midrule
            \multirow{10}{*}{Wildfire} 
                            & LANDFIRE PROGRAM  & California       & Satellite         & $128\times128$                 & 15 mins            &-                     & \cite{burge2023recurrent}\\
                            & FARSITE           & Regional         & Synthetic         & 30 m                           & 15 mins            &-                     & \cite{burge2023recurrent}\\     
                            & NASA-MODIS Terra  & California       & Satellite         & 1 km                           & 5 mins             & 2017-2018            & \cite{chowdhury2021mitigating}\\
                            & MERRA-2           & California       & Reanalysis        & $0.5^\circ\times0.625^\circ $  & 1 hour             &2017-2018             & \cite{chowdhury2021mitigating}\\  
                            & USGS              & Regional         & Satellite         & 30 m                           & -                  &2017-2018             & \cite{chowdhury2021mitigating}\\
                            & AICC              & Regional, Alaska & Satellite         & $400\times350$                 & Daily              &2002-2018             & \cite{marjani2023firepred}\\ 
                            & NRC               & Regional, Canada & Satellite         & 30 m                           & Daily              & 2002-2018            & \cite{marjani2023firepred}\\
                            & VIIRS             & South Africa     & Satellite         & 375 m                          & 1 hour             &2012-2014             & \cite{perumal2020comparison}\\
                            & VIIRS             & California       & Satellite         & 375 m                          & Daily              &2012-2021             & \cite{masrur2024capturing}\\    
                            & Percolation model & Regional         & Synthetic         &$110\times110$                  & 5 mins             &-                     & \cite{masrur2024capturing}\\
                                
        \bottomrule
    \end{tabular}
}
\label{tab:datasets}
\end{table*}

\section{Model Architectures}
\label{sec:model_arch}
%
%
\subsection{Convolutional Neural Networks}
Convolutional Neural Networks (CNNs)~\cite{lecun1995convolutional} are a specialized type of neural network designed for processing structured grid data, such as images. The convolutional layer usually utilizes convolutional kernels to process the input data, performing convolution operations to extract features like edges, textures, and patterns~\cite{li2021survey}.
This is often followed by a pooling layer to reduce the spatial dimensions of the feature maps, making the network computationally more efficient and focusing on the most important information.
%

They are widely used in tasks related to computer vision, such as image classification~\cite{he2016deep}, object detection~\cite{ren2016faster}, and segmentation~\cite{he2017mask}. 
Moreover, CNNs could be categorized into Conv1D, Conv2D, and Conv3D according to the sliding dimension of convolutional kernels~\cite{kiranyaz20211d}.

\subsection{Recurrent Neural Networks}
Recurrent Neural Networks (RNNs)~\cite{medsker2001recurrent} is a type of neural network particularly suited for tasks involving time-dependent or sequential data, such as time series forecasting~\cite{sbrana2020n}, natural language processing~\cite{mikolov2011extensions, zhang2017context}, and speech recognition~\cite{yadav2022survey}.
The key idea behind this is to recurrently learn from a sequence of data with an internal (hidden) state, which includes as inputs the previous hidden states and current input. The learning or update rule is:
\begin{equation}
    \begin{aligned}
        h_t &= \sigma(\mathbf{W}_x x_t + \mathbf{W}_h h_{t-1} + b_h), \\
        y_t &= \sigma(\mathbf{W}_y h_t + b_y),
    \end{aligned}
\end{equation}
where $h_t$ is the hidden state at $t$-th time step, $x_t$ is the input at $t$-th time step, $y_t$ is the output at the same time step, $\mathbf{W}_x$, $\mathbf{W}_h$, and $\mathbf{W}_y$ are the weight matrices, $b_h$ and $b_y$ are the biases, and $\sigma$ is the activation function (e.g., tanh or ReLU).
%

However, RNNs often suffer from gradient vanishing and gradient explosion while modeling long sequences. Long Short-Term Memory~\cite{hochreiter1997long} (LSTM) and Gated Recurrent Unit~\cite{chung2014empirical} (GRU) have been proposed to alleviate such a problem by well-designed gates to forget and filter information.

\subsection{Graph Neural Networks}
Graph Neural Networks (GNNs)~\cite{scarselli2008graph} is designed to work on graph-structured data, $\mathcal{G}=(\mathcal{V}, \mathcal{E})$, consisting of a set of nodes $\mathcal{V}$ and a set of edges $\mathcal{E}$. These nodes and edges represent the entities and the dependent relationships among these entities, respectively.
%
Spatio-temporal Graph Neural Networks (ST-GNNs)~\cite{yu2017spatio} is an extension of GNNs designed to model both spatial and temporal dependencies in dynamic graph-structured data changing over time, $\mathcal{G}_t=(\mathcal{V}, \mathcal{E}, t)$. 
Here, nodes $\mathcal{V}$ refer to spatial locations, and edges $\mathcal{E}$ refer to spatial relationships. Each node $v_t^{i}$ represents the feature vector at the corresponding location $i$ and time $t$. 
For each node, the message-passing technique~\cite{gilmer2017neural} is often employed to capture the spatial dependencies on its neighbors. The temporal dependencies between graph snapshots can be modeled with the sequential models aforementioned.
For the message passing, hidden states $h_t^i$ at each node are updated based on messages (feature vectors) $v_{t+1}^{i}$ according to:
\begin{equation}
    \begin{aligned}
        v_{t+1}^{i} &= \sum_{j \in N(i)} M_t(h_t^i, h_t^j, e_{ij}), \\
        h_{t+1}^{i} &= \sigma(h_t^i, v_{t+1}^{i}),
    \end{aligned}
\end{equation}
where in the sum, $N(i)$ denotes the neighbors of $i^{th}$ node in graph $\mathcal{G}$. After iterative updates $k$ time steps, the final output of the whole graph at time $t+k$ can be computed with a readout function $\mathcal{O}$:
\begin{equation}
    y_{t+k} = \mathcal{O}(\{h_{t+k}^i \mid i \in \mathcal{G}\}).
\end{equation}

\subsection{Transformer and Vision Transformer}
To overcome the limitations of RNNs, which stem from their inherent sequential processing, the Transformer model~\cite{vaswani2017attention} has emerged as a powerful alternative. Its core innovation lies in the use of parallel processing through the \emph{attention} mechanism, enabling it to capture dependencies between any parts of a sequence without the need for sequential steps~\cite{wen2022transformers}. The \emph{attention} mechanism is described as follows: 
\begin{equation}
    \text{Attention}(\mathbf{Q}, \mathbf{K}, \mathbf{V}) = \text{softmax}\left(\frac{\mathbf{Q} \mathbf{K}^T}{\sqrt{d_k}}\right)\mathbf{V},
\end{equation}
where the $d_k$ denotes the dimension of the key, $\mathbf{Q} \in \mathbb{R}^{n \times d_k}$, $\mathbf{K} \in \mathbb{R}^{m \times d_k}$, and $\mathbf{V} \in \mathbb{R}^{m \times d_v}$ are the query matrix, key matrix, and value matrix, respectively. These three matrices are computed by linear transformations from the original input sequence $\mathbf{X} \in \mathbb{R}^{n \times d}$ with learnable weight matrices $\mathbf{W}_q \in \mathbb{R}^{d \times d_k}$, $\mathbf{W}_k \in \mathbb{R}^{d \times d_k}$, $\mathbf{W}_v \in \mathbb{R}^{d \times d_v}$, as
\begin{equation}
    \mathbf{Q} = \mathbf{X} \mathbf{W}_q, \mathbf{K} = \mathbf{X} \mathbf{W}_k, \mathbf{V} = \mathbf{X} \mathbf{W}_v.
\end{equation}

\paragraph{Vision Transformer.} The Vanilla Transformer was originally proposed for dealing with sequences. 
Vision Transformer (ViT)~\cite{dosovitskiy2020image} is a variant tailed to process images and has shown powerful performance compared to convolutional neural networks (CNNs). 
ViT models divide the input image into a grid of smaller, non-overlapping patches. Each patch is treated similarly to a ``word" in natural language processing, and the patches are then flattened into vectors. 
Positional embeddings are added to these patch embeddings to mark the relative positions of patches in the image, helping models understand the image's spatial layout. Subsequently, the additive embeddings are fed into the Vanilla Transformer layer to leverage the \emph{attention} mechanism. We refer readers to look into Figure 1 in~\cite{dosovitskiy2020image}.

\subsection{Mamba and Vision Mamba}
We start by introducing the State Space Models (SSMs). SSMs represent the evolution of the system’s internal states and make predictions of what their next state could be. For sequence modeling, SSMs map a sequence $x(t) \in \mathbb{R}^L \mapsto y(t) \in \mathbb{R}^L$ through an implicit latent state $h(t) \in \mathbb{R}^{L \times N}$:
\begin{equation}
    \begin{aligned}
        h'(t) &= \mathbf{A}h(t) + \mathbf{B} x(t), \\
        y(t) &= \mathbf{C} h(t),
    \end{aligned}
\end{equation}
where $\mathbf{A} \in \mathbb{R}^{N \times N}$ and $\mathbf{B}, \mathbf{C} \in \mathbb{R}^{N \times 1}$ are learnable matrices. The continuous sequence is discretized by a step size $\Delta$, and the discretized SSM model is represented as:
\begin{equation}
    \begin{aligned}
        h_t &= \bar{\mathbf{A}} h_{t-1} + \bar{\mathbf{B}} x_t, \\
        y_t &= \mathbf{C} h_t,
    \end{aligned}
\end{equation}
where discretization rule can be achieved by zero-order hold~\cite{zhang2007comparison} $\bar{\mathbf{A}} = \exp(\Delta \mathbf{A})$ and $\bar{\mathbf{B}} = (\Delta \mathbf{A})^{-1}(\exp(\Delta \mathbf{A}) - \mathbf{I}) \cdot \Delta \mathbf{B}$.
The structured state-space model (S4), a variant of the vanilla SSM, improves long-range dependency modeling by utilizing the High-order Polynomial Projection Operators (HiPPO)~\cite{gu2020hippo}.

\paragraph{Mamba.} S4 applies the same parameters $\mathbf{A}$ and $\mathbf{B}$ to each ``token'' of input, which is challenging to identify the importance of each input.
Selective State Space Model (Mamba)~\cite{gu2023mamba} incorporates a selection mechanism such that parameters that affect interactions along the sequence are input-dependent (parameters $\Delta$, $\mathbf{A}$, $\mathbf{B}$ are functions of the input), enabling capturing contextual information in long sequences. 
Besides, Mamba possesses efficient hardware-aware designs. It utilizes three computing acceleration techniques (kernel fusion, parallel scan, and recomputation) to materialize the hidden state $h$ only in more efficient levels of the GPU memory hierarchy.

\paragraph{Vision Mamba.} Vision Mamba~\cite{zhu2024vision} is a variant of Mamba used for image modeling. Similar to Vision Transformer, Vision Mamba first splits the input image into patches and then projects them into patch tokens, but leverages bidirectional SSMs (Mamba blocks) to replace attention mechanisms as the image encoder to model the sequence of tokens. Therefore, Vision Mamba can be well-tailed for 2-D grid weather data, e.g., MetMamba~\cite{qin2024metmamba}.

\subsection{Generative Adversarial Networks}
Generative Adversarial Networks (GANs)~\cite{goodfellow2014generative,mirza2014conditional} were originally proposed to learn a generative model to generate realistic images via adversarial training. Specifically, GANs simultaneously train two neural networks adversarially: a \texttt{Generator G} and a \texttt{Discriminator D}. 
The Generator learns the underlying data distribution and generates produce samples that can effectively fool the discriminator, while the discriminator differentiates between the samples generated by the generator and the real samples by outputting the corresponding probabilities. 
This training process can be regarded as a two-player zero-sum game~\cite{washburn1995two}, ultimately ending when the discriminator is unable to distinguish between the generator-generated samples and the real samples, i.e., $D(x)=\frac{1}{2}$.

GANs have widely used for image generation~\cite{xu2018attngan}, super-resolution~\cite{harder2022generating}, style transferring~\cite{zheng2022p}, and image-based weather forecasting~\cite{chen2022dynamic, choi2023pct, cheng2023highway}.

\subsection{Diffusion Models}
Diffusion Models (DMs)~\cite{ho2020denoising,song2020denoising} are the other type of generative models that have gained significant popularity in computer vision~\cite{saharia2022palette, croitoru2023diffusion}, natural language processing~\cite{hertz2022prompt, li2023diffusion}, due to their ability to produce high-quality, realistic samples. Diffusion models work in two processes: \emph{forward diffusion process} and \emph{reverse denoising process}. 
In the forward process, data (e.g., an image) is gradually “noised” by adding small amounts of Gaussian noise over multiple steps until it becomes nearly pure noise. This process is usually fixed and non-learnable, where each step incrementally increases the noise.
The reverse process is learnable, where the model learns how to gradually remove noise, step-by-step, to recover a realistic sample from a noisy starting point. This iterative denoising process helps to learn the intricate, high-dimensional data distribution.

Mathematically, the \textit{forward process} transforms an input  $\mathbf{x}_0$ with a data distribution of $q(\mathbf{x}_0)$ to a white Gaussian noise vector $\mathbf{x}_N$ in $N$ diffusion steps. It can be described as a Markov chain that gradually adds Gaussian noise to the input according to a variance schedule $\{\beta_1, \dots, \beta_N\} \in (0, 1)$:
\begin{equation}
    q(\mathbf{x}_{1:N} \mid \mathbf{x}_0) = \prod_{n=1}^N q(\mathbf{x}_n \mid \mathbf{x}_{n-1}),
\end{equation}
where at each step $n \in [1, N]$, the diffused sample $\mathbf{x}_n$ is obtained with $q(\mathbf{x}_n \mid \mathbf{x}_{n-1}) = \mathcal{N} \left( \mathbf{x}_n; \sqrt{1 - \beta_n} \mathbf{x}_{n-1}, \beta_n \mathbf{I} \right)$.

In the \textit{reverse process}, the \textit{denoiser network}, $p_\theta(\cdot)$, is used to recover $\mathbf{x}_0$ by gradually denoising $\mathbf{x}_n$ starting from a Gaussian noise $\mathbf{x}_N$ sampled from $\mathcal{N}(0, \mathbf{I})$. This process is presented as:
\begin{equation}
    p_\theta(\mathbf{x}_{0:N}) = p(\mathbf{x}_N) \prod_{n=1}^N p_\theta(\mathbf{x}_{n-1} \mid \mathbf{x}_n).
\end{equation}

In weather and climate domains, diffusion models have been applied to precipitation nowcasting~\cite{asperti2308precipitation, gao2024prediff}, atmospheric downscaling~\cite{ling2024diffusion, mardani2023generative}, weather forecasting~\cite{shi2024codicast, andrae2024continuous}.

\end{document}